\documentclass[10pt,twocolumn,letterpaper]{article}

\usepackage[pagenumbers]{cvpr} %

\usepackage[utf8]{inputenc} %
\usepackage[T1]{fontenc}    %
\usepackage{url}            %
\usepackage{booktabs}       %
\usepackage{amsfonts}       %
\usepackage{nicefrac}       %
\usepackage{microtype}      %
\usepackage{xcolor}         %
\usepackage{wrapfig}

\usepackage{multirow}

\definecolor{cadmiumgreen}{rgb}{0.0, 0.42, 0.24}
\definecolor{darkpink}{rgb}{0.91, 0.33, 0.5}
\definecolor{mygreen}{RGB}{20, 99, 4}
\definecolor{ForestGreen}{RGB}{27,132,27}
\definecolor{ultramarine}{RGB}{0,32,96}

\usepackage{listings}
\usepackage{arydshln}

\usepackage{adjustbox}

\usepackage{booktabs}
\usepackage{soul}

\usepackage{graphicx}
\usepackage{subcaption}

\definecolor{cvprblue}{rgb}{0.21,0.49,0.74}
\usepackage[pagebackref,breaklinks,colorlinks,allcolors=cvprblue]{hyperref}

\def\paperID{14181} %
\def\confName{CVPR}
\def\confYear{2026}

\title{Hallucination Score: Towards Mitigating Hallucinations in Generative Image Super-Resolution}

\vspace{-1mm}
\author{%
{Weiming Ren}$^{1*\dagger}$
\hspace{4mm}%
{Raghav Goyal}$^{2*}$
\hspace{4mm}%
{Zhiming Hu}$^{2*}$
\hspace{4mm}%
{Tristan Aumentado-Armstrong}$^{2*}$
\\
{Iqbal Mohomed$^{2}$}
\hspace{14mm}
{Alex Levinshtein$^{2}$}
\\[1mm]
{$^{1}$University of Waterloo} \hspace{13mm} {$^{2}$AI Center -- Toronto, Samsung Electronics}\\
{ \centering
{\tt\small w2ren@uwaterloo.ca}, %
{\tt\small \{raghav.goyal, zhiming.hu, tristan.a, i.mohomed, alex.lev\}@samsung.com} 
}
}

\begin{document}

\twocolumn[{
\renewcommand\twocolumn[1][]{#1}
\maketitle
\begin{center}
    \captionsetup{type=figure}
    \includegraphics[width=\textwidth]{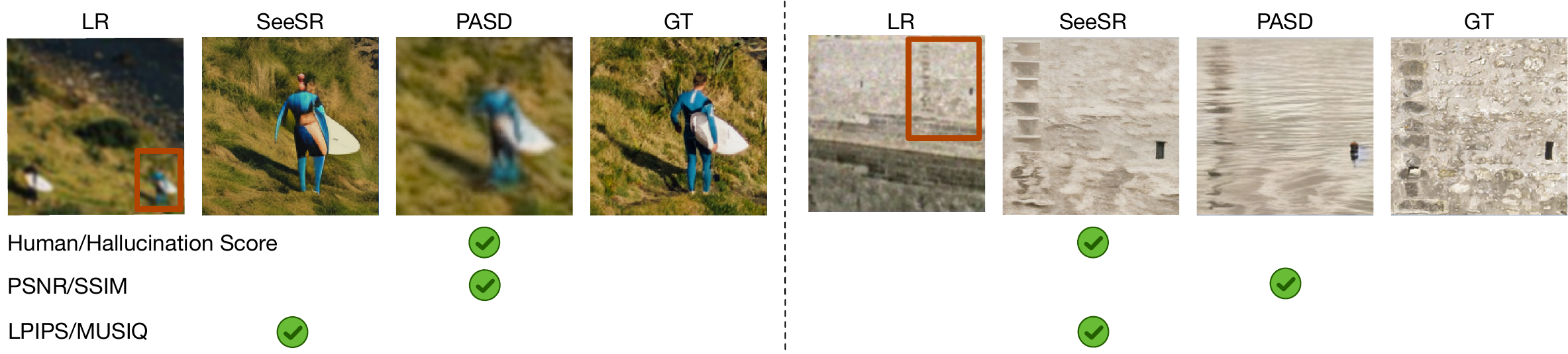}
\captionof{figure}{\textbf{Hallucination score for image super-resolution}. The outputs of state-of-the-art super-resolution (SR) models (e.g., SeeSR~\cite{wu2024seesr} and PASD~\cite{yang2024pasd}) often contain significant hallucinations, as seen in the example images above. %
For each example set, we show the outputs of two SR models and 
    the \textit{preference} of a given metric for each output, via a green checkmark in its row;
    for instance, in the left inset, LPIPS prefers the SeeSR output, while SSIM favours the PASD one. 
While human evaluators and our proposed \textit{hallucination score} (HS) can identify hallucinatory outputs,
traditional metrics (PSNR, SSIM, MUSIQ, and LPIPS) often fail 
to do so. %
Further, notice that the HS does not always align with existing metrics, as it captures complementary aspects of SR quality.
}
    \label{fig:teaser}
\end{center}
\vspace{0.9mm}
}]

\def\thefootnote{*}\footnotetext{Equal primary contribution}
\def\thefootnote{$\dagger$}\footnotetext{Work done as an intern at AI Center -- Toronto, Samsung Electronics}

\begin{abstract}
Generative super-resolution (GSR) currently sets the state-of-the-art in terms of perceptual image quality, overcoming the ``regression-to-the-mean'' blur of prior non-generative models. However, from a human perspective, such models do not fully conform to the optimal balance between quality and fidelity. Instead, a different class of artifacts, in which generated details fail to perceptually match the low resolution image (LRI) or ground-truth image (GTI), is a critical but under-studied issue in GSR, limiting its practical deployment. In this work, we focus on measuring, analyzing, and mitigating these artifacts (\ie, ``hallucinations''). We observe that hallucinations are not well-characterized with existing image metrics or quality models, as they are orthogonal to both exact fidelity and no-reference quality. Instead, we take advantage of multimodal large language models (MLLMs) by constructing a prompt that assesses hallucinatory visual elements and generates a ``Hallucination Score'' (HS). We find that HS is closely aligned with human evaluations, and also provides complementary insights to prior image metrics used for super-resolution (SR) models. Finally, we propose a few efficient HS proxies and demonstrate how diffusion-based GSR models can be fine-tuned to mitigate hallucinations, leveraging HS proxies as differentiable reward functions.
\end{abstract}

\vspace{-1.0em}
\section{Introduction}
\label{sec:intro}

Single-image super-resolution (SR) is inherently ill-posed, with every low-resolution (LR) input corresponding to a multimodal distribution of possible high-resolution (HR) solutions \cite{schultz1994bayesian}. For standard regressive (\ie, non-generative) models, outputs are integrated over the solution space, resulting in blurriness. This is a natural consequence of training with pixel-space reconstruction losses, which attain their optima via averaging possible solutions in pixel space; this induces the so-called ``regression-to-the-mean'' effect (\eg, \cite{bruna2016super,delbracio2023inversion}). While perceptual metrics (\eg, \cite{zhang2018unreasonable,ding2020image}) can reduce this problem, they cannot fully remove it.

In contrast, for GSR methods, the model can ``sample'' a particular solution, with much less impact from such averaging~\cite{delbracio2023inversion}. %
This leads to improved realism, better image quality, and less blurriness (\eg, \cite{wang2024exploiting,wu2024seesr,gao2023implicit,yang2024pasd,moser2024diffusion}). Further, it allows sampling multiple solutions (\ie, ``explorable'' SR \cite{bahat2020explorable}). However, a different problem naturally arises, referred to as ``hallucinations'': unlike the blurry outputs that characterize uncertainty for regressive models, GSR can output images that are sharp and detailed, yet completely \textit{incorrect} and \textit{perceptually jarring} (see Fig.~\ref{fig:teaser}). 
Such solutions may be plausible according to the data manifold learned by the GSR model; however, they are often perceptually unacceptable.
In some cases, hallucinations can completely change the semantic meaning of the image, while in others they can severely alter the geometric interpretation of the scene.

\begin{figure}[t]
    \centering
    \includegraphics[width=0.99\linewidth]{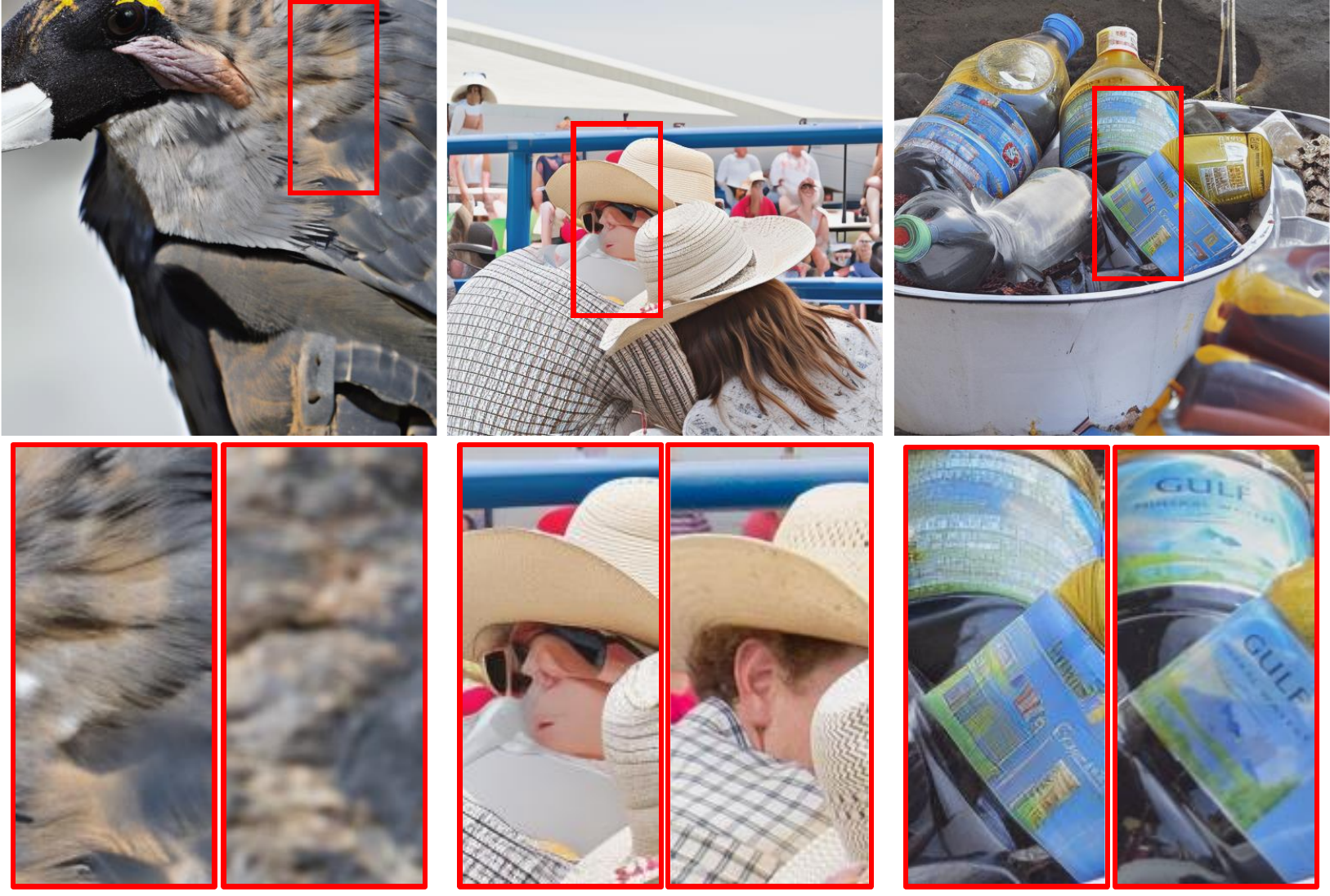}
    \vspace{-0.3em}
    \caption{
    \textbf{Examples of hallucinations.}
    Top: SeeSR outputs \cite{wu2024seesr}; bottom: zoom-ins of SR (left) with GT (right).
    From left to right, we see:
    (i) \textit{incorrect semantics}, 
        wrongly adding feathers to the stone;
    (ii) \textit{visually jarring scene alterations},
        despite coarse semantic preservation;
    and
    (iii) \textit{textual artifacts}. 
    Notice the textures appear realistic and sharp, but are perceptually unappealing.
    }
    \vspace{-0.5em}
    \label{fig:qualhal}
    
\end{figure}

The consequence of hallucinated content is severe: for instance, in real-world settings, such as digital zoom on cameras or mobile phones, current GSR models cannot be trusted to output acceptable details -- the risk of alienating users with perceptually damaged content, worse than simple blur, is too high. 
Such models can completely change text or alter faces to different identities as well (see Fig.~\ref{fig:qualhal}). Ideally, therefore, we could identify such problematic model outputs, to help us design more trustworthy GSR approaches.

However, %
these issues are non-trivial to detect and characterize.
While low-level metrics (\eg, $L_2$ distance, SSIM \cite{wang2004image}) will detect such hallucinations, they do not allow for perceptually plausible variations from the ground truth which are required in GSR. Indeed, it is well-known that such metrics correlate poorly with human sensibilities 
(\eg, \cite{zhang2018unreasonable,girod1993mse,mannos1974effects}).
Differently, recent full-reference (FR-IQA)~\cite{ding2021comparison} and no-reference (NR-IQA) \cite{ke2021musiq,wu2024qalign} image quality assessment metrics  allow for perceptually plausible variations from the ground-truth image, but they cannot detect hallucinations effectively. 
FR-IQA metrics do not capture the various semantic and perceptual factors 
that characterize subjective judgments of SR output quality 
(as we demonstrate in \S\ref{sec:analysis}). 
NR-IQA metrics will not detect details as hallucinatory as long as the \textit{quality} of the details is high.
Thus, existing approaches cannot effectively detect GSR hallucinations and allow for perceptually plausible differences at the same time; indeed,
        as shown in Fig.~\ref{fig:teaser},
        they may agree or disagree with human judgment, depending on the scenario.

In this work, we aim to bridge this gap by constructing an automated rater that detects hallucinations and allows for semantically plausible perceptual differences from ground-truth based on recent powerful multimodal large language models (MLLMs). It is called \textit{hallucination score} (HS),
which we show correlates well to human perceptual decisions.
We examine the existing image distance and similarity metrics, confirming that they correlate poorly with our measure; however, we observe that certain semantics-aware deep features (\eg, DINOv2~\cite{oquab2023dinov2} and CLIP~\cite{radford2021learning}) correlate the best with HS.
Motivated by these analyses, we
propose a scalable and differentiable approach to reduce the hallucinations based on those strong semantic representations.

We summarize our contributions as follows:
(i) {we define hallucinations in the GSR context, and devise our MLLM-based HS to measure them};
(ii) {we conduct user studies and extensively analyze existing image metrics, similarity measures, and quality models, finding that HS
    (a) closely correlates to human opinion, and 
    (b) forms a complementary evaluation dimension}; 
and 
(iii) We propose a few proxies that can effectively approximate MLLM-based HS and human ratings. Using differentiable HS proxies, we demonstrate how to 
directly reduce GSR hallucinations through reward back-propagation, 
without sacrificing realism or fidelity.

\section{Related Work}
\label{sec:relwork}

\begin{figure*}[ht]
    \centering
    \includegraphics[width=\linewidth]{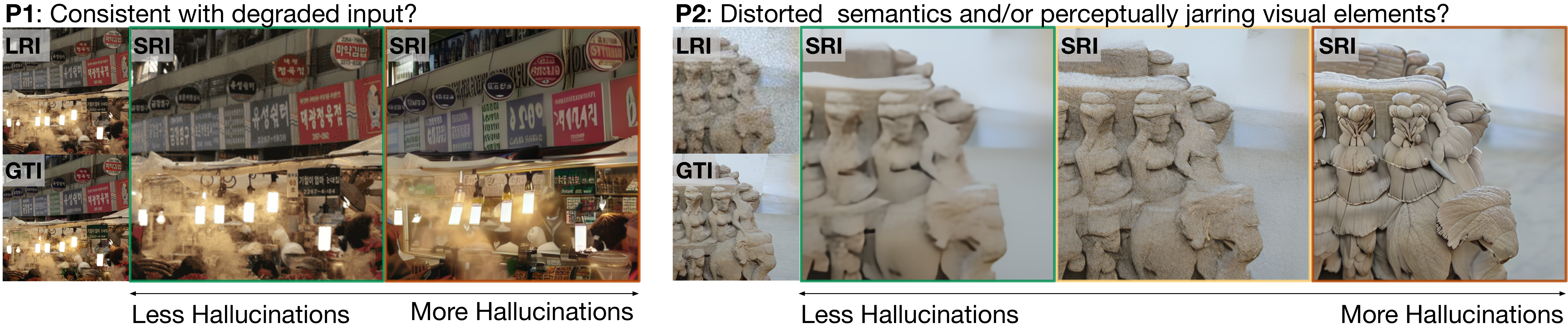}
    \vspace{-1.6em}
    \caption{
        \textbf{Illustration of our hallucination definition.}
        Property \textbf{P1} defines SRI content as hallucinatory if it cannot be plausibly degraded into LRI content.
        Property \textbf{P2} considers a continuum from
            blurred content (due to uncertainty)
            and/or innocuous detail changes
            (less hallucinatory)
            to
            perceptually salient
            and/or semantically severe
            distortions
            (highly hallucinatory). %
    }
    \label{fig:defillus}
    \vspace{-0.7em}
\end{figure*}

\noindent
\textbf{Generative SR.}
While generative adversarial networks (GANs) (\eg, \cite{wang2021real,wang2018esrgan,li2024sed,ledig2017photo,park2023content}) and other techniques
(\eg, \cite{lugmayr2020srflow,yao2023local,guo2022lar,zhang2025augmenting})
have improved results in GSR,
the most successful recent models have been diffusion-based (\eg, \cite{wang2024exploiting, yang2024pasd, wu2024seesr, lin2024diffbir,noroozi2024you,wu2024one,moser2024diffusion, sun2024pisasr, chen2025faithdiff}).
For instance, recent approaches such as StableSR~\cite{wang2024exploiting}, PASD~\cite{yang2024pasd}, and SeeSR~\cite{wu2024seesr} have employed conditional diffusion models that leverage features or tags extracted from LR images to guide the SR process. 
The fundamental appeal of using generative models is two-fold: 
(a) it directly tackles the ``regression-to-the-mean'' problem (\eg, \cite{jo2021tackling,delbracio2023inversion}) and (b) it enables better controllability via sampling (\ie, ``exploration'' \cite{bahat2020explorable}). 
However, LR-derived control signals are often noisy 
(\eg, incorrect semantics extracted from LR), 
which may cause hallucinations in the generated high-resolution content. 
Our analysis reveals several instances where these methods fall prey to this issue.
In our work, we specifically target this problem, aiming to improve existing diffusion-based GSR.

\vspace{0.3em}
\noindent
\textbf{Image Quality Assessment Metrics.} %
SR losses and evaluations necessarily span across reconstruction fidelity and perceptual quality, due to the tradeoff between them \cite{blau2018perception,blau2019rethinking}.  
Common low-level full-reference (FR) distortion measures include 
$L_p$ distances, %
SSIM \cite{wang2004image}, and others (\eg, frequency-domain \cite{fuoli2021fourier,liu2023spectral,wang2023spatial,chen2024low}, uncertainty-aware \cite{ning2021uncertainty}, edge-focused \cite{ma2020structure,sun2008image}).
In contrast, especially in GSR (\eg, \cite{wu2024seesr,yang2024pasd}),
perceptual evaluations rely on NR-IQA models 
(\eg, \cite{wu2024qalign,agnolucci2024arniqa,chen2024topiq,ke2021musiq,ying2020patches,mittal2012making,heusel2017gans}),
which examine \textit{general} image quality, 
though SR-specific ones also exist
\cite{khrulkov2021neural,ma2017learning}.
Others have considered NR artifact detection via image statistics
(\eg, \cite{liang2022details,xie2023desra}).
Finally, perceptually oriented FR-IQA metrics \cite{ding2021comparison}, 
which generally compare neural embeddings, 
balance distortion with NR quality: 
\eg,  LPIPS \cite{zhang2018unreasonable} and its variants \cite{ghazanfari2023r,kettunen2019lpips,ghildyal2022stlpips}, DISTS \cite{ding2020iqa}, and others (\eg, \cite{johnson2016perceptual,fu2023dreamsim,rad2019srobb,mechrez2019maintaining,you2024descriptive}).
Other editing tasks %
also compare images via semantics, such as CLIP \cite{radford2021learning} similarity (\eg, \cite{brooks2023instructpix2pix,mirzaei2024watch}),
or segmentations (\eg, \cite{myers2020semantic,cherian2019sem}).
In this work, we focus on %
\textit{hallucinations}, related to the %
degree of perceptual ``wrongness'' a restoration incurs, 
in the context of the low-resolution and ground-truth image.
Without a reference, NR-IQA cannot account for this context; conversely, existing FR-IQA fails to combine the
low-level, semantic, and perceptual aspects 
necessary to measure hallucinations.

\vspace{0.3em}
\noindent
\textbf{Hallucination Mitigation in Image Generation.} 
In the unconditional generation context, hallucinations can be defined as ``non-factual'' outputs (\eg, \cite{lim2024evaluating}); however, this perspective is less applicable to SR, where the primary concern is the trade-off between perceptual quality and reconstruction fidelity \cite{blau2018perception}. 
Other prior works \cite{aithal2025understanding,cohen2025looks} relate hallucinations to the fundamental limitations of generative models. 
Specifically, Aithal et al.~\cite{aithal2025understanding} define hallucinations as image content that is out-of-distribution with respect to the training data. 
However, this does not account for the perceptual (\ie, human) aspects of hallucinations, nor for the specific reference-based structure of SR. Separately, others~\cite{cohen2025looks} have considered hallucination as synonymous with entropy (\ie, the uncertainty that induces incorrect but realistic details), and thus closely relates to the perception-distortion tradeoff. 
While this approach relates closely to ours, in that incorrect but realistic details may also be hallucinatory under our definition, it does not necessarily differentiate between various (wrong but realistic) details that humans would judge very differently in terms of quality (\ie, quantifying subjective degrees of hallucination). Further, estimating entropy for real-world image sizes remains an open research problem. In contrast, our method focuses on the perceptual facets of GSR, and we devise a practical method of measuring hallucinations, via modern MLLMs, that is sensitive to the \textit{level} of spurious content present.

\begin{figure*}[htb]
    \centering
    \includegraphics[width=\textwidth]{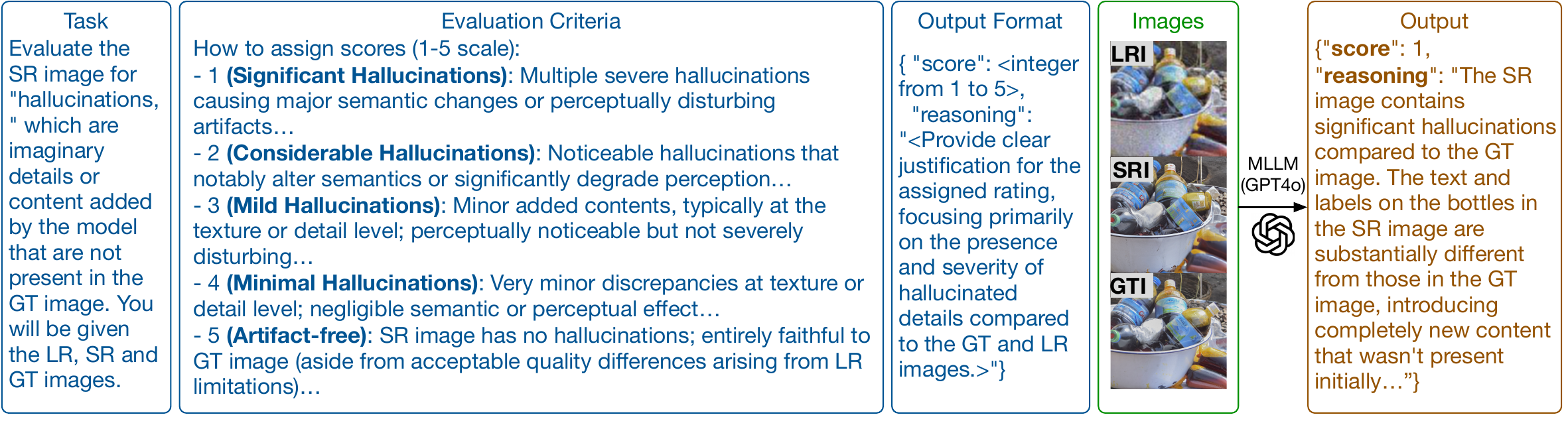}
    \vspace{-1.5em}
    \caption{\textbf{Generating hallucination scores with GPT-4o.} We construct a prompt comprising three essential parts: task introduction, evaluation criteria, and output format. This detailed prompt is then combined with input images and fed into the MLLM model (GPT-4o \cite{hurst2024gpt}) to obtain hallucination scores and accompanying explanations. The full prompt can be found in Supp.~Fig.~\ref{fig:promptfull}.
    }
    \label{fig:prompt}
    \vspace{-0.3em}
\end{figure*}

\section{Defining and Characterizing Hallucinations}
\label{sec:def}

In the context of GSR, hallucination refers to the generation of  image content that is perceptually ``incorrect'', 
\textit{relative to} 
(i) the low-resolution input image (LRI), and
(ii) the ground-truth high-resolution reference image (GTI).
Specifically, we define hallucinations in a super-resolved image (SRI) to have the following properties (see also Fig.~\ref{fig:defillus}):

\noindent    
$\bullet\,$\textbf{P1}:~SRI content that could not be plausibly present in the LRI is necessarily a hallucination.

\noindent
$\bullet\,$\textbf{P2}:~SRI content that differs from the GTI is  hallucinatory 
    to the extent that the generated visual elements are semantically different or
    perceptually recognizable
    as anomalous.

Property \textbf{P1} is simply inherited from the SR problem itself, demanding there exists some realistic degradation that maps the SRI to the LRI.
Property \textbf{P2}, however, fundamentally relies on the subjective judgment of human visual perception.
It does \textit{not} ask that the SRI shares the exact details of the GTI; for instance, new textural details that a human observer would not notice as out-of-place are acceptable (non-hallucinatory or low hallucinatory).

However, if the added details changed the \textit{semantics of the scene} (\eg, significant alterations of scene elements) or generated \textit{perceptually unpleasant details} (\eg, incorrect facial features, unreadable or distorted text) when compared to LRI or GTI, they should be labeled as hallucinations. 
Importantly, this definition is orthogonal to general image quality (\eg, NR-IQA), yet does not demand reconstructive preservation of the GTI. For instance, a regressive SR model that outputs a blurry image could have low image quality, but also no hallucinations (see ``Bicubic'' in Table \ref{tab:results_injection}). Conversely, a GSR model can have high general quality (\ie, sharp generated details), but could have a hallucination level that is low (details do not seem out-of-place, whether or not they match the GTI) or high (details are obviously anomalous).

\subsection{MLLM-based Hallucination Scoring} 
\label{subsec:mllmhalluscore}

\begin{figure*}[t]
    \centering
    \includegraphics[width=0.99\textwidth]{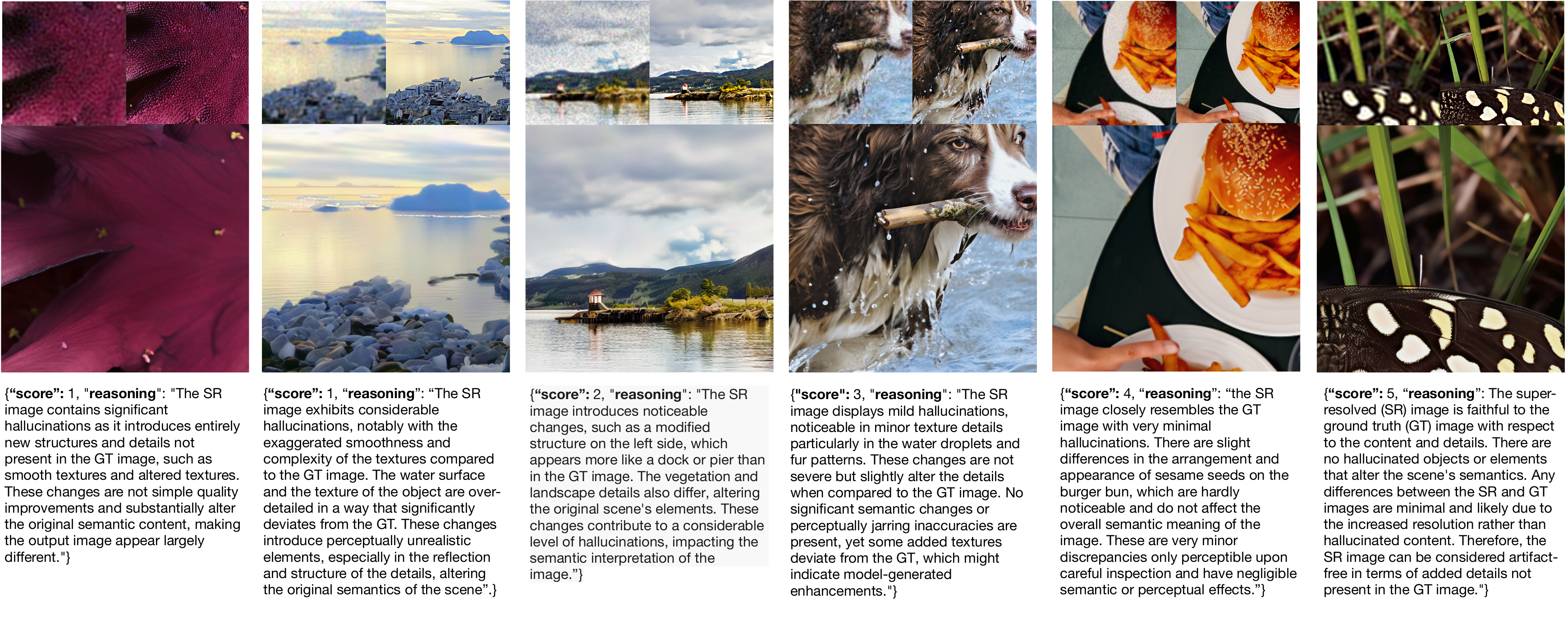}
    \vspace{-0.5em}
    \caption{
    \textbf{Qualitative examples of our MLLM-based hallucination score.}
    In this figure, we show six example outputs from the MLLM given the LRI (top-left), GTI (top-right), SRI (bottom) and the prompt as inputs. Each output includes a numerical score on a 1-5 scale with detailed explanations justifying the assigned score. The results demonstrate the MLLM's ability to effectively identify critical hallucination issues in each image and assign accurate hallucination scores accordingly.}
    \vspace{-0.5em}
    \label{fig:mllm_sample_outputs}
\end{figure*}

While human-rated image quality assessment (IQA) is the gold standard, it is fundamentally unscalable across datasets and models, especially as both evolve. As such, we investigate the use of an Multimodal Large Language Model (MLLM) for generating scores that mimic human judgments, according to the definition above. %
We use GPT-4o \cite{hurst2024gpt} as our primary model, but also test Qwen2.5-VL \cite{qwen,bai2025qwen2} (though in \S\ref{sec:correlationan}, we find it has lower correlation to human judgment), as our method is agnostic to the choice of MLLM. 
To query the model, we design a tailored prompt that incorporates an description of the task of hallucination scoring, as well as an evaluation criteria and output format as shown in Fig.~\ref{fig:prompt}. 
The model outputs both a numerical score, which we call the GPT-HS, and 
a justification for its decision (\ie, an explanation of its estimate),
given the LRI, SRI, GTI, and prompt. 
The HS describes the level of hallucination as an integer from 1-5, with $1$ indicating significant semantic alterations or jarring effects, and $5$ representing minimal or no hallucination. 
The complete prompt can be found in Supp.~Fig.~\ref{fig:promptfull}.
Illustrative example outputs from the MLLM are shown in Fig.~\ref{fig:mllm_sample_outputs}.
These demonstrate the model's ability to detect semantic changes and identify disturbing scenes in the SRIs, yielding scores that accurately reflect the extent of hallucination present 
(see Supp.~\S\ref{supp:qualitative} for more examples).

\subsection{Efficient Proxies for Hallucination Scoring}
\label{subsec:hs:proxy}
MLLMs provide state-of-the-art results on many tasks, but are computationally inefficient and memory-intensive.
There are also complications in their use as differentiable optimization targets, as we consider in \S\ref{sec:results}.
We therefore investigate training efficient and differentiable proxies for HS estimation, \textit{using  MLLM-based model to generate training data}. 

\noindent
\textbf{MLLM-HS Dataset.}
We build a dataset of $\sim$31K pairs of SRIs 
(from Swin2SR \cite{conde2022swin2sr}, SeeSR \cite{wu2024seesr}, PASD \cite{yang2024pasd}, and StableSR \cite{yang2024pasd}) with associated GPT-derived HSs, from
LSDIR \cite{li2023lsdir}, DIV2K \cite{agustsson2017ntire}, DIV8K \cite{gu2019div8k}, and Flickr2K \cite{timofte2017ntire}.
However, we ensure that 
(i) models are never run on their own training data
and (ii) there is no overlap with our analysis and evaluation datasets (see Supp.~\S\ref{supp:proxy:dataset}).

\noindent
\textbf{HS Proxy Designs.}
We consider three architectures:
a convolutional neural network (CNN), 
an adaptation of a DINO-based deep feature metric,
and the open-weights MLLM
{Qwen2.5-VL-7B}
(see Supp.~\S\ref{supp:proxy:model} for details).

\noindent
$\bullet\,$\textit{CNN}:
starting from an ImageNet-pretrained ResNet-50 (RN50) \cite{he2016deep}, we modify the first layer to take nine channels (LQ, GT, and SR) and the last to output the scalar HS. 

\noindent
$\bullet\,$\textit{DINO-HS}:
{we devise a simple approach for calculating image similarity} via deep features, which we fine-tune to reproduce the HS.
Denote the estimated HS via 
$\widehat{h} = h_s(S_c(f(I_\mathrm{SR}),f(I_\mathrm{GT})))$, where $f$ is a DINO-based feature extractor \cite{darcet2023vision}, $S_c$ is cosine similarity, and $h_s$ alters the similarity to match the HS.
For stability, similar to prior work (\eg, \cite{xu2023imagereward}), we only allow a subset of layers of $f$ to be trained.
Our use of deep features is motivated by our findings in \S\ref{sec:correlationan} that a metric based on such semantics-aware models, like DINO \cite{caron2021emerging,oquab2023dinov2}, naturally correlates to HS.

\noindent
$\bullet\,$\textit{Qwen-HS}:
we also fine-tune the smaller, open-weights MLLM Qwen2.5-VL-7B (denoted Qwen-HS). We use the same GPT-derived dataset for training, as GPT-HS correlates better to human scores than untuned Qwen2.5-VL-7B (\S\ref{sec:correlationan}). 
More specifically, we apply standard supervised fine-tuning, where not only the score but also the explanatory text  {(\ie, the reasoning)} are used to train the model.

\section{Metric Analysis}
\label{sec:analysis}

We first demonstrate that HS correlates well with human opinion, including our trained proxies (which build on the GPT-based HS), while existing metrics are insufficiently sensitive to hallucinations.
Additional analysis finds that HS is complementary to these metrics.
Altogether, these suggest  
(i) the utility of HS for evaluation
and
(ii) the potential of our proxies for fine-tuning GSR models to mitigate hallucinations, without necessarily damaging performance according to traditional metrics, as we show in \S\ref{sec:results}.

\subsection{Existing Metrics and Similarities}
\label{sec:existingmetrics}

We first investigate the relation of existing image metrics, similarities, and quality measures to hallucinations.
To this end, we comprehensively analyze a variety of such methods commonly employed in SR (see Supp. \S\ref{supp:andetails} for details):  

\newcommand{\bl}{$\bullet$~}
\vspace{0.2em}
\noindent
\bl\textit{Pixel-Level Distortion.}
We use mean-squared error (MSE) and SSIM \cite{wang2004image} to measure low-level colour-space distance.

\vspace{0.2em}
\noindent
\bl\textit{FR-IQA Metrics.}
We consider the commonly used LPIPS \cite{zhang2018unreasonable} and DISTS \cite{ding2020image} metrics, which are sensitive to textures and other mid-level visual signals.

\vspace{0.2em}
\noindent
\bl\textit{NR-IQA Metrics.}
We apply the popular MUSIQ \cite{ke2021musiq} model to estimate SR image quality. 
In addition, we measure sharpness via the Laplacian magnitude (\textit{\eg}, \cite{gallego2019focus}); this also enables us to see which models incur blur when the output is uncertain (\ie, regression-to-the-mean).

\vspace{0.2em}
\noindent
\bl\textit{Semantic Segmentation Divergence (SSD).}
Since a semantic class change often implies hallucinatory content, a natural approach is estimate the categorical changes between the GTI and SRI. 
To do so, we extract tags or common object categories on the GTI using the Recognize Anything model (RAM++ \cite{zhang2023recognize,huang2023open}), 
segment with OpenSeeD \cite{zhang2023simple}, and compute the mean per-pixel KL divergence.

\vspace{0.1em}
\noindent
\bl\textit{Neural Feature Distance.}
We extract features via two well-known visual encoders: 
DINO \cite{caron2021emerging,oquab2023dinov2} and CLIP \cite{radford2021learning},
specifically DINOv2 with registers \cite{darcet2023vision} and OpenCLIP \cite{cherti2023reproducible}.
In both cases, we consider both the spatial tokens (\texttt{*-ST}) and class token (\texttt{*-CLS}), along with the use of intermediate layers (\texttt{*-interm}).
We then compute the cosine distance on the GTI and SRI features. 

\vspace{0.1em}
\noindent
\bl\textit{Neural Correspondence Features.}
Hallucinations relate closely to semantic correspondences, in that they are often perceptually difficult to relate back to the GTI. %
Hence, we 
build off
a recent correspondence model, TLR \cite{zhang2024telling}, which combines StableDiffusion 1.5 \cite{rombach2022high} and DINOv2 \cite{oquab2023dinov2} features, as well as DeepViT \cite{amir2021deep}, which relies on multi-scale log-binned DINOv1 \cite{caron2021emerging} features.

\subsection{Correlation Analyses}
\label{sec:correlationan}

\newcommand{\blap}[1]{\vbox to 0pt{\hbox{#1}\vss}}

{
\setlength{\tabcolsep}{0.04in}
\fontsize{9}{10}\selectfont
\begin{table*}[t]
    \centering
    \aboverulesep = 0mm
    \belowrulesep = 0mm %
    \caption{
        \textbf{Correlations to Human Judgments.}
        We show Pearson ($\rho_P$) and Spearman ($\rho_S$) correlations between human scores (aggregated per image via mean or majority) and a variety of image metrics and similarities (see \S\ref{sec:existingmetrics}). 
        We find that our GPT-based HS, as well as our proxies trained with GPT-HS-derived data, generally have the highest correlations, with deep feature distances (particularly DINO) closely following.
        These motivate our claims that
        (i) existing methods do not capture human notions of hallucination (and thus our HS can act as a complementary evaluation) and 
        (ii) our proxies have potential as optimization targets. See Supp.~\S\ref{supp:human} for more details.
    }
    \begin{adjustbox}{max width=0.99\linewidth}
    \begin{tabular}{cc|ccccccccccccccccc|ccc} %
    \toprule
    & \multirow{2}{*}{\shortstack[c]{Human\\ Score}} & 
    \multirow{2}{*}{PSNR} & \multirow{2}{*}{SSIM} & \multirow{2}{*}{DISTS} & \multirow{2}{*}{LPIPS} & \multirow{2}{*}{MUSIQ} & \multirow{2}{*}{Sharpness} & \multirow{2}{*}{SSD} & \multirow{2}{*}{DeepViT} & \multirow{2}{*}{TLR} & \multicolumn{3}{c}{DINO} & \multicolumn{3}{c}{CLIP} & \multirow{2}{*}{Qwen-7B} &
    \multirow{2}{*}{GPT-HS} & \multicolumn{3}{c}{HS Proxies (via GPT-HS)} \\
    \cmidrule(ll){12-14} \cmidrule(ll){15-17} \cmidrule(ll){20-22}
     & & & & & & & & & &  & ST  & CLS & interm & ST & CLS & interm & & & CNN & DINO-HS & Qwen-HS \\
     \midrule
    \multirow{2}{*}{$\rho_P$} &
    Mean 
    & 0.36 & 0.28 & -0.09 & 0.23 & -0.16 & -0.12 & 0.08 & 0.37 & 0.37 & 0.38 & 0.31 & 0.53 & 0.46 & 0.53 & 0.47 & 0.42 & 0.55 & 0.49 & 0.68 & 0.66
     \\
    & Majority 
    & 0.30 & 0.21 & -0.11 & 0.16 & -0.19 & -0.10 & 0.04 & 0.35 & 0.32 & 0.35 & 0.26 & 0.45 & 0.41 & 0.50 & 0.42 & 0.37 & 0.50 & 0.43 & 0.62 & 0.60
     \\
    \midrule
     \multirow{2}{*}{$\rho_S$} &
     Mean &  
         0.37 & 0.28 & -0.09 & 0.25 & -0.17 & -0.24 & 0.15 & 0.38 & 0.35 &
         0.40 & 0.34 & 0.57 & 0.45 & 0.50 & 0.48 & 0.43
          & 0.56 &  0.51 & 0.71 & 0.70
     \\
     & Majority & 
        0.27 & 0.18 & -0.12 & 0.17 & -0.18 & -0.21 & 0.07 & 0.35 & 0.29 &
        0.36 & 0.29 & 0.47 & 0.40 & 0.47 & 0.42 & 0.37
         & 0.51 &  0.44 & 0.63 & 0.62
     \\
    \bottomrule
    \end{tabular}
    \end{adjustbox}
    \label{tab:humancorrelations}
\end{table*}
}

{
\setlength{\tabcolsep}{0.04in}
\fontsize{9}{10}\selectfont
\begin{table*}[t]
    \centering
    \aboverulesep = 0mm
    \belowrulesep = 0mm %
    
    \caption{
        \textbf{Correlations to GPT-derived Hallucination Score (HS).}
        Correlations (Pearson $\rho_P$ and Spearman $\rho_S$) use the full SS-TS (not the subset used for human study in Table~\ref{tab:humancorrelations}) via four SR models (12K images).
        Columns: affinity or metric functions.
        With respect to GPT-HS,
        we see that 
        (i) existing models do not correlate strongly,
        and
        (ii) our proxies correlate best (and therefore can substitute as optimization objectives), 
        but are also not identical. 
        For this reason, we consider HS evaluation via multiple proxies in \S\ref{sec:results}. See Supp.~\S\ref{sec:add-heatmap-individual} for more details. 
    }
    \begin{adjustbox}{max width=0.99\linewidth}
    \begin{tabular}{lcccccccccccccccccc} %
    \toprule
    \multirow{2}{*}{} & \multirow{2}{*}{PSNR} & \multirow{2}{*}{SSIM} & \multirow{2}{*}{DISTS} & \multirow{2}{*}{LPIPS} & \multirow{2}{*}{MUSIQ} & \multirow{2}{*}{Sharpness} & \multirow{2}{*}{SSD} & \multirow{2}{*}{DeepViT} & \multirow{2}{*}{TLR} & \multicolumn{3}{c}{DINO} & \multicolumn{3}{c}{CLIP} & 
    \multicolumn{3}{c}{~HS Proxies (via GPT-HS)~} 
    \\
    \cmidrule(ll){11-13} \cmidrule(ll){14-16} \cmidrule(ll){17-19}
     & & & & & & & & &  & ST  & CLS & interm & ST & CLS & interm &  CNN & DINO-HS & Qwen-HS \\
     \midrule
    $\rho_P$ &
        0.27 & 0.23 & 0.03 & 0.16 & -0.23 & -0.14 & 0.08 & 0.30 & 0.28 & 0.35 & 0.28 & 0.29 & 0.35 & 0.33 & 0.36 & 0.48 & 0.64 & 0.60
    \\
     $\rho_S$ & 0.25 & 0.22 & 0.02 & 0.17 & -0.23 & -0.22 & 0.14 & 0.30 & 0.27 & {0.33} & 0.26 & 0.32 & {0.33} & 0.31 & {0.35} & 0.48 & 0.63 & 0.60 \\

    \bottomrule
    \end{tabular}
    \end{adjustbox}
    
    \label{tab:correlations}
\end{table*}
}

\noindent\textbf{Dataset.}
We utilize the StableSR Test Set (SS-TS) \cite{wang2024exploiting}, derived from DIV-2K Val \cite{div2k} with RealESRGAN degradations \cite{wang2021real}. 
It consists of 3K crops from 92 images. 

\noindent\textbf{Comparison to Human Judgments.}
We conduct a user study on a subset of the SS-TS (one random crop per image), where 11 users rated the hallucinations in the outputs of three GSR models (PASD~\cite{yang2024pasd}, SeeSR~\cite{wu2024seesr}, and StableSR~\cite{wang2024exploiting}; 276 images total). See Supp.~\S\ref{supp:sstshuman} for details.
In Table~\ref{tab:humancorrelations}, we consider the correlations between these human scores and the various metrics, including GPT-HS.

\noindent
$\bullet\,$  
Our Qwen-HS and DINO-HS proxies, both trained with GPT-HS examples, best mirror human judgments.
The former provides an explanation with its score, while the latter is significantly more efficient.
GPT-HS itself also correlates strongly, with the next highest value in all cases except one.
Finally, the feature distances perform well out-of-the-box, particularly DINO, motivating our DINO-HS architecture.

\noindent
$\bullet\,$
Human perceptual IQA includes inherent variance.
Regarding inter-rater agreement, the mean pairwise Spearman correlation between users is
0.54. %
Thus, on average, \textit{the correlation between humans is 
on par with %
the correlation between GPT-HS and the human mean}, suggesting GPT-HS is a good proxy for human judgment, 
with significant discrepancies attributable to task-inherent variability.

\noindent
$\bullet\,$
Further, we found the \textit{per-image} standard deviations (SDs), across human user scores, to be 0.80 on average, with 85.1\% of images having SD $\leq 1$. 
Similarly, GPT-HS and the human average score have a mean absolute difference of 0.92.
In other words, both the individual raters and GPT-HS generally stay within one point of the human mean.

\noindent
$\bullet\,$ 
Since the discrete GPT-HSs are comparable to human scores, we can measure accuracy: GPT-HS exactly equals the human majority on 29.0\% of samples and is within one point 79.7\% of the time (for human mean, 61.2\%).
Human cross-rater accuracy is similar:
users give identical scores for 34.1\%, and are within one point for 79.2\%, of ratings.

Together, these results suggest that 
GPT-HS and its proxies could be useful surrogates for human notions of hallucination. %
See Supp.~\ref{supp:human} for additional details and visualizations.

\noindent
\textbf{Hallucination Insensitivity of Existing Methods.}
Given that GPT-HS is an appropriate measure of hallucinations, we more comprehensively evaluate its relation to existing metrics.
We therefore construct a larger dataset (12K images with HSs), applying four models (the diffusion-based StableSR~\cite{wang2024exploiting}, SeeSR \cite{wu2024seesr}, and PASD \cite{yang2024pasd}, as well as the regression-based Swin2SR \cite{conde2022swin2sr}) to the full SS-TS.

The results are presented in Table~\ref{tab:correlations}. 
Unsurprisingly, low-level metrics (PSNR and SSIM) correlate positively with GPT-HS, as they favour blurrier images, rather than the invented details that form hallucinations \cite{blau2018perception}.
Moreover, the NR-IQA metrics, MUSIQ and Sharpness, correlate \textit{negatively} with GPT-HS, as they only consider SRI quality in isolation, whereas hallucinations are often superficially realistic.
In contrast, the semantics-aware neural distances correlate strongly to GPT-HS, particularly DINO (known to exhibit low-level human visual traits \cite{cai2025computer}), motivating its use as the basis of our differentiable proxy.
Finally, our proxies correlate best to GPT-HS (higher than inter-human agreement), but still retain some differences; hence, we report all three in our evaluations. %
See Supp.~\S\ref{sec:additional_stats} and~\S\ref{sec:add-heatmap-individual} for details.

\noindent
\textbf{Further Analyses.}
In Supp.~\S\ref{supp:nrhs}, we examine \textit{no reference} (NR) HS estimation (i.e., without using the GT image). %
While this setting shows reduced correlation to human judgment, 
the relatively small gap suggests that NR-HS may be promising for future work. 
Further, in Supp.~\S\ref{supp:promptrobustness}, we demonstrate the robustness of GPT-HS with respect to prompt wording.

\begin{figure}[htb]
    \centering
    \includegraphics[width=0.99\linewidth]{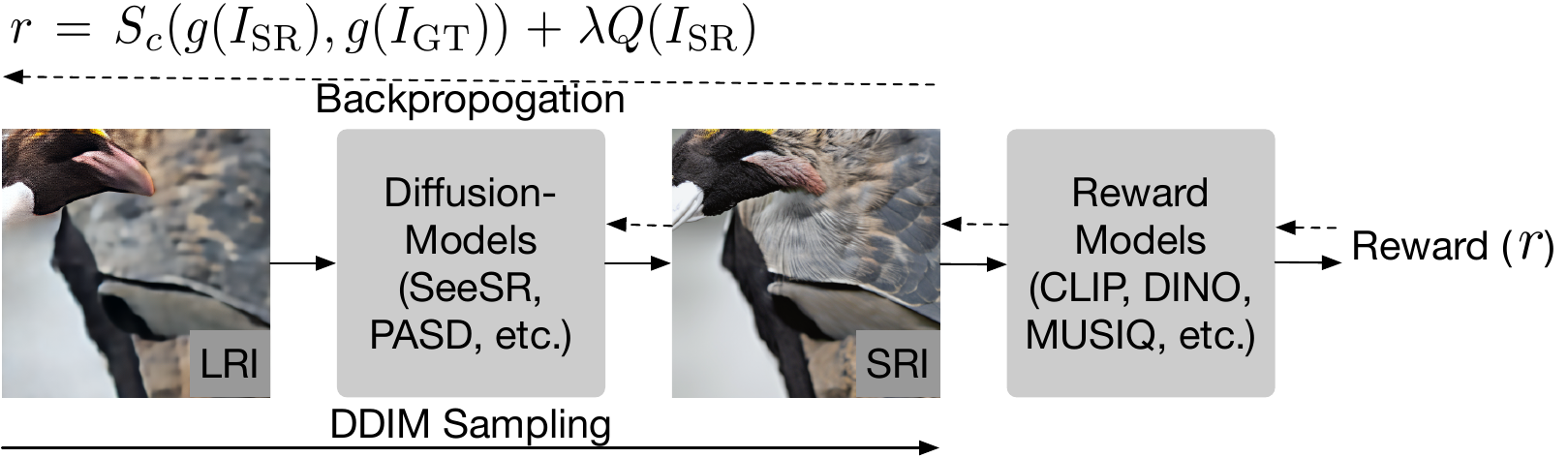}
    \vspace{-0.5em}
    \caption{\textbf{Fine-tuning GSR models to mitigate hallucinations}. We construct a semantics-based differentiable proxy for HS as a reward model, which is then back-propagated through the denoising steps~\cite{prabhudesai2023aligning,clark2023directly} to align GSR models. 
    }
    \label{fig:injection}
    \vspace{-0.7em}
\end{figure}

\section{Mitigating Hallucinations in GSR}
\label{sec:results}

Our analyses in \S\ref{sec:analysis} demonstrate that 
our HS %
is an effective surrogate for measuring hallucinations.
We therefore apply our differentiable HS proxy
as a reward function to fine-tune diffusion-based GSR methods via AlignProp \cite{prabhudesai2023aligning}. 
Empirically, this algorithm reduces hallucinations %
while preserving or even improving other evaluation metrics.

\newcommand{\dinohs}{DINO-HS}

{
\setlength{\tabcolsep}{0.04in}
\fontsize{9}{10}\selectfont
\aboverulesep = 0mm
\belowrulesep = 0mm %
\begin{table*}[t]
    \centering
    \caption{
    \textbf{SR Results.}
        We divide results into standard models (upper parts) and our adapted models trained using reward backpropagation \cite{prabhudesai2023aligning} with \textit{+DINO-HS+MUSIQ} (lower parts).
        Not only do our fine-tuned models obtain improved HS score, but they do so without blurring the image, as measured by our superior results according to the various NR-IQA metrics.
        Further, while our model versions do tend to have lower \textit{pixel-level} fidelity, they actually have better \textit{perceptual} fidelity (LPIPS and DISTS) in most cases.
    }
    \begin{adjustbox}{max width=\linewidth}
    \begin{tabular}{ll|ccccccccccc}
    \toprule
      & Model & PSNR $\uparrow$ & SSIM $\uparrow$ & LPIPS $\downarrow$ & DISTS $\downarrow$ & MUSIQ $\uparrow$ & CLIPIQA $\uparrow$ & QAlign $\uparrow$ &  Sharpness $\uparrow$ & GPT-HS $\uparrow$ & Qwen-HS $\uparrow$ & \dinohs $\uparrow$ \\\hline
      \multirow{10}{*}{SS-TS} 
      & Bicubic & 25.04 & 0.634 & 0.704 & 0.337 & 19.86 & 0.312 & 1.15 & 0.90 & \textbf{4.67} & \textbf{3.30} & \textbf{3.67} \\
      & Swin2SR \cite{conde2022swin2sr} & \textbf{25.75} & \textbf{0.681} & 0.473 & 0.295 & 44.37 & 0.299 & 2.20 & 6.57 & 3.38 & 3.17 & 3.39 \\
      & RealESRGAN \cite{wang2021real} & 24.04 & 0.631 & 0.313 & 0.212 & 62.22 & 0.547 & 3.35 & 73.02 & 2.78 & 2.84 & 2.86 \\
      & StableSR \cite{wang2024exploiting} & 23.26 & 0.573 & {0.311} & {0.205} & {65.92} & {0.677} & {3.53} & \textbf{105.01} & 3.36 & 3.00 & 3.33\\
      & PiSA \cite{sun2025pixel} & 23.87 & 0.606 & \textbf{0.282} & \textbf{0.193} & \textbf{69.68} & \textbf{0.693} & \textbf{3.88} & 73.29 & 3.58 & 3.23 & 3.60\\
      \cdashline{2-13}\noalign{\vskip 0.5ex}
      & SeeSR \cite{wu2024seesr} & \textbf{23.68} & \textbf{0.604} & 0.319 & 0.197 & 68.67 & 0.694 &3.98 & 84.01 & 2.99 & 2.77 &  3.17  \\
      & \quad+DINO-HS+MUSIQ &  23.23 & 0.595 & \textbf{0.252} & \textbf{0.185} & \textbf{70.49} & \textbf{0.743} & \textbf{3.98} & \textbf{135.99} & \textbf{3.87} & \textbf{3.46} & \textbf{3.99} \\
      
      \cdashline{2-13}\noalign{\vskip 0.5ex}
      & PASD \cite{yang2024pasd} & \textbf{23.55} & \textbf{0.598} & 0.369 & 0.214 & 65.54 & 0.635 & 3.75 & 82.59 & 2.54 & 2.42 & 2.48 \\
      & \quad+DINO-HS+MUSIQ & 22.69 & 0.579 & \textbf{0.262} & \textbf{0.186} & \textbf{69.52} & \textbf{0.746} & \textbf{3.84} & \textbf{175.71} & \textbf{3.83}  & \textbf{3.36} & \textbf{3.90} \\
    \hline\noalign{\vskip 0.5ex}

      \multirow{10}{*}{RealSR}
        & Bicubic & 27.11 & 0.756 & 0.456 & 0.263 & 25.81 & 0.310 & 1.66 & 0.95 & \textbf{4.56} & \textbf{3.63} & \textbf{3.98} \\
        & Swin2SR \cite{conde2022swin2sr} & \textbf{27.29} & \textbf{0.801} & 0.291 & 0.237 & 53.14 & 0.303 & 2.51 & 13.26 & 3.57 & 3.13 & 3.46 \\
        & RealESRGAN \cite{wang2021real} & 25.58 & 0.759 & 0.272 & 0.207 & 60.61 & 0.450 & 3.11 & 48.99 & 2.96 & 2.69 & 2.96 \\
        & StableSR \cite{wang2024exploiting} & 24.65 & 0.708 & 0.300 & 0.214 & 65.88 & 0.623 & 3.28 & \textbf{75.74} & 3.22 & 2.68 & 3.31 \\
        & PiSA \cite{sun2025pixel} & 25.50 & 0.742 & \textbf{0.267} & \textbf{0.204} & \textbf{70.14} & \textbf{0.669} & \textbf{3.63} & 51.53 & 3.11 & 2.92 & 3.47 \\
        \cdashline{2-13}\noalign{\vskip 0.5ex}
        
        & SeeSR \cite{wu2024seesr} & \textbf{25.15} & \textbf{0.721} & 0.301 & 0.223 & 69.81 & 0.670 & \textbf{3.72} & 86.99 & 2.92 & 2.60 & 3.13 \\
        & \quad+DINO-HS+MUSIQ & 23.98 & 0.718 & \textbf{0.278} & \textbf{0.200} & \textbf{70.13} & \textbf{0.729} & 3.68 & \textbf{106.23} & \textbf{3.45} & \textbf{3.10} & \textbf{3.88} \\
        \cdashline{2-13}\noalign{\vskip 0.5ex}
        
        & PASD \cite{yang2024pasd} & \textbf{25.75} & \textbf{0.735} & 0.296 & 0.213 & 62.52 & 0.534 & 3.30 & 43.47 & 2.89 & 2.52 & 2.81\\
        & \quad+DINO-HS+MUSIQ & 23.62 & 0.716 & \textbf{0.269} & \textbf{0.197} & \textbf{69.47} & \textbf{0.719} & \textbf{3.59} & \textbf{104.88} & \textbf{3.62} & \textbf{2.99} & \textbf{3.71}
        \\
     \hline\noalign{\vskip 0.5ex}
     \multirow{10}{*}{DRealSR} 
        & Bicubic & \textbf{30.54} & 0.830 & 0.461 & 0.279 & 22.59 & 0.319 & 1.47 & 0.38 & \textbf{4.76} & \textbf{3.95} & \textbf{4.14} \\
        & Swin2SR \cite{conde2022swin2sr} & 29.98 & \textbf{0.843} & 0.330 & 0.251 & 43.58 & 0.325 & 2.23 & 4.07 & 3.68 & 3.69 & 3.63 \\
        & RealESRGAN \cite{wang2021real} & 28.40 & 0.801 & \textbf{0.286} & \textbf{0.211} & 54.87 & 0.454 & 2.91 & 27.07 &3.27 & 3.41 & 3.23 \\
        & StableSR \cite{wang2024exploiting} & 28.03 & 0.754 & 0.328 & 0.227 & 58.51 & 0.636 & 3.06 & \textbf{40.08} & 3.51 & 3.41 & 3.45 \\
        & PiSA \cite{sun2025pixel} & 28.31 & 0.780 & 0.296 & 0.217 & \textbf{66.10} & \textbf{0.697} & \textbf{3.58} & 30.66 & 3.62 & 3.60 & 3.59 \\
        \cdashline{2-13}\noalign{\vskip 0.5ex}
        
        & SeeSR \cite{wu2024seesr} & \textbf{28.07} & \textbf{0.768} & \textbf{0.317} & 0.232 & 65.09 & 0.691 & \textbf{3.59} & 48.21 & 3.11 & 3.14 & 3.15 \\
        & \quad+DINO-HS+MUSIQ & 26.52 & 0.739 & 0.326 & \textbf{0.221} & \textbf{65.19} & \textbf{0.742} & 3.52 & \textbf{55.36} & \textbf{3.80} & \textbf{3.65} & \textbf{3.86}
        \\
        \cdashline{2-13}\noalign{\vskip 0.5ex}
        
        & PASD \cite{yang2024pasd} & \textbf{28.05} & \textbf{0.779} & \textbf{ 0.319} & 0.230 & 58.48 & 0.572 & 3.27 & 29.66 & 2.72 & 2.85 & 2.70 \\
        & \quad+DINO-HS+MUSIQ & 25.10 & 0.719 & 0.328 & \textbf{0.227} & \textbf{65.04} & \textbf{0.729} & \textbf{3.41} & \textbf{58.42} & \textbf{3.74} & \textbf{3.57} & \textbf{3.75}
        \\
      \bottomrule   
    \end{tabular}
    \end{adjustbox}
    \vspace{-0.8em}

    \label{tab:results_injection}
\end{table*}
}

\vspace{0.3em}
\noindent\textbf{Method.}
For HS-based optimization,
we focus on SeeSR \cite{wu2024seesr} and PASD \cite{yang2024pasd}, 
which are representative semantics-aware diffusion models, based on common GSR architectures (ControlNet \cite{zhang2023adding} and UNet \cite{ronneberger2015u};
\eg, \cite{wu2024seesr,yang2024pasd,lin2024tasr,lin2024diffbir}).
Further, despite impressive visual quality, they had more prevalent hallucinations (lower HSs) than others.

We visualize the architecture in Fig.~\ref{fig:injection}. Our method leverages gradient-based reward fine-tuning methods, developed to align 
generative models %
to human preferences \cite{prabhudesai2023aligning, clark2023directly}. 
In our case, we extend AlignProp \cite{prabhudesai2023aligning} to diffusion-based GSR, keeping the same design choices, except for the additional ControlNet, which is kept unchanged. 
To reduce hallucinations, we utilize our HS proxy, DINO-HS, as a differentiable reward model.
We then fine-tune the GSR model to maximize this reward, via backpropagation through the denoising steps.
To avoid excessively disrupting the diffusion prior, 
we train only LoRA weights (rank $4$),
as in AlignProp \cite{prabhudesai2023aligning}.

More specifically, our reward model consists of two terms:
$ r = S_c(g(I_\mathrm{SR}),g(I_\mathrm{GT}))
        + \lambda Q(I_\mathrm{SR}),$
where $g$ is a neural feature extractor,
$S_c$ is cosine similarity, and 
$Q$ is an NRIQA model, MUSIQ \cite{ke2021musiq}, which prevents the GSR method from decreasing perceptual quality (\eg, blur) to increase HS as a trivial solution
(see \textit{Ablations} below). %
For $g$, we focus on our HS proxy, DINO-HS (\S\ref{subsec:hs:proxy}), trained on GPT-HS scores 
(denoted \textit{+DINO-HS+MUSIQ}).
See Supp.~\S\ref{sec:add-results} for more details, as well as additional results, including various configurations of DINO, CLIP, LPIPS, and MSE.

\begin{table*}[t]
\centering
\caption{
    \textbf{Ablations and Variations}.
    Via SeeSR (col.~2), we consider several variations on our DINO-based approach (cols.~3-8), as well as alternative objective terms to DINO (cols.~9-11).
    Note columns 2 and 6 appear in Table \ref{tab:results_injection}.
    By default,
    $\lambda=0.05$ for the DINO-based models. Due to differing scales, 
    MSE and LPIPS use $\lambda=0.001$ and $\lambda = 0.2$. 
    We see that
    (i) fine-tuning greatly improves HS,
        particularly with DINO,
    (ii) the MUSIQ term is useful for maintaining NR quality, and 
    (iii) while DINO-HS greatly improves DINO's human correlation, it only modestly improves it as a reward function (\ie, much of the benefit is from DINO itself, which we originally identified via our correlation studies in \S\ref{sec:correlationan}),
    and
    (iv)
    other objectives cannot reduce HS as effectively as DINO.
    See also Supp.~\ref{sec:add-results} 
    for additional results.
}
\vspace{-0.7em}
\label{tab:ablation-combined}
\resizebox{0.99\linewidth}{!}{%
\begin{tabular}{lcccccccccccc}  %
\toprule
\multirow{2}{*}{Metric} & \multirow{2}{*}{SeeSR} & \multicolumn{2}{c}{+ DINO-HS} & \multicolumn{3}{c}{+ DINO-HS interm + $\lambda \cdot $MUSIQ}  & \multirow{2}{*}{\shortstack[c]{+ DINO-interm\\ ~ + MUSIQ }} & \multicolumn{3}{c}{Other Objectives} 
\\ \cmidrule(l){3-4} \cmidrule(l){5-7} \cmidrule(l){9-11} \cmidrule(l){12-13}  
 &  & last & interm & $\lambda\mathord{=}0.1$ & $\lambda\mathord{=}0.05$ & $\lambda\mathord{=}0.01$ & & MSE & MSE+MUSIQ & LPIPS+MUSIQ  \\ 
 \midrule
PSNR $\uparrow$ & 23.68 & 24.49 & 23.58 & 22.81 & 23.23 & 23.66 & 23.02 & 25.94 & 26.08 & 23.66 \\
LPIPS $\downarrow$ & 0.319 & 0.434 & 0.256 & 0.272 & 0.252 & 0.254 & 0.255 & 0.453 & 0.446 & 0.248 \\
MUSIQ $\uparrow$ & 68.67 & 31.76 & 59.56 & 72.74 & 70.49 & 63.62 & 70.33 & 44.0 & 50.96 & 71.49 \\
QAlign $\uparrow$ & 3.98 & 1.94 & 3.24 & 4.11 & 3.98 & 3.55 & 3.92 & 2.27 & 2.59 & 3.99
\\
Sharpness $\uparrow$ & 84.01 & 5.87 & 70.57 & 147.22 & 135.99 & 80.55 & 126.65 & 7.70 & 6.55 & 101.13
\\
GPT-HS $\uparrow$ & 2.99 & 4.26 & 3.98 & 3.61 & 3.87 & 3.92 & 3.85 & 3.65 & 3.38 & 3.32
 \\ 
Qwen-HS $\uparrow$ & 2.77 & 3.93 & 3.58 & 3.26 & 3.46 & 3.55  & 3.45 & 3.49 & 3.24 & 2.96
\\
\dinohs $\uparrow$ & 3.17 & 4.10 & 4.03 & 3.82 & 3.99 & 3.97 & 3.95 & 3.66 & 3.43 & 3.57
\\
 \bottomrule
\end{tabular}%
} %
\vspace{-0.7em}
\end{table*}

\noindent\textbf{Settings.} 
GSR models are initialized from their pretrained checkpoints.
For data, we combine DIV-2K/8K~\cite{div2k,gu2019div8k} and Flickr2K \cite{agustsson2017ntire}, with RealESRGAN \cite{wang2021real} degradations. 
Inference follows the default configurations (DDIM \cite{song2020denoising} for SeeSR; UniPC \cite{zhao2023unipc} for PASD).
See Supp.~\S\ref{sec:add-results} for details.

\noindent\textbf{Evaluation.} 
Our task is 4$\times$ image super-resolution, which we evaluate on both synthetic and real-world datasets. For synthetic, we use the StableSR \cite{wang2024exploiting} test set (SS-TS; see \S\ref{sec:correlationan}), which has
3K DIV2K-Val crops using RealESRGAN \cite{wang2021real} degradations.
For real-world, we use RealSR \cite{cai2019toward} and DRealSR \cite{wei2020component}. 
We employ an array of reference-based and non-reference-based metrics. For FR-IQA, we apply pixel-level metrics (PSNR and SSIM \cite{wang2004image}) and perceptual metrics (LPIPS \cite{kettunen2019lpips} and DISTS \cite{ding2020image}). 
For NR-IQA, we employ MUSIQ \cite{ke2021musiq}, CLIPIQA \cite{wang2023exploring}, QAlign \cite{wu2024qalign}, and sharpness.

\noindent\textbf{Results.}
We aggregate our results in Table \ref{tab:results_injection}.
We compare to bicubic upsampling (Bicubic) and Swin2SR, along with four diffusion-based GSR models (StableSR, SeeSR, PASD, and PiSA), which span the perception-distortion trade-off \cite{blau2018perception}.
In particular, Bicubic and the non-diffusion Swin2SR perform very well on low-level metrics (PSNR, SSIM), but quite poorly according to NR-IQA metrics. 
In addition, our HS consistently scores Bicubic and Swin2SR the highest, as they output blurry, rather than hallucinatory, content when confronted by uncertainty in the LRI.

\begin{figure}
    \centering
    \includegraphics[width=0.99\linewidth]{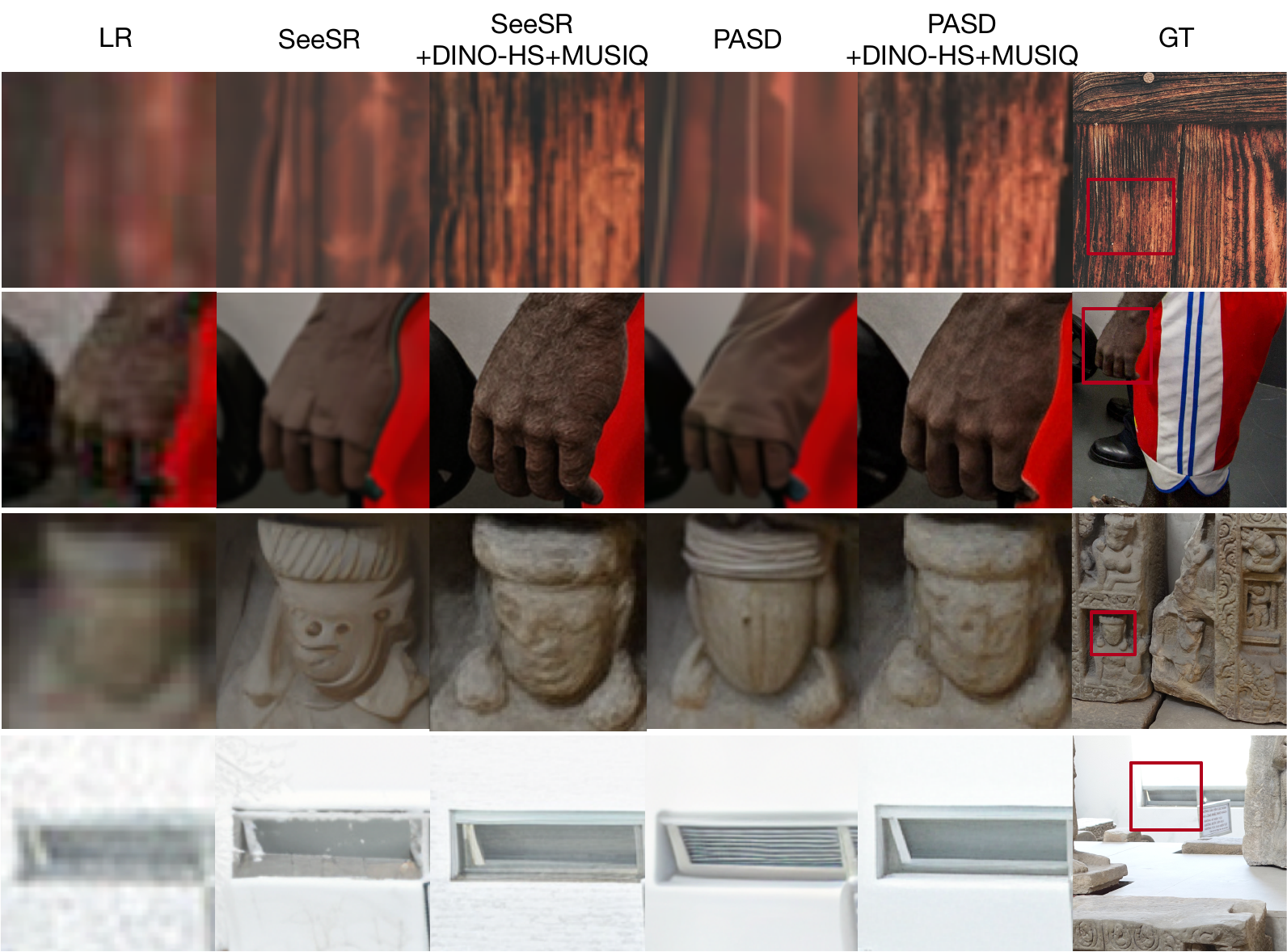}
    \vspace{-0.5em}
    \caption{
        \textbf{Qualitative results.}
        We compare SeeSR and PASD with their aligned variants, SeeSR / PASD + DINO-HS-interm+MUSIQ. We see our models preserve the semantics of the scene better while also generating sharp details (\eg, our model corrected the false ``clothed'' hand).
        See also Supp.~\S\ref{sec:add-results}
        for additional visualizations.
    }
    \vspace{-0.7em}
    \label{fig:visinjection}
\end{figure}

Our primary comparison, however, is between 
the base GSR models (SeeSR and PASD) and our fine-tuned versions, via DINO-HS.
We see that reward-based optimization
greatly reduces hallucinations (as measured by our HS functions),
but without sacrificing other metrics.
Indeed, our adapted model is generally \textit{improved} in terms of realism and quality, according to NR-IQA measures, suggesting the high HS is \textit{not} due to blurry outputs (as for Swin2SR and Bicubic).
Further, though our aligned models do incur reduced low-level (pixel-space) fidelity (PSNR and SSIM), they improve perceptual fidelity (LPIPS and DISTS) in most cases.
Overall, our approach improves hallucinations, while achieving comparable, and even improving, perceptual quality.
For visual comparison, we show sample outputs in
Fig.~\ref{fig:visinjection}.

\vspace{0.3em}
\noindent\textbf{Ablations.}
We present several variations of our approach on SeeSR
in Table~\ref{tab:ablation-combined}.
(i) \textit{Last vs.\ intermediate layers}: 
our DINO-HS proxy utilizes the last layer outputs for HS estimation (see \S\ref{subsec:hs:proxy} and Supp.~\S\ref{supp:proxy:model}), obtaining high correlation to human scores (\S\ref{sec:correlationan}).
However, for AlignProp, using this directly as a reward is devastating to image quality; instead, including intermediate layers produces much better perceptual fidelity (LPIPS) and quality (MUSIQ), while still improving HS. 
(ii) \textit{MUSIQ factors ($\lambda$)}: unsurprisingly, we observe higher $\lambda$ leads to higher perceived quality (MUSIQ), but lower fidelity and HS. %
Our choice of optimal $\lambda$ (=$0.05$) is driven by 
(a) not going below the quality of the base variant
(e.g., MUSIQ and QAlign)
but also
(b) attaining the best HS and perceptual fidelity (LPIPS) possible.
(iii) \textit{Proxy training:}
    while our HS reward
    (DINO-HS; col.~6) improves over the \textit{un}tuned DINO (col.~8), the changes are modest, suggesting DINO itself 
    is more fundamental to our performance than 
    HS-based tuning 
    (which may be unsurprising, 
    given DINO was identified by its HS correlation).
    However, note that proxy tuning remains essential for obtaining high correlation to humans (\S\ref{sec:correlationan}).
(iv) \textit{Alternative objectives:} compared to DINO, using other rewards (MSE or LPIPS) does not improve HS as effectively. 
See also Supp.\S\ref{sec:add-results} and \S\ref{supp:extraremarks} for additional ablations and variations of our approach.

\section{Conclusion}
\label{sec:conc}
\vspace{-0.2em}
We have considered the problem of hallucinations in GSR, including its definition, its measurement via HS, 
its relation to existing metrics, and a carefully designed approach to ameliorating it.
While our HS (a) closely matches human judgments, and (b) is complementary to existing metrics, it is computed via an MLLM, which is both expensive and difficult to optimize through. 
Building on DINO, we construct a differentiable proxy for HS, and leverage it as a reward function for GSR fine-tuning,
mitigating hallucinations while preserving, or even improving, other metrics.
We believe future work, such as localizing hallucinated regions in SRI, will bring GSR closer to practical use.

{
    \small
    \bibliographystyle{ieeenat_fullname}
    \bibliography{main}

@String { AAAI         = {{Proceedings of the National Conference on Artificial Intelligence ({AAAI})}} }

@String { ACCV         = {{Proceedings of the Asian Conference on Computer Vision ({ACCV})}} }

@String { CVIU         = {{Computer Vision and Image Understanding ({CVIU})}} }

@String { CVPR         = {{Proceedings of the {IEEE} Conference on Computer Vision and Pattern Recognition ({CVPR})}} }

@String { CVPRW        = {{Proceedings of the {IEEE} Conference on Computer Vision and Pattern Recognition Workshops ({CVPRW})}} }

@String { ECCV         = {{Proceedings of the European Conference on Computer Vision ({ECCV})}} }

@String { ICCV         = {{Proceedings of the International Conference on Computer Vision ({ICCV})}} }

@String { ICCVW         = {{Proceedings of the International Conference on Computer Vision Workshops ({ICCVW})}} }

@String { ICLR         = {{Proceedings of the International Conference on Learning Representations ({ICLR})}} }

@String { ICML         = {{Proceedings of the International Conference on Machine Learning ({ICML})}} }

@String { IJCV         = {{International Journal of Computer Vision ({IJCV})}} }

@String { MICCAI       = {{Proceedings of the International Conference on Medical Image Computing and Computer Assisted Intervention ({MICCAI})}} }

@String { NIPS         = {{Neural Information Processing Systems ({NeurIPS})}} }

@String { NeuRIPS      = {{Neural Information Processing Systems ({NeurIPS})}} }

@String { PAMI         = {{{IEEE} Transactions on Pattern Analysis and Machine Intelligence ({PAMI})}} }

@String { TIP          = {{{IEEE} Transactions on Image Processing}} }

@String { WACV         = {{Proceedings of the {IEEE} Workshop on Applications of Computer Vision ({WACV})}} }

@String { CVPR         = {{{IEEE} Conference on Computer Vision and Pattern Recognition ({CVPR})}} }

@String { CVPRW        = {{{IEEE} Conference on Computer Vision and Pattern Recognition Workshops ({CVPRW})}} }

@String { ECCV         = {{European Conference on Computer Vision ({ECCV})}} }

@String { ECCVW        = {{European Conference on Computer Vision Workshops ({ECCVW})}} }

@String { ICCV         = {{International Conference on Computer Vision ({ICCV})}} }

@String { ICCVW        = {{International Conference on Computer Vision Workshops ({ICCVW})}} }

@String { ICLR         = {{International Conference on Learning Representations ({ICLR})}} }

@String { ICML         = {{International Conference on Machine Learning ({ICML})}} }

@String { TMLR         = {{Transactions on Machine Learning Research {(TMLR)}} }}

@String { WACV         = {{Winter Conference on Applications of Computer Vision ({WACV})}} }

@inproceedings{he2016deep,
  title={Deep residual learning for image recognition},
  author={He, Kaiming and Zhang, Xiangyu and Ren, Shaoqing and Sun, Jian},
  booktitle=CVPR,
  year={2016}
}

@article{cohen1960coefficient,
  title={A coefficient of agreement for nominal scales},
  author={Cohen, Jacob},
  journal={Educational and psychological measurement},
  volume={20},
  number={1},
  pages={37--46},
  year={1960},
  publisher={Sage Publications Sage CA: Thousand Oaks, CA}
}

@article{scikitlearn,
  title={Scikit-learn: Machine Learning in {P}ython},
  author={Pedregosa, F. and Varoquaux, G. and Gramfort, A. and Michel, V.
          and Thirion, B. and Grisel, O. and Blondel, M. and Prettenhofer, P.
          and Weiss, R. and Dubourg, V. and Vanderplas, J. and Passos, A. and
          Cournapeau, D. and Brucher, M. and Perrot, M. and Duchesnay, E.},
  journal={Journal of Machine Learning Research},
  volume={12},
  pages={2825--2830},
  year={2011}
}

@inproceedings{ronneberger2015u,
  title={U-net: Convolutional networks for biomedical image segmentation},
  author={Ronneberger, Olaf and Fischer, Philipp and Brox, Thomas},
  booktitle=MICCAI,
  year={2015},
}

@article{song2020denoising,
  title={Denoising diffusion implicit models},
  author={Song, Jiaming and Meng, Chenlin and Ermon, Stefano},
  journal=ICLR,
  year={2021}
}

@inproceedings{johnson2016perceptual,
  title={Perceptual losses for real-time style transfer and super-resolution},
  author={Johnson, Justin and Alahi, Alexandre and Fei-Fei, Li},
  booktitle=ECCV,
  year={2016},
}

@article{ding2021comparison,
  title={Comparison of full-reference image quality models for optimization of image processing systems},
  author={Ding, Keyan and Ma, Kede and Wang, Shiqi and Simoncelli, Eero P},
  journal=IJCV,
  year={2021},
}

@inproceedings{blau2018perception,
  title={The perception-distortion tradeoff},
  author={Blau, Yochai and Michaeli, Tomer},
  booktitle=CVPR,
  year={2018}
}

@inproceedings{blau2019rethinking,
  title={Rethinking lossy compression: The rate-distortion-perception tradeoff},
  author={Blau, Yochai and Michaeli, Tomer},
  booktitle=ICML,
  year={2019},
}

@inproceedings{zhang2025augmenting,
  title={Augmenting Perceptual Super-Resolution via Image Quality Predictors},
  author={Zhang, Fengjia and Rangrej, Samrudhdhi B and Aumentado-Armstrong, Tristan and Fazly, Afsaneh and Levinshtein, Alex},
  booktitle=CVPR,
  year={2025}
}

@inproceedings{fu2023dreamsim,
title={{DreamSim}: Learning New Dimensions of Human Visual Similarity using Synthetic Data},
author= {Fu, Stephanie and Tamir, Netanel and Sundaram, Shobhita and Chai, Lucy and Zhang, Richard and Dekel, Tali and Isola, Phillip},
booktitle=NEURIPS,
year={2023}
}

@inproceedings{brooks2023instructpix2pix,
  title={{InstructPix2Pix}: Learning to follow image editing instructions},
  author={Brooks, Tim and Holynski, Aleksander and Efros, Alexei A},
  booktitle=CVPR,
  year={2023}
}

@inproceedings{mirzaei2024watch,
  title={Watch your steps: Local image and scene editing by text instructions},
  author={Mirzaei, Ashkan and Aumentado-Armstrong, Tristan and Brubaker, Marcus A and Kelly, Jonathan and Levinshtein, Alex and Derpanis, Konstantinos G and Gilitschenski, Igor},
  booktitle=ECCV,
  year={2024},
}

@article{mittal2012making,
  title={Making a ``completely blind'' image quality analyzer},
  author={Mittal, Anish and Soundararajan, Rajiv and Bovik, Alan C},
  journal={IEEE Signal processing letters},
  volume={20},
  number={3},
  pages={209--212},
  year={2012},
  publisher={IEEE}
}

@article{xie2023desra,
  title={{DeSRA}: detect and delete the artifacts of {GAN}-based real-world super-resolution models},
  author={Xie, Liangbin and Wang, Xintao and Chen, Xiangyu and Li, Gen and Shan, Ying and Zhou, Jiantao and Dong, Chao},
  journal=ICML,
  year={2023}
}

@inproceedings{liang2022details,
  title={Details or artifacts: A locally discriminative learning approach to realistic image super-resolution},
  author={Liang, Jie and Zeng, Hui and Zhang, Lei},
  booktitle=CVPR,
  year={2022}
}

@article{you2024descriptive,
  title={Descriptive image quality assessment in the wild},
  author={You, Zhiyuan and Gu, Jinjin and Li, Zheyuan and Cai, Xin and Zhu, Kaiwen and Dong, Chao and Xue, Tianfan},
  journal={arXiv preprint arXiv:2405.18842},
  year={2024}
}

@article{ghazanfari2023r,
  title={{R-LPIPS}: An adversarially robust perceptual similarity metric},
  author={Ghazanfari, Sara and Garg, Siddharth and Krishnamurthy, Prashanth and Khorrami, Farshad and Araujo, Alexandre},
  journal={arXiv preprint arXiv:2307.15157},
  year={2023}
}

@article{kettunen2019lpips,
  title={{E-LPIPS}: robust perceptual image similarity via random transformation ensembles},
  author={Kettunen, Markus and H{\"a}rk{\"o}nen, Erik and Lehtinen, Jaakko},
  journal={arXiv preprint arXiv:1906.03973},
  year={2019}
}

@inproceedings{zhang2018unreasonable,
  title={The unreasonable effectiveness of deep features as a perceptual metric},
  author={Zhang, Richard and Isola, Phillip and Efros, Alexei A and Shechtman, Eli and Wang, Oliver},
  booktitle=CVPR,
  year={2018}
}

@article{delbracio2023inversion,
  title={Inversion by Direct Iteration: An Alternative to Denoising Diffusion for Image Restoration},
  author={Delbracio, Mauricio and Milanfar, Peyman},
  journal=TMLR,
  year={2023}
}

@inproceedings{khrulkov2021neural,
  title={Neural side-by-side: Predicting human preferences for no-reference super-resolution evaluation},
  author={Khrulkov, Valentin and Babenko, Artem},
  booktitle=CVPR,
  year={2021}
}

@article{ding2020image,
  title={Image quality assessment: Unifying structure and texture similarity},
  author={Ding, Keyan and Ma, Kede and Wang, Shiqi and Simoncelli, Eero P},
  journal=PAMI,
  year={2020},
}

@inproceedings{clark2023directly,
  title={Directly fine-tuning diffusion models on differentiable rewards},
  author={Clark, Kevin and Vicol, Paul and Swersky, Kevin and Fleet, David J},
  booktitle=ICLR,
  year={2024}
}

@inproceedings{park2023content,
  title={Content-aware local {GAN} for photo-realistic super-resolution},
  author={Park, JoonKyu and Son, Sanghyun and Lee, Kyoung Mu},
  booktitle=ICCV,
  year={2023}
}

@inproceedings{li2024sed,
  title={{SeD}: Semantic-aware discriminator for image super-resolution},
  author={Li, Bingchen and Li, Xin and Zhu, Hanxin and Jin, Yeying and Feng, Ruoyu and Zhang, Zhizheng and Chen, Zhibo},
  booktitle=CVPR,
  year={2024}
}

@inproceedings{wang2018esrgan,
  title={{ESRGAN}: Enhanced super-resolution generative adversarial networks},
  author={Wang, Xintao and Yu, Ke and Wu, Shixiang and Gu, Jinjin and Liu, Yihao and Dong, Chao and Qiao, Yu and Change Loy, Chen},
  booktitle=ECCVW,
  year={2018}
}

@inproceedings{wang2021real,
  title={Real-{ESRGAN}: Training real-world blind super-resolution with pure synthetic data},
  author={Wang, Xintao and Xie, Liangbin and Dong, Chao and Shan, Ying},
  booktitle=ICCV,
  year={2021}
}

@inproceedings{sun2024pisasr,
  title={Pixel-level and Semantic-level Adjustable Super-resolution: A Dual-{LoRA} Approach},
  author={Sun, Lingchen and Wu, Rongyuan and Ma, Zhiyuan and Liu, Shuaizheng and Yi, Qiaosi and Zhang, Lei},
  booktitle=CVPR,
  year={2025}
}

@inproceedings{
wu2024qalign,
title={{Q-Align}: Teaching {LMM}s for Visual Scoring via Discrete Text-Defined Levels},
author={Wu, Haoning and Zhang, Zicheng and Zhang, Weixia and Chen, Chaofeng and Liao, Liang and Li, Chunyi and Gao, Yixuan and Wang, Annan and Zhang, Erli and Sun, Wenxiu and Yan, Qiong and Min, Xiongkuo and Zhai, Guangtao and Lin, Weisi},
booktitle=ICML,
year={2024},
}

@inproceedings{wu2024one,
  title={One-step effective diffusion network for real-world image super-resolution},
  author={Wu, Rongyuan and Sun, Lingchen and Ma, Zhiyuan and Zhang, Lei},
  booktitle=NeurIPS,
  year={2024}
}

@inproceedings{guo2022lar,
  title={{LAR-SR}: A local autoregressive model for image super-resolution},
  author={Guo, Baisong and Zhang, Xiaoyun and Wu, Haoning and Wang, Yu and Zhang, Ya and Wang, Yan-Feng},
  booktitle=CVPR,
  year={2022}
}

@inproceedings{lim2024evaluating,
  title={Evaluating Image Hallucination in Text-to-Image Generation with Question-Answering}, 
  author={Lim, Youngsun and Choi, Hojun and Shim, Hyunjung},
  booktitle=AAAI,
  year={2025}
}

@InProceedings{div2k,
author = {Agustsson, Eirikur and Timofte, Radu},
title = {{NTIRE} 2017 Challenge on Single Image Super-Resolution: Dataset and Study},
booktitle = CVPRW,
year = {2017}
}

@inproceedings{agnolucci2024arniqa,
  title={{ARNIQA}: Learning distortion manifold for image quality assessment},
  author={Agnolucci, Lorenzo and Galteri, Leonardo and Bertini, Marco and Del Bimbo, Alberto},
  booktitle=WACV,
  year={2024}
}

@article{chen2024topiq,
  title={{TOPIQ}: A top-down approach from semantics to distortions for image quality assessment},
  author={Chen, Chaofeng and Mo, Jiadi and Hou, Jingwen and Wu, Haoning and Liao, Liang and Sun, Wenxiu and Yan, Qiong and Lin, Weisi},
  journal={IEEE Transactions on Image Processing},
  year={2024},
}

@inproceedings{wang2023exploring,
  title={Exploring {CLIP} for assessing the look and feel of images},
  author={Wang, Jianyi and Chan, Kelvin CK and Loy, Chen Change},
  booktitle=AAAI, 
  year={2023}
}

@inproceedings{ke2021musiq,
  title={{MUSIQ}: Multi-scale image quality transformer},
  author={Ke, Junjie and Wang, Qifei and Wang, Yilin and Milanfar, Peyman and Yang, Feng},
  booktitle=ICCV,
  year={2021}
}

@inproceedings{ying2020patches,
  title={From patches to pictures ({PaQ-2-PiQ}): Mapping the perceptual space of picture quality},
  author={Ying, Zhenqiang and Niu, Haoran and Gupta, Praful and Mahajan, Dhruv and Ghadiyaram, Deepti and Bovik, Alan},
  booktitle=CVPR,
  year={2020}
}

@article{ma2017learning,
  title={Learning a no-reference quality metric for single-image super-resolution},
  author={Ma, Chao and Yang, Chih-Yuan and Yang, Xiaokang and Yang, Ming-Hsuan},
  journal=CVIU,
  year={2017},
}

@inproceedings{chen2024low,
  title={Low-res leads the way: Improving generalization for super-resolution by self-supervised learning},
  author={Chen, Haoyu and Li, Wenbo and Gu, Jinjin and Ren, Jingjing and Sun, Haoze and Zou, Xueyi and Zhang, Zhensong and Yan, Youliang and Zhu, Lei},
  booktitle=CVPR,
  year={2024}
}

@inproceedings{bahat2020explorable,
  title={Explorable super resolution},
  author={Bahat, Yuval and Michaeli, Tomer},
  booktitle=CVPR,
  year={2020}
}

@article{wang2004image,
  title={Image quality assessment: from error visibility to structural similarity},
  author={Wang, Zhou and Bovik, Alan C and Sheikh, Hamid R and Simoncelli, Eero P},
  journal=TIP,
  year={2004},
}

@article{wang2024exploiting,
  title={Exploiting diffusion prior for real-world image super-resolution},
  author={Wang, Jianyi and Yue, Zongsheng and Zhou, Shangchen and Chan, Kelvin CK and Loy, Chen Change},
  journal=IJCV,
  year={2024},
}

@inproceedings{wu2024seesr,
  title={{SeeSR}: Towards semantics-aware real-world image super-resolution},
  author={Wu, Rongyuan and Yang, Tao and Sun, Lingchen and Zhang, Zhengqiang and Li, Shuai and Zhang, Lei},
  booktitle=CVPR,
  year={2024}
}

@inproceedings{gao2023implicit,
  title={Implicit diffusion models for continuous super-resolution},
  author={Gao, Sicheng and Liu, Xuhui and Zeng, Bohan and Xu, Sheng and Li, Yanjing and Luo, Xiaoyan and Liu, Jianzhuang and Zhen, Xiantong and Zhang, Baochang},
  booktitle=CVPR,
  year={2023}
}

@inproceedings{yang2024pasd,
  title={Pixel-aware {Stable Diffusion} for realistic image super-resolution and personalized stylization},
  author={Yang, Tao and Wu, Rongyuan and Ren, Peiran and Xie, Xuansong and Zhang, Lei},
  booktitle=ECCV,
  year={2024},
}

@article{moser2024diffusion,
  title={Diffusion models, image super-resolution, and everything: A survey},
  author={Moser, Brian B and Shanbhag, Arundhati S and Raue, Federico and Frolov, Stanislav and Palacio, Sebastian and Dengel, Andreas},
  journal={IEEE Transactions on Neural Networks and Learning Systems},
  year={2024},
  publisher={IEEE}
}

@inproceedings{bruna2016super,
  title={Super-resolution with deep convolutional sufficient statistics},
  author={Bruna, Joan and Sprechmann, Pablo and LeCun, Yann},
  booktitle=ICLR,
  year={2016}
}

@article{schultz1994bayesian,
  title={A {Bayesian} approach to image expansion for improved definition},
  author={Schultz, Richard R and Stevenson, Robert L},
  journal=TIP,
  year={1994},
}

@inproceedings{lugmayr2020srflow,
  title={{SRFlow}: Learning the super-resolution space with normalizing flow},
  author={Lugmayr, Andreas and Danelljan, Martin and Van Gool, Luc and Timofte, Radu},
  booktitle=ECCV,
  year={2020},
}

@inproceedings{heusel2017gans,
  title={{GANs} trained by a two time-scale update rule converge to a local {Nash} equilibrium},
  author={Heusel, Martin and Ramsauer, Hubert and Unterthiner, Thomas and Nessler, Bernhard and Hochreiter, Sepp},
  booktitle=NIPS,
  year={2017}
}

@article{oquab2023dinov2,
  title={{DINOv2}: Learning robust visual features without supervision},
  author={Oquab, Maxime and Darcet, Timoth{\'e}e and Moutakanni, Th{\'e}o and Vo, Huy and Szafraniec, Marc and Khalidov, Vasil and Fernandez, Pierre and Haziza, Daniel and Massa, Francisco and El-Nouby, Alaaeldin and others},
  journal={arXiv preprint arXiv:2304.07193},
  year={2023}
}

@inproceedings{amir2021deep,
  title={Deep {ViT} Features as Dense Visual Descriptors},
  author={Amir, Shir and Gandelsman, Yossi and Bagon, Shai and Dekel, Tali},
  booktitle={European Conference on Computer Vision Workshops (ECCVW)},
  year={2022}
}

@inbook{girod1993mse,
author = {Girod, Bernd},
title = {What's wrong with mean-squared error?},
year = {1993},
publisher = {MIT Press},
booktitle = {Digital Images and Human Vision},
pages = {207–220},
}

@article{mannos1974effects,
  title={The effects of a visual fidelity criterion of the encoding of images},
  author={Mannos, James and Sakrison, David},
  journal={IEEE transactions on Information Theory},
  volume={20},
  number={4},
  pages={525--536},
  year={1974},
}

@inproceedings{radford2021learning,
  title={Learning transferable visual models from natural language supervision},
  author={Radford, Alec and Kim, Jong Wook and Hallacy, Chris and Ramesh, Aditya and Goh, Gabriel and Agarwal, Sandhini and Sastry, Girish and Askell, Amanda and Mishkin, Pamela and Clark, Jack and others},
  booktitle=ICML,
  year={2021},
}

@inproceedings{cherti2023reproducible,
  title={Reproducible scaling laws for contrastive language-image learning},
  author={Cherti, Mehdi and Beaumont, Romain and Wightman, Ross and Wortsman, Mitchell and Ilharco, Gabriel and Gordon, Cade and Schuhmann, Christoph and Schmidt, Ludwig and Jitsev, Jenia},
  booktitle=CVPR,
  year={2023}
}

@inproceedings{yu2024scaling,
  title={Scaling up to excellence: Practicing model scaling for photo-realistic image restoration in the wild},
  author={Yu, Fanghua and Gu, Jinjin and Li, Zheyuan and Hu, Jinfan and Kong, Xiangtao and Wang, Xintao and He, Jingwen and Qiao, Yu and Dong, Chao},
  booktitle=CVPR,
  year={2024}
}

@inproceedings{chen2025faithdiff,
  title={{FaithDiff}: Unleashing diffusion priors for faithful image super-resolution},
  author={Chen, Junyang and Pan, Jinshan and Dong, Jiangxin},
  booktitle=CVPR,
  year={2025}
}

@inproceedings{duan2025dit4sr,
  title={{DiT4SR}: Taming diffusion transformer for real-world image super-resolution},
  author={Duan, Zheng-Peng and Zhang, Jiawei and Jin, Xin and Zhang, Ziheng and Xiong, Zheng and Zou, Dongqing and Ren, Jimmy S and Guo, Chunle and Li, Chongyi},
  booktitle=ICCV,
  year={2025}
}

@inproceedings{zhang2023simple,
  title={A simple framework for open-vocabulary segmentation and detection},
  author={Zhang, Hao and Li, Feng and Zou, Xueyan and Liu, Shilong and Li, Chunyuan and Yang, Jianwei and Zhang, Lei},
  booktitle=ICCV,
  year={2023}
}

@inproceedings{zhang2023recognize,
title={{Recognize Anything}: A Strong Image Tagging Model},
author={Zhang, Youcai and Huang, Xinyu and Ma, Jinyu and Li, Zhaoyang and Luo, Zhaochuan and Xie, Yanchun and Qin, Yuzhuo and Luo, Tong and Li, Yaqian and Liu, Shilong and others},
booktitle={Proceedings of the {IEEE} Conference on Computer Vision and Pattern Recognition Workshops ({CVPRW}).},
year={2024}
}

@article{huang2023open,
  title={Open-set image tagging with multi-grained text supervision},
  author={Huang, Xinyu and Huang, Yi-Jie and Zhang, Youcai and Tian, Weiwei and Feng, Rui and Zhang, Yuejie and Xie, Yanchun and Li, Yaqian and Zhang, Lei},
  journal={arXiv preprint arXiv:2310.15200},
  year={2023}
}

@inproceedings{jo2021tackling,
  title={Tackling the ill-posedness of super-resolution through adaptive target generation},
  author={Jo, Younghyun and Oh, Seoung Wug and Vajda, Peter and Kim, Seon Joo},
  booktitle=CVPR,
  year={2021}
}

@inproceedings{darcet2023vision,
  title={Vision transformers need registers},
  author={Darcet, Timoth{\'e}e and Oquab, Maxime and Mairal, Julien and Bojanowski, Piotr},
  booktitle=ICLR,
  year={2024}
}

@inproceedings{caron2021emerging,
  title={Emerging properties in self-supervised vision transformers},
  author={Caron, Mathilde and Touvron, Hugo and Misra, Ishan and J{\'e}gou, Herv{\'e} and Mairal, Julien and Bojanowski, Piotr and Joulin, Armand},
  booktitle=ICCV,
  year={2021}
}

@inproceedings{rombach2022high,
  title={High-resolution image synthesis with latent diffusion models},
  author={Rombach, Robin and Blattmann, Andreas and Lorenz, Dominik and Esser, Patrick and Ommer, Bj{\"o}rn},
  booktitle=CVPR,
  year={2022}
}

@inproceedings{zhang2024telling,
  title={Telling left from right: identifying geometry-aware semantic correspondence},
  author={Zhang, Junyi and Herrmann, Charles and Hur, Junhwa and Chen, Eric and Jampani, Varun and Sun, Deqing and Yang, Ming-Hsuan},
  booktitle=CVPR,
  year={2024}
}

@inproceedings{yao2023local,
  title={Local implicit normalizing flow for arbitrary-scale image super-resolution},
  author={Yao, Jie-En and Tsao, Li-Yuan and Lo, Yi-Chen and Tseng, Roy and Chang, Chia-Che and Lee, Chun-Yi},
  booktitle=CVPR,
  year={2023}
}

@inproceedings{zhang2023adding,
  title={Adding conditional control to text-to-image diffusion models},
  author={Zhang, Lvmin and Rao, Anyi and Agrawala, Maneesh},
  booktitle=ICCV,
  year={2023}
}

@article{lin2024tasr,
  title={{TASR}: Timestep-Aware Diffusion Model for Image Super-Resolution},
  author={Lin, Qinwei and Sun, Xiaopeng and Gao, Yu and Zhong, Yujie and Li, Dengjie and Zhao, Zheng and Wang, Haoqian},
  journal={arXiv preprint arXiv:2412.03355},
  year={2024}
}

@inproceedings{ledig2017photo,
  title={Photo-realistic single image super-resolution using a generative adversarial network},
  author={Ledig, Christian and Theis, Lucas and Husz{\'a}r, Ferenc and Caballero, Jose and Cunningham, Andrew and Acosta, Alejandro and Aitken, Andrew and Tejani, Alykhan and Totz, Johannes and Wang, Zehan and others},
  booktitle=CVPR,
  year={2017}
}

@article{pan2021exploiting,
  title={Exploiting deep generative prior for versatile image restoration and manipulation},
  author={Pan, Xingang and Zhan, Xiaohang and Dai, Bo and Lin, Dahua and Loy, Chen Change and Luo, Ping},
  journal=PAMI,
  year={2021},
}

@inproceedings{ma2020structure,
  title={Structure-preserving super resolution with gradient guidance},
  author={Ma, Cheng and Rao, Yongming and Cheng, Yean and Chen, Ce and Lu, Jiwen and Zhou, Jie},
  booktitle=CVPR,
  year={2020}
}

@inproceedings{sun2008image,
  title={Image super-resolution using gradient profile prior},
  author={Sun, Jian and Xu, Zongben and Shum, Heung-Yeung},
  booktitle=CVPR,
  year={2008},
}

@inproceedings{cherian2019sem,
  title={{Sem-GAN}: Semantically-consistent image-to-image translation},
  author={Cherian, Anoop and Sullivan, Alan},
  booktitle=WACV,
  year={2019},
}

@inproceedings{myers2020semantic,
  title={Semantic pixel distances for image editing},
  author={Myers-Dean, Josh and Wehrwein, Scott},
  booktitle=CVPRW,
  year={2020}
}

@inproceedings{gallego2019focus,
  title={Focus is all you need: Loss functions for event-based vision},
  author={Gallego, Guillermo and Gehrig, Mathias and Scaramuzza, Davide},
  booktitle=CVPR,
  year={2019}
}

@inproceedings{cohen2025looks,
  title={Looks too good to be true: An information-theoretic analysis of hallucinations in generative restoration models},
  author={Cohen, Regev and Kligvasser, Idan and Rivlin, Ehud and Freedman, Daniel},
  booktitle=NEURIPS,
  year={2025}
}

@inproceedings{fuoli2021fourier,
  title={{Fourier} space losses for efficient perceptual image super-resolution},
  author={Fuoli, Dario and Van Gool, Luc and Timofte, Radu},
  booktitle=ICCV,
  year={2021}
}

@inproceedings{ning2021uncertainty,
  title={Uncertainty-driven loss for single image super-resolution},
  author={Ning, Qian and Dong, Weisheng and Li, Xin and Wu, Jinjian and Shi, Guangming},
  booktitle=NEURIPS,
  year={2021}
}

@inproceedings{mechrez2019maintaining,
  title={Maintaining natural image statistics with the contextual loss},
  author={Mechrez, Roey and Talmi, Itamar and Shama, Firas and Zelnik-Manor, Lihi},
  booktitle=ACCV,
  year={2019},
}

@inproceedings{rad2019srobb,
  title={{SROBB}: Targeted perceptual loss for single image super-resolution},
  author={Rad, Mohammad Saeed and Bozorgtabar, Behzad and Marti, Urs-Viktor and Basler, Max and Ekenel, Hazim Kemal and Thiran, Jean-Philippe},
  booktitle=ICCV,
  year={2019}
}

@inproceedings{liu2023spectral,
  title={Spectral {Bayesian} uncertainty for image super-resolution},
  author={Liu, Tao and Cheng, Jun and Tan, Shan},
  booktitle=CVPR,
  year={2023}
}

@inproceedings{wang2023spatial,
  title={Spatial-frequency mutual learning for face super-resolution},
  author={Wang, Chenyang and Jiang, Junjun and Zhong, Zhiwei and Liu, Xianming},
  booktitle=CVPR,
  year={2023}
}

@inproceedings{aithal2025understanding,
  title={Understanding hallucinations in diffusion models through mode interpolation},
  author={Aithal, Sumukh K and Maini, Pratyush and Lipton, Zachary and Kolter, J Zico},
  booktitle=NEURIPS,
  year={2025}
}

@inproceedings{conde2022swin2sr,
  title={{Swin2SR}: {SwinV2} transformer for compressed image super-resolution and restoration},
  author={Conde, Marcos V and Choi, Ui-Jin and Burchi, Maxime and Timofte, Radu},
  booktitle=ECCVW,
  year={2022},
}

@article{bai2025qwen2,
  title={Qwen2.5-{VL} technical report},
  author={Bai, Shuai and Chen, Keqin and Liu, Xuejing and Wang, Jialin and Ge, Wenbin and Song, Sibo and Dang, Kai and Wang, Peng and Wang, Shijie and Tang, Jun and others},
  journal={arXiv preprint arXiv:2502.13923},
  year={2025}
}

@article{qwen,
  title={Qwen Technical Report},
  author={Jinze Bai and Shuai Bai and Yunfei Chu and Zeyu Cui and Kai Dang and Xiaodong Deng and Yang Fan and Wenbin Ge and Yu Han and Fei Huang and Binyuan Hui and Luo Ji and Mei Li and Junyang Lin and Runji Lin and Dayiheng Liu and Gao Liu and Chengqiang Lu and Keming Lu and Jianxin Ma and Rui Men and Xingzhang Ren and Xuancheng Ren and Chuanqi Tan and Sinan Tan and Jianhong Tu and Peng Wang and Shijie Wang and Wei Wang and Shengguang Wu and Benfeng Xu and Jin Xu and An Yang and Hao Yang and Jian Yang and Shusheng Yang and Yang Yao and Bowen Yu and Hongyi Yuan and Zheng Yuan and Jianwei Zhang and Xingxuan Zhang and Yichang Zhang and Zhenru Zhang and Chang Zhou and Jingren Zhou and Xiaohuan Zhou and Tianhang Zhu},
  journal={arXiv preprint arXiv:2309.16609},
  year={2023}
}

@inproceedings{li2023lsdir,
  title={{LSDIR}: A large scale dataset for image restoration},
  author={Li, Yawei and Zhang, Kai and Liang, Jingyun and Cao, Jiezhang and Liu, Ce and Gong, Rui and Zhang, Yulun and Tang, Hao and Liu, Yun and Demandolx, Denis and others},
  booktitle=CVPRW,
  year={2023}
}

@inproceedings{timofte2017ntire,
  title={{NTIRE} 2017 challenge on single image super-resolution: Methods and results},
  author={Timofte, Radu and Agustsson, Eirikur and Van Gool, Luc and Yang, Ming-Hsuan and Zhang, Lei and others},
  booktitle=CVPRW,
  year={2017}
}

@article{xu2023imagereward,
  title={{ImageReward}: Learning and evaluating human preferences for text-to-image generation},
  author={Xu, Jiazheng and Liu, Xiao and Wu, Yuchen and Tong, Yuxuan and Li, Qinkai and Ding, Ming and Tang, Jie and Dong, Yuxiao},
  journal=NEURIPS,
  year={2023}
}

@article{bai2025vision,
  title={Vision-Language Models as Differentiable Semantic and Spatial Rewards for Text-to-{3D} Generation},
  author={Bai, Weimin and Li, Yubo and Luo, Weijian and Chen, Wenzheng and Sun, He},
  journal={arXiv preprint arXiv:2509.15772},
  year={2025}
}

@inproceedings{ghildyal2022stlpips,
  title={Shift-tolerant Perceptual Similarity Metric},
  author={Abhijay Ghildyal and Feng Liu},
  booktitle=ECCV,
  year={2022}
}

@article{ding2020iqa,
  title={Image Quality Assessment: Unifying Structure and Texture Similarity},
  author={Ding, Keyan and Ma, Kede and Wang, Shiqi and Simoncelli, Eero P.},
  journal = PAMI,
  year={2022},
}

@inproceedings{lin2024diffbir,
  title={{DiffBIR}: Toward blind image restoration with generative diffusion prior},
  author={Lin, Xinqi and He, Jingwen and Chen, Ziyan and Lyu, Zhaoyang and Dai, Bo and Yu, Fanghua and Qiao, Yu and Ouyang, Wanli and Dong, Chao},
  booktitle=ECCV,
  year={2024},
}

@inproceedings{noroozi2024you,
  title={You only need one step: Fast super-resolution with stable diffusion via scale distillation},
  author={Noroozi, Mehdi and Hadji, Isma and Martinez, Brais and Bulat, Adrian and Tzimiropoulos, Georgios},
  booktitle=ECCV,
  year={2024},
}

@article{hurst2024gpt,
  title={{GPT-4o} system card},
  author={Hurst, Aaron and Lerer, Adam and Goucher, Adam P and Perelman, Adam and Ramesh, Aditya and Clark, Aidan and Ostrow, AJ and Welihinda, Akila and Hayes, Alan and Radford, Alec and others},
  journal={arXiv preprint arXiv:2410.21276},
  year={2024}
}

@inproceedings{cai2025computer,
  title={Do computer vision foundation models learn the low-level characteristics of the human visual system?},
  author={Cai, Yancheng and Yin, Fei and Hammou, Dounia and Mantiuk, Rafal},
  booktitle=CVPR,
  year={2025}
}

@article{kingma2014adam,
  title={Adam: A method for stochastic optimization},
  author={Kingma, Diederik P and Ba, Jimmy},
  journal={arXiv preprint arXiv:1412.6980},
  year={2014}
}

@inproceedings{schuhmann2022laionb,
  title={{LAION}-5B: An open large-scale dataset for training next generation image-text models},
  author={Christoph Schuhmann and
          Romain Beaumont and
          Richard Vencu and
          Cade W Gordon and
          Ross Wightman and
          Mehdi Cherti and
          Theo Coombes and
          Aarush Katta and
          Clayton Mullis and
          Mitchell Wortsman and
          Patrick Schramowski and
          Srivatsa R Kundurthy and
          Katherine Crowson and
          Ludwig Schmidt and
          Robert Kaczmarczyk and
          Jenia Jitsev},
  booktitle=NeurIPS,
  year={2022},
}

@inproceedings{shao2019objects365,
  title={Objects365: A large-scale, high-quality dataset for object detection},
  author={Shao, Shuai and Li, Zeming and Zhang, Tianyuan and Peng, Chao and Yu, Gang and Zhang, Xiangyu and Li, Jing and Sun, Jian},
  booktitle=ICCV,
  year={2019}
}

@inproceedings{lin2014microsoft,
  title={Microsoft {COCO}: Common objects in context},
  author={Lin, Tsung-Yi and Maire, Michael and Belongie, Serge and Hays, James and Perona, Pietro and Ramanan, Deva and Doll{\'a}r, Piotr and Zitnick, C Lawrence},
  booktitle=ECCV,
  year={2014},
}

@inproceedings{liu2021swin,
  title={Swin transformer: Hierarchical vision transformer using shifted windows},
  author={Liu, Ze and Lin, Yutong and Cao, Yue and Hu, Han and Wei, Yixuan and Zhang, Zheng and Lin, Stephen and Guo, Baining},
  booktitle=ICCV,
  year={2021}
}

@article{min2019spair,
  title={{SPair-71k}: A large-scale benchmark for semantic correspondence},
  author={Min, Juhong and Lee, Jongmin and Ponce, Jean and Cho, Minsu},
  journal={arXiv preprint arXiv:1908.10543},
  year={2019}
}

@inproceedings{gu2019div8k,
  title={{DIV8K}: Diverse {8K} resolution image dataset},
  author={Gu, Shuhang and Lugmayr, Andreas and Danelljan, Martin and Fritsche, Manuel and Lamour, Julien and Timofte, Radu},
  booktitle=ICCVW,
  year={2019},
}

@inproceedings{zhao2023unipc,
  title={{UniPC}: A unified predictor-corrector framework for fast sampling of diffusion models},
  author={Zhao, Wenliang and Bai, Lujia and Rao, Yongming and Zhou, Jie and Lu, Jiwen},
  booktitle=NEURIPS,
  year={2023}
}

@inproceedings{cai2019toward,
  title={Toward real-world single image super-resolution: A new benchmark and a new model},
  author={Cai, Jianrui and Zeng, Hui and Yong, Hongwei and Cao, Zisheng and Zhang, Lei},
  booktitle=ICCV,
  year={2019}
}

@inproceedings{wei2020component,
  title={Component divide-and-conquer for real-world image super-resolution},
  author={Wei, Pengxu and Xie, Ziwei and Lu, Hannan and Zhan, Zongyuan and Ye, Qixiang and Zuo, Wangmeng and Lin, Liang},
  booktitle=ECCV,
  year={2020},
}

@article{prabhudesai2023aligning,
  title={Aligning text-to-image diffusion models with reward backpropagation},
  author={Prabhudesai, Mihir and Goyal, Anirudh and Pathak, Deepak and Fragkiadaki, Katerina},
  journal={arXiv preprint arXiv:2205.01917},
  year={2023}
}

@inproceedings{agustsson2017ntire,
  title={{NTIRE} 2017 challenge on single image super-resolution: Dataset and study},
  author={Agustsson, Eirikur and Timofte, Radu},
  booktitle=CVPRW,
  year={2017}
}

@inproceedings{sun2025pixel,
  title={Pixel-level and semantic-level adjustable super-resolution: A dual-lora approach},
  author={Sun, Lingchen and Wu, Rongyuan and Ma, Zhiyuan and Liu, Shuaizheng and Yi, Qiaosi and Zhang, Lei},
  booktitle=CVPR,
  year={2025}
}
}

\appendix
\clearpage
{
\begin{center}
\large{\textbf{Hallucination Score: Towards Mitigating Hallucinations in Generative Image Super-Resolution}} \\
\vspace{-5pt}

\begin{center}
\large{Supplementary Material}
\end{center}
\end{center}
}
\setcounter{section}{0}
\renewcommand\thesection{\Alph{section}}

\section{Limitations and Future Work}
\label{subsec:lim:broad:impact}

While this paper introduces a new metric called Hallucination Score (HS) and a method to reduce hallucination in generative super resolution, there are several avenues for future research. One limitation of our approach is that it evaluates hallucinations at the image level; a more nuanced analysis could investigate localizing hallucinatory regions within an image, potentially object-centric, which would be particularly valuable in practical applications where selective remedies for hallucinatory artifacts could be explored. Additionally, we relied on a proxy based on DINO and CLIP to approximate MLLM outputs due to computational constraints. Future work could explore developing a lightweight version of an MLLM, enabling direct back-propagation through the model and potentially leading to better results. Moreover, one could investigate the effectiveness of loss based on mid-level features while training diffusion-based GSR models in the first place.

\section{More Information on the GPT-based Hallucination Score Generation}
\label{sec:prompt-mllm}
\subsection{Prompt and Experimental Setup}
\begin{figure*}[htb]
    \centering
    \includegraphics[width=0.98\textwidth]{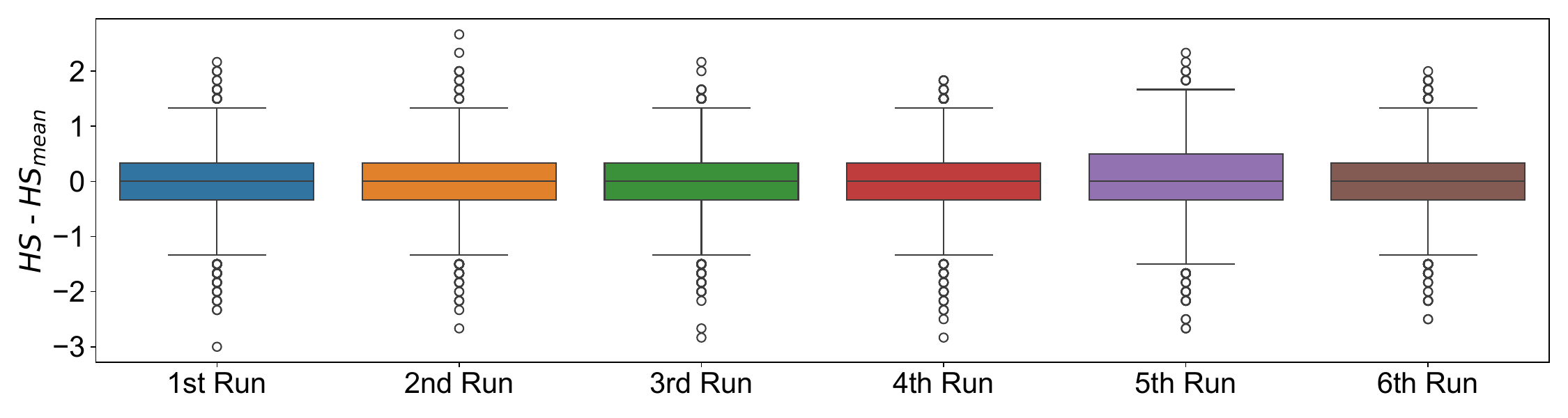}
    \caption{\textbf{Differences of HS across multiple runs}. We calculate the mean of HS ($HS_{mean}$) across all the six runs for each image and plot the differences between the $HS$ of each run with their mean ($HS_{mean}$).
    }
    \label{fig:hs_stats}
\end{figure*}

We provide the complete prompt, which we abbreviate in Fig.~\ref{fig:prompt} and use in conjunction with the GPT-4o-2024-08-06 model, in Fig.~\ref{fig:promptfull}. Moreover,  we investigate the stability of HS scores generated by MLLM across multiple runs. Specifically, we generate the HS six times on the same set of 3000 images in the SS-TS dataset (cropped from DIV2K-Val), super-resolved by the StableSR model. After that, we calculate the mean HS per image across those runs, denoted by $HS_{mean}$. For each run, we plot the score differences between the score for an image in the current run and the mean score for that image across all six runs. The results are shown in Fig.~\ref{fig:hs_stats}. As we can see, the differences for the HS of each image is minimal across several runs.

In terms of latency and cost, each set of inputs to GPT-4o consists of the LR, SR, and GT, along with the prompt. The cost of processing 3000 examples is $\sim$5 USD and takes $\sim$8 minutes.

\paragraph{Prompt Robustness}
\label{supp:promptrobustness}
To check the dependence of HS on prompt wording,
we generated two alternative prompts by asking GPT to reword the original one. We display one of these rewordings in Fig.~\ref{fig:prompt_reword_v2}.
Operating on the SS-TS, both rewordings obtain a Spearman correlation of 0.66 to the original.
For reference, humans with the same task description have a \textit{lower} correlation of 0.54 (\ie, average pairwise inter-rater agreement; see \S\ref{sec:correlationan}).

\subsection{Additional MLLM-Based Metric Statistics}
\label{sec:additional_stats}
In addition, we provide HS statistics in Table~\ref{tab:mllmstats}, finding that diffusion-based approaches (especially SeeSR and PASD) tend to hallucinate more than the non-diffusion-based Swin2SR. 
Indeed, Swin2SR not only has the highest mean HS, but also the smallest number of outputs ($19.3\%$) with the score of $1$ or $2$ (\ie, significant and considerable hallucination; see Fig.~\ref{fig:prompt}). 
To an extent, we also find that ``easy'' and ``hard'', in terms of hallucination, is dependent on %
image content itself, not just model choice.
Specifically, the diffusion models have an average correlation with each other of $0.34$, suggesting non-trivial concordance across models (\ie, the same image tends to be similarly rated across models). 
Interestingly, this does not depend on diffusion: the average correlation between Swin2SR and the other GSR models is similar ($0.31$).
\begin{table}[htb]
    \centering
    \caption{\textbf{GPT-based Hallucination Scores of SR models.}
        Values are computed over full StableSR Test Set (SS-TS; 3K images).
        The better scores of the non-generative Swin2SR conform to the intuition that GSR is more prone to hallucinate.}
    \begin{tabular}{l|c:ccccc}
        \multirow{2}{*}{Method} & \multirow{ 2}{*}{\shortstack{Mean\\Score}} & \multicolumn{5}{c}
        {Score Percentages} \\
                                &                        & 1 & 2 & 3 & 4 & 5 \\\hline
        Swin2SR  & 3.38 & 6.5 & 12.8 & 33.2 & 30.7 & 16.8 \\
        StableSR & 3.36 & 5.9 & 19.0 & 26.6 & 30.1 & 18.4 \\
        SeeSR    & 2.99 & 14.2 & 23.7 & 25.0 & 22.8 & 14.3 \\
        PASD     & 2.45 & 26.3 & 30.2 & 22.6 & 13.4 & 7.5
    \end{tabular}
    \label{tab:mllmstats}
\end{table}

\begin{figure*}
    \centering
    \begin{minipage}{0.95\textwidth}
        \begin{lstlisting}[
    basicstyle=\small, %
]
You will receive three images for evaluation:
1. **Ground Truth (GT)**: The reference high-resolution image.
2. **Low-Resolution Input (LR)**: The degraded, low-resolution input image provided to an AI model.
3. **Super-Resolved Image (SR)**: The output high-resolution image generated by an AI super-resolution model based solely on the LR image.
**Task:**  
Evaluate the SR image for "hallucinations," which are imaginary details or content added by the model that are not present in the GT image.
#### Criteria for Evaluation:
- **Hallucinations** are newly added visual contents that significantly differ from the GT image. 
- Mere **lack of detail**, blurry textures, or lower image quality (due to severe damage in the LR image) should **not** be considered hallucinations. Such artifacts are understandable, given original input limitations.
- Focus specifically on added details that **change the semantic meaning** (new objects, significant alterations of scene elements) or generate **perceptually jarring inaccuracies** (e.g., incorrect facial features, unreadable or distorted text).
#### How to assign scores (1-5 scale):
- **1 (Significant Hallucinations):** Multiple severe hallucinations causing major semantic changes or perceptually disturbing artifacts, such as completely invented objects, critically incorrect text, or distorted faces.
- **2 (Considerable Hallucinations):** Noticeable hallucinations that notably alter semantics or significantly degrade perception (e.g., introducing partially incorrect objects, faces, or text).
- **3 (Mild Hallucinations):** Minor added contents, typically at the texture or detail level, slightly affecting semantic interpretation; perceptually noticeable but not severely disturbing.
- **4 (Minimal Hallucinations):** Very minor discrepancies at texture or detail level only perceptible upon careful inspection; negligible semantic or perceptual effect.
- **5 (Artifact-free):** SR image has no hallucinations; entirely faithful to GT image (aside from acceptable quality differences arising from LR limitations).
Your response must strictly adhere to the following JSON format and include brief but clear reasoning for your evaluation:
```json
{
  "score": <integer from 1 to 5>,
  "reasoning": "<Provide clear justification for the assigned rating, focusing primarily on the presence and severity of hallucinated details compared to the GT and LR images.>"
}
```
Output nothing else besides this JSON.    
\end{lstlisting}
    \end{minipage}
    \caption{
        Complete Prompt.
        We show the full prompt, used to obtain our MLLM-based Hallucination Score (HS). See also Fig.~\ref{fig:prompt}.
    }
    \label{fig:promptfull}
\end{figure*}

\begin{figure*}
    \centering
    \begin{minipage}{0.95\textwidth}
        \begin{lstlisting}[
    basicstyle=\tiny, %
]
You will be provided with three images for evaluation:

Ground Truth (GT): The authentic high-resolution reference image.

Low-Resolution (LR): The degraded input image used by the super-resolution model.

Super-Resolved (SR): The model's high-resolution output generated solely from the LR image.

Task:
Judge the SR image for the presence of hallucinations-visual content created by the model that does not appear in the GT image.

Evaluation Criteria:
Hallucinations refer to fabricated details that differ noticeably from the GT.

Do not count poor quality, blur, or missing detail as hallucinations if they stem from limitations in the LR image.

Focus on any additions that change semantic interpretation (e.g., made-up objects, incorrect features) or introduce jarring inconsistencies (e.g., mangled text, unnatural shapes).

Scoring System (1 to 5 scale):
1 (Extensive Hallucinations): Multiple major artifacts or fabricated elements that strongly disrupt scene understanding or realism.

2 (Strong Hallucinations): Clearly visible hallucinated features that interfere with interpretation or coherence.

3 (Mild Hallucinations): Some minor, invented content-mostly at the fine detail level-that slightly affects perception.

4 (Subtle Hallucinations): Few and minor discrepancies; perceptually negligible or hard to notice.

5 (No Hallucinations): SR is completely consistent with the GT aside from acceptable differences due to LR degradation.

Please respond strictly using the following JSON format and include a brief rationale for the score:

```json
{
  "score": <integer from 1 to 5>,
  "reasoning": "<Provide a clear explanation for the given rating, focusing mainly on the presence and impact of hallucinated elements compared to the GT and LR images.>"
}
```
Return only this JSON - do not include any extra comments or formatting.
\end{lstlisting}
    \end{minipage}
    \caption{
        We show another variation of the prompt used in the prompt robustness experiment (\S\ref{supp:promptrobustness}). 
        See also the full prompt, in Fig.~\ref{fig:promptfull}, and the illustration of the prompt in Fig.~\ref{fig:prompt}.
    }
    \label{fig:prompt_reword_v2}
\end{figure*}

\section{Models Used in Correlation Analysis}
\label{supp:andetails}

In this section, we provide additional details on the choices of the off-the-shelf models, their architectures, and the method to compute cosine distance between GTI and SRI images needed to obtain correlations in Table \ref{tab:correlations} and \S\ref{sec:correlationan} in the main paper.

\subsection{Neural Feature Distance}
As discussed in \S\ref{sec:existingmetrics} in the main paper, we compute cosine distance between features extracted from DINOv2 \cite{oquab2023dinov2} and CLIP \cite{radford2021learning} on GTI and SRI. For both DINOv2 and CLIP, we consider two versions, one using spatial tokens (\texttt{*-ST}) and the other, CLS token (\texttt{*-CLS}).

\paragraph{DINOv2} We adopt DINOv2 with registers \cite{darcet2023vision} with ViT-B/14 model architecture\footnote{\url{https://github.com/facebookresearch/dinov2}}. We resize the input images from $512$ to $518$ in order to be compatible with the patch size of $14$. For \texttt{DINO-CLS}, we extract CLS token feature of dimensions $1\times768$, and for \texttt{DINO-ST} we extract patch token features of dimensions $37 \times 37 \times768$. We note that both CLS and patch token features are obtained after normalization using \texttt{nn.LayerNorm}, excluding the tokens specific to registers. For \texttt{*-interm} we obtain intermediate features from layers $1,3,5,7,9,11$, where the $11^{th}$ layer is the last layer.

\paragraph{CLIP} We use OpenCLIP \cite{cherti2023reproducible} with ViT-B/16 model architecture pre-trained on LAION-2B \cite{schuhmann2022laionb}. We take the input images of size $512$. For \texttt{CLIP-CLS}, we extract normalized CLS token feature of dimensions $1\times768$, and for \texttt{CLIP-ST} we extract normalized patch token features of dimensions $32 \times 32\times768$. We note that normalization refers to division with L2-norm along feature dimension, consistent with OpenCLIP \cite{cherti2023reproducible}. Similar to above, for \texttt{*-interm} we obtain intermediate features from layer indices $1,3,5,7,9,11$, where the $11^{th}$ layer is the last layer.

Lastly, to obtain distance, we compute cosine distance between extracted features from GTI and SRI, and take a mean on the distances across spatial tokens in the case of \texttt{*-ST} to obtain a scalar.

\subsection{Semantic Segmentation Divergence (SSD)}

To estimate semantic changes between the GTI and SRI, we use an Open Vocabulary Semantic Segmentation framework, OpenSeeD \cite{zhang2023simple}. 
As a first step, we extract tags or common object categories on GTI using Recognize Anything model (RAM++ \cite{zhang2023recognize,huang2023open}). We then use the resulting tags to define vocabulary for object categories in OpenSeeD, followed by segmentation results on GTI and SRI in the form of per-pixel distribution over the pre-extracted tags.

For OpenSeeD, we use the provided checkpoint on open vocabulary model pre-trained on panoptic segmentation (COCO 2017 \cite{lin2014microsoft}) and object detection tasks (Objects365 \cite{shao2019objects365}), with Swin-T \cite{liu2021swin} as the backbone.

Finally, we compute KL divergence on the resulting per-pixel distributions between the GTI and SRI, and average across pixels to obtain the final distance.

\subsection{Neural Correspondence Features}
\label{supp:subsec:neuralcorres}
\paragraph{Telling Left from Right (TLR).} We follow the default setup in TLR\footnote{\url{https://github.com/Junyi42/geoaware-sc}} \cite{zhang2024telling} which uses Stable Diffusion 1.5 \cite{rombach2022high} and DINOv2 ViT-B/14 \cite{oquab2023dinov2} to obtain fused multi-scale features, and applies a four bottleneck residual layers pre-trained on SPair-71k \cite{min2019spair} dataset, to obtain semantic correspondence. In our case, we simply fetch post-processed features on GTI and SRI and obtain cosine distance.

\paragraph{DeepViT.} We use the DeepViT\footnote{\url{https://github.com/ShirAmir/dino-vit-features}} \cite{min2019spair} feature extractor based on the DINOv1 ViT-S/8 architecture. Specifically, the features are obtained from the $9^{th}$ layer, followed by log-binning for additional spatial context. We then measure the cosine distance between the resulting features from GTI and SRI.

\section{Correlation Analysis of Human Ratings}
\label{supp:human}

\subsection{Dataset}
\label{supp:sstshuman}
The StableSR Test Set (SS-TS) \cite{wang2024exploiting} consists of patches derived from 92 whole images (a subset of the 100 DIV2K-Val \cite{div2k} images). To ensure image diversity, we extract one crop (patch) from each image. Specifically, we select the crop with the median position, or roughly at the center of the image. We then super-resolve these crops with the three GSR models (PASD~\cite{yang2024pasd}, SeeSR~\cite{wu2024seesr}, and StableSR~\cite{wang2024exploiting}), and ask 11 human raters to evaluate the hallucination levels present.

\subsection{Additional Statistics}
\begin{figure*}[t]
    \centering
    \includegraphics[width=0.99\linewidth]{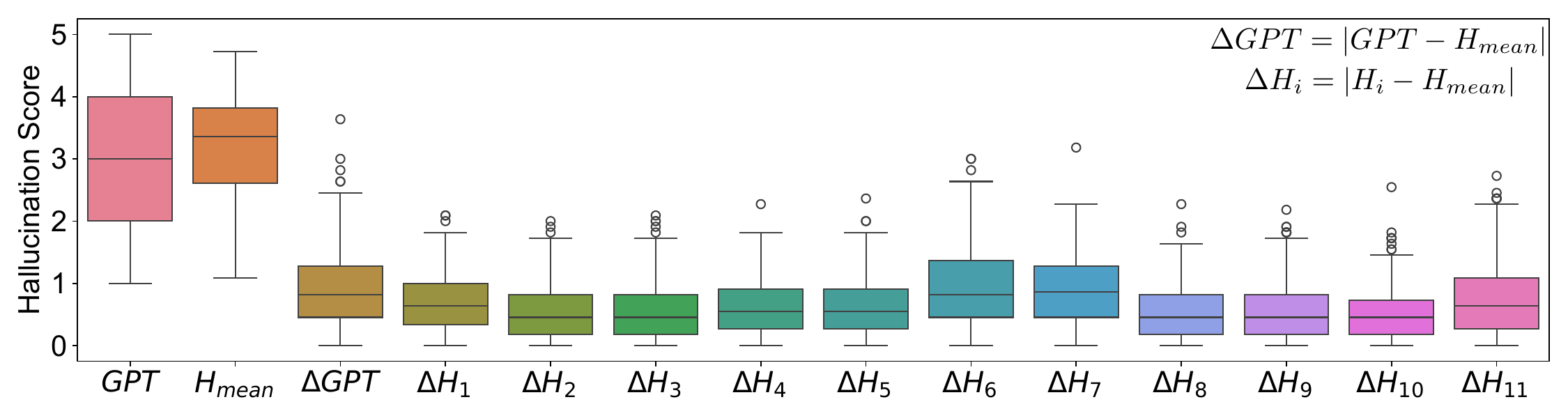}
    \vspace{-0.5em}
    \caption{
        \textbf{Comparison of GPT with Human scores}. In a user study with 276 SR output images, each rated (1-5) by 11 human evaluators, we plot the absolute difference between mean of human scores ($H_{mean}$, averaged across humans per image) with humans and MLLM denoted by $\Delta \text{H}_i$ and $\Delta \text{GPT}$ respectively, where $i$ denotes one of 11 total humans. We observe $\Delta \text{GPT}$ is well within the range of human inter-rater variability. %
    }
    \label{fig:score-dist-bar-plot}
    \vspace{-0.5em}
\end{figure*}

\begin{figure*}[htb]
    \centering
    \includegraphics[width=1\textwidth]{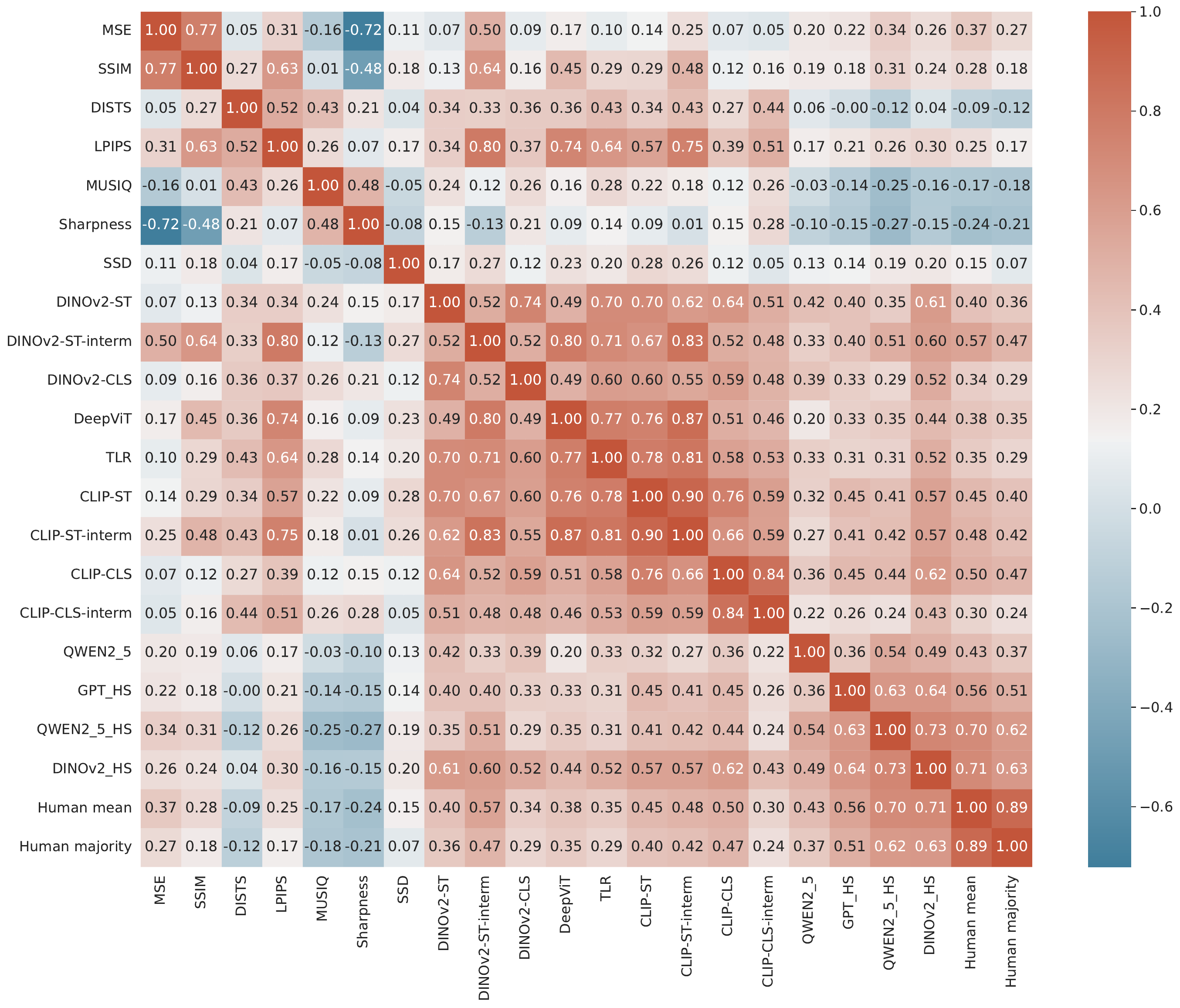}
    \caption{\textbf{Spearman correlation heatmap of human evaluation with GPT-4o and other metrics}. This map extends Table \ref{tab:humancorrelations}. We found that (i) humans (= \texttt{Human mean} and \texttt{Human majority}) have high correlations ($0.56$ and $0.51$, respectively) with GPT-4o \cite{hurst2024gpt} (=\texttt{GPT-HS}) scores compared to other perceptual, semantic and feature-based metrics described in \S\ref{sec:correlationan}. Further (ii), among the \textit{un}tuned metrics, neural feature distances based on DINOv2 \cite{oquab2023dinov2} and CLIP \cite{cherti2023reproducible,radford2021learning} correlates the most with GPT-4o, especially their intermediate feature variants (\texttt{*-interm}).
    However (iii), our HS models fine-tuned on GPT-HS outputs (\S\ref{supp:proxy:model}), namely Qwen-HS and DINO-HS, have the highest correlation to both human scores (mean and majority) and GPT-HS itself, by a significant margin.
    The user study was conducted on median crops (roughly centered) obtained from all the $92$ DIV-2K val \cite{div2k} images used by the StableSR Test Set (SS-TS) \cite{wang2024exploiting}. Eleven human subjects rated the images (from 1-5) on the SR outputs from three diffusion-based models (\ie, StableSR, SeeSR, and PASD), totalling $276$ images  ($92 \times 3$). 
    \textbf{Note}: Spearman correlations done on less than 500 samples are indicative of trends but not the exact values. 
    }
    \label{fig:heatmap-human-eval}
\end{figure*}

In the user study, for each of the diffusion-based models (\ie, StableSR, SeeSR, and PASD), human annotators assigned a score in the range of 1 to 5 for the $92$ SRIs, while given the corresponding LRI and GTI as the reference.
In \S\ref{subsec:mllmhalluscore} and Fig.~\ref{fig:score-dist-bar-plot}, we show the distribution of scores from GPT is well within the range of human inter-rater variability. In this section, similar to Table \ref{fig:mllm_sample_outputs} of the main paper, we additionally visualize a heatmap of Spearman rank correlations among human average and human majority scores, along with the metrics described in \S\ref{sec:correlationan} across $276$ ($92 \times 3$) images, shown in Fig. \ref{fig:heatmap-human-eval}. Human aggregate (mean / majority) scores are computed per image across all human raters (11 in total). We further note that Spearman correlations performed on less than 500 samples\footnote{\url{https://docs.scipy.org/doc/scipy/reference/generated/scipy.stats.spearmanr.html}} are indicative of trends but not the exact values.

\noindent
\textbf{Inter-rater Agreement}. 
As an additional measure of inter-rater agreement, we compute the Cohen-$\kappa$ \cite{scikitlearn,cohen1960coefficient} between users, obtaining a pairwise mean of 0.50 (std.~dev.~$0.122$).

In Fig.~\ref{fig:score-dist-bar-plot},
we also plot absolute difference in scores between human mean with (i) MLLM (denoted as $\Delta\text{GPT}$), and (ii) each human ($\Delta\text{H}_i$). We observe $\Delta\text{GPT}$ to have similar statistical properties as the humans $\Delta\text{H}_i$, where specifically the median and quantiles lie within a similar range. 
This shows $\Delta \text{GPT}$ is well within the range of human inter-rater variability. See also the discussion in \S\ref{sec:correlationan}.

\section{Correlation Analysis of GPT-HS}
\label{sec:add-heatmap-individual}
We follow up on the analysis described in \S\ref{sec:correlationan}, and 
provide correlation heatmaps and average metrics for the individual models.

\begin{table}[t]
\centering
\caption{\textbf{Average over metrics on the SS-TS dataset.} As a companion to Table \ref{tab:correlations} in the main paper, we aggregate and average the metrics across the SS-TS dataset (\ie, the 3K DIV-2K validation crops with degradations, released by \cite{wang2024exploiting}). Last column (``Combined'') is the aggregated result across the four models. 
}
    \begin{adjustbox}{max width=0.99\linewidth}
\begin{tabular}{@{}lcccc:c@{}}
\toprule
Metric & StableSR & SeeSR  & PASD & Swin2SR & Combined\\ \midrule
MSE ($\times$ 1e3) $\downarrow$ & 9.487 & 8.589 & 8.248 & \textbf{5.934} & 8.064\\
SSIM $\uparrow$ & 0.534 & 0.567 & 0.578 & \textbf{0.648} & 0.582\\
DISTS $\downarrow$ & 0.205 & \textbf{0.197} & 0.220 & 0.295 & 0.229\\
LPIPS $\downarrow$ & \textbf{0.311} & 0.319 & 0.375 & 0.473 & 0.370\\
MUSIQ $\uparrow$ & 65.918 & \textbf{68.672} & 64.079 & 44.372 & 60.76\\
Sharpness $\uparrow$ & \textbf{105.01} & 84.01 & 56.94 & 6.57 & 63.13\\
SSD ($\times$ 1e3) $\downarrow$ & \textbf{7.621} & 7.844 & 9.428 & 12.872 & 9.441 \\
DINOv2-ST $\downarrow$ & \textbf{0.351} & 0.356 & 0.432 & 0.432 & 0.393\\
DINOv2-ST-interm $\downarrow$ & \textbf{0.111} & 0.117 & 0.135 & 0.161 & 0.131\\
DINOv2-CLS $\downarrow$ & \textbf{0.297} & 0.317 & 0.441 & 0.454 & 0.377\\
DeepViT $\downarrow$ & \textbf{0.199} & 0.204 & 0.234 & 0.254 & 0.222\\
TLR $\downarrow$ & \textbf{0.221} & 0.223 & 0.257 & 0.293 & 0.248\\
CLIP-ST $\downarrow$ & 0.385 & \textbf{0.381} & 0.427 & 0.443 & 0.409\\
CLIP-ST-interm $\downarrow$ & 0.285 & \textbf{0.284} & 0.315 & 0.322 & 0.301\\
CLIP-CLS $\downarrow$ & 0.152 & \textbf{0.150} & 0.206 & 0.264 & 0.193\\
GPT-HS $\uparrow$ & 3.361 & 2.992 & 2.455 & \textbf{3.383} & 3.048\\ 
Qwen-HS $\uparrow$ & 2.997 & 2.770 & 2.415 & \textbf{3.166} & 2.837\\ 
DINOv2-HS $\uparrow$ & 3.326 & 3.176 & 2.483 & \textbf{3.395} & 3.095\\ 
\bottomrule
\end{tabular}%
\end{adjustbox}
\label{tab:avg-metrics-across-models}
\end{table}

\paragraph{Average metrics.}
In Table \ref{tab:correlations} of the main paper, we presented Spearman correlation of MLLM with the metrics described in  \S\ref{sec:correlationan}. In this section, we provide an average across the SS-TS dataset (3K images) for each metric in Table \ref{tab:avg-metrics-across-models}. 
The average across metrics help us compare their absolute values across various types of models. We observe non-diffusion approach (Swin2SR) perform best with MSE and SSIM, suggesting high fidelity compared to diffusion-based models. On the other hand, diffusion-based models outperform on perceptual quality (\eg, LPIPS, MUSIQ). Within diffusion-based models, StableSR and SeeSR perform better than PASD over semantic-aware metrics (DINO/CLIP) and GPT-4o score, indicating lower hallucinatory artifacts.

\begin{figure*}[]
    \centering
    \includegraphics[width=0.99\textwidth]{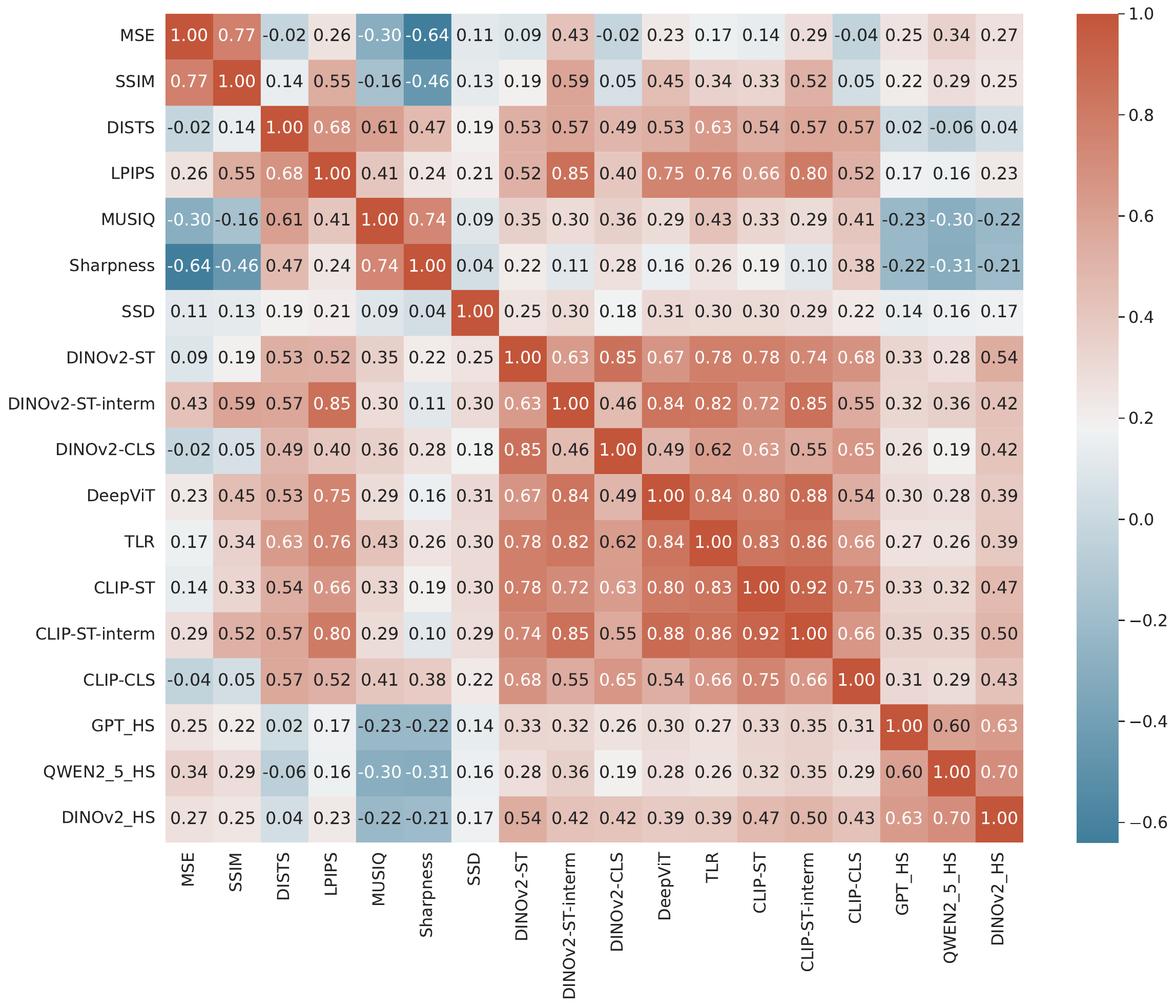}
    \caption{\textbf{Spearman correlation heatmap for combined models.} 
    This map extends Table~\ref{tab:correlations} to show the pairwise correlations between all metrics for the combined models (StableSR, SeeSR, PASD, and Swin2SR), run on the SS-TS for each (12K crops in total), rather than only the correlation to GPT-HS.
    Note that the correlations to GPT-HS for existing metrics and affinities are relatively low (excluding our trained HS proxies), with none going above 0.35 in correlation.
    This suggests that GPT-HS measures a notion of hallucination that is not captured well by existing methods.
    In contrast, our fine-tuned proxies (trained on GPT-HS outputs) have substantial correlations (0.60 and 0.63), similar to the magnitude of human inter-rater agreement (0.54; see \S\ref{sec:correlationan}) and human-mean-to-GPT correlation (0.56); further,
    note that Qwen-HS and DINO-HS have a 0.70 correlation.
    Thus, since all three methods still have non-trivial disagreements with each other, we utilize all three in our evaluations in Table~\ref{tab:results_injection}.
    See also Fig.~\ref{fig:heatmap-human-eval} for pairwise correlations including human scoring. 
    }
\label{fig:heatmaps-combined-models-ss}
\end{figure*}

\paragraph{Spearman correlation heatmap for combined models.}
In Fig.~\ref{fig:heatmaps-combined-models-ss}, we show Spearman correlation heatmap for combined (StableSR, SeeSR, PASD, and Swin2SR) models across 12K (4 $\times$ 3K, from the SS-TS) images. In particular, we observe last-layer features from DINO/CLIP do not correlate well with MSE/SSIM compared to MLLM (GPT), suggesting the efficacy of higher-level semantic concepts to capture hallucinatory artifacts compared to low-level metrics.

\paragraph{HS Types.}
As shown in Table \ref{tab:correlations}, DINO-HS and Qwen-HS best correlate to GPT-HS. Further, despite being trained on GPT-HS outputs, they actually outperform GPT-HS in terms of human correlation (Table \ref{tab:humancorrelations}).
On the full SS-TS (12K crops, as in Table~\ref{tab:correlations}), we find that DINO-HS  and Qwen-HS have a correlation (both Pearson and Spearman) of 0.70, similar to their correlations with GPT. 
This suggests that the two trained proxies are strongly correlated. 
For comparison, inter-human Spearman correlation is 0.54 (see also \S\ref{sec:correlationan}).
Notice that Qwen-HS and GPT-HS provide textual explanations along with their discrete scores; however, the benefits of DINO-HS include superior efficiency (in memory and time), as well as the presence of a continuous score. Thus, we consider all three metrics in our evaluation. 
We remark that we briefly attempted to optimize Qwen-HS with AlignProp. However, we found the training to be unstable, sometimes resulting in a model that outputs severe artifacts. %
For this reason, as well as computational efficiency, we turned to our DINO-based proxy fine-tuning approach instead (as described in \S\ref{sec:results}).

\section{Hallucination Score Proxy Details}

\subsection{MLLM-HS Proxy Training Dataset}
\label{supp:proxy:dataset}
We require a dataset of LQ, SR, and GT images, along with associated GPT-derived HSs, in order to train our proxy models.
To do so, we run Swin2SR, PASD, and SeeSR to obtain SR outputs.
Specifically, for each model, we only generate samples via datasets that were not yet seen by the model, to ensure the resulting outputs simulate the behaviour of a ``new'' test input. Specifically, we used LSDIR for PASD and Swin2SR, while for SeeSR we combine DIV2K, DIV8K, and Flickr2K. Since Swin2SR has relatively few hallucinations, we generated less data for it (only $\sim$2000 images). The resulting dataset has 30,245 training tuples, plus an additional 303 held-out examples for validation. It does not include DIV2K-Val, which forms the basis of the SS-TS we use for analysis and evaluation, nor does it include the evaluation sets RealSR and DRealSR.

\subsection{Architecture and Training Details}
\label{supp:proxy:model}

\noindent
\paragraph{CNN-Based Architecture.}
As noted in the main paper, we trained a ResNet-50 (pretrained on ImageNet), to regress GPT-HS score.
The input is three images (LQ, SR, and GT), so nine channels, while the output is simply a scalar, trained via the GPT-HS scores on model outputs (see \S\ref{supp:proxy:dataset} above).
Note that we also trained a no-reference (NR) version of the CNN architecture (see \S\ref{supp:nrhs}). The only difference, compared to the standard version, is that the NR-CNN takes in two images (six channels, for LR and SR), instead of three.

\noindent
\paragraph{DINO-HS Architecture.}
Given the good correlation properties of DINO (see Tables \ref{tab:humancorrelations} and \ref{tab:correlations}), we fine-tune it to obtain our DINO-HS approximator.
In particular, we assume that we can build off the metric we defined for correlation analysis, namely the cosine similarity of the DINO features of the GTI versus those of SRI. 
Formally, we define $\widehat{h} = h_s(S_c(f(I_\mathrm{SR}),f(I_\mathrm{GT})))$, where $f$ is a DINO-based feature extractor \cite{darcet2023vision} (the DINOv2-B model with registers), $S_c$ is cosine similarity, and $h_s$ alters the similarity to match the HS.
We remark that we use the post-normalized spatial tokens of the last layer (\ie, eleventh block, denoted \texttt{x\_norm\_patchtokens}) to compute the cosine similarity per token, followed by averaging. The learnable scale and shift, $h_s$, map the scalar similarity to the HS.
This procedure is used for direct HS estimation.
However, for using the DINO-HS model as a reward in AlignProp (see \S\ref{sec:results}), we slightly modify the procedure.
Namely, we take the outputs of the odd blocks (1, 3, 5, 7, 9, and 11), concatenate them together, and compute the final cosine similarity on the result.
We find that this provides a more performant reward: without doing this, the resulting fine-tuned model experiences a severe drop in both perceptual quality (according to NR-IQA metrics, such as MUSIQ) and fidelity (\eg, LPIPS).

\noindent
\paragraph{Training Details: CNN and DINO-HS.}
Both models are trained with a combined regression and correlation loss:
$ \mathcal{L}(\mathcal{I}, S) = 
      (\gamma_r/n_b) \sum_j w_j || \widehat{S}(\mathcal{I})_j - S_j ||_a 
    + \gamma_c \mathcal{P}[\widehat{S}(\mathcal{I})] $,
where 
$\mathcal{I}$ is the data batch of length $n_b$ 
    (with GT GPT-HS scores $S$),
$\mathcal{P}$ is the Pearson correlation,
and $\widehat{S}$ is the estimated HS.
The loss weights ($\gamma_r = 1, \gamma_c = 0.5$) and parameters $a = 2$ are set empirically.
Due to the severe class imbalance in the data (namely, the scores one to five have the following percentages: 29.9, 32.6, 19.8, 12.3, and 5.4), 
we weight each sample by the rarity of its label ($w_j = (1/f_{\ell(j)})^p$, where $\ell$ is the label (HS), $f_\ell$ is the frequency of label $\ell$, and $p$ is a hyper-parameter we empirically set to 0.75).
For the CNN, we remark that ablating the correlation loss and class imbalance reweighting causes the correlation to human scores to decline 
(Spearman to human mean: 0.51 vs.\ 0.45; human majority: 0.44 vs.\ 0.41).
Both models are optimized by Adam \cite{kingma2014adam}. 
DINO-HS and CNN have learning rates $10^{-6}$ and $2\times10^{-4}$, and batch sizes $n_b$ of 24 and 64.
For DINO-HS, we only optimize the MLPs and attention matrices of the last four blocks (8, 9, 10, and 11), to prevent catastrophic forgetting of the rich information in the original DINO.
The CNN allows all weights to be trained.
In general, we chose hyper-parameters and early stopping times by checking the correlation to GPT on a held-out validation set (as mentioned in \S\ref{supp:proxy:dataset}).

\paragraph{Alternative MLLM.}
\label{supp:alternativemllm}
We also considered Qwen2.5-VL-7B model \cite{qwen,bai2025qwen2}, 
    which reduces cost, accessibility, and efficiency issues with GPT.
Further, 
    we obtain a finetuned model (denoted Qwen-HS), 
    using our dataset of HS-labeled images \textit{from GPT} (see \S\ref{subsec:hs:proxy} and \S~\ref{supp:proxy:dataset}). More specifically, we finetune the Qwen2.5-VL-7B model and perform SFT with the dataset. The model takes GT, LR, and SR images, in that order, along with the prompt shown in Fig.~\ref{fig:promptfull} as inputs, and generates HS and the corresponding reasoning as the output. In this training, we fine-tune the LLM and visual merger modules, leaving the vision encoder frozen, for 1 epoch with a learning rate of $1e^{-6}$ and a batch size of $128$.
In terms of correlation to humans, untuned Qwen underperforms GPT (human mean: 0.43 vs 0.56; majority: 0.37 vs 0.51), but Qwen-HS actually \textit{out}performs GPT (0.70/0.62 for mean/majority), despite being trained on GPT outputs. 
This may be due to fine-tuning reallocating model capacity.
Interestingly, while Qwen-HS has a 0.54 rank correlation to GPT (on 12K images, via SS-TS on four GSR models), the models are usually close in score:
a difference in HS
of 0, 1, 2, 3, and 4 occur with frequency
0.378, 0.446, 0.143, 0.027, and 0.006.
In words, 82.4\% of Qwen-GPT judgment pairs are within one HS.

We remark that our Qwen-HS model could, in theory, be utilized for direct optimization (which we perform in \S\ref{sec:results} via DINO-HS), as others have considered (\eg, \cite{bai2025vision}). 
However, our preliminary experiments found this process to be unstable and unable to compete with our adapted deep features proxy. We leave further investigation to future work.

\begin{figure*}
    \centering
    \begin{minipage}{0.95\textwidth}
        \begin{lstlisting}[
    basicstyle=\tiny, %
]

You will receive two images:

1. **Low-Resolution Input (LR):** A degraded image that serves as the input to a super-resolution model.  
2. **Super-Resolved Image (SR):** The high-resolution image generated by the model based solely on the LR input.

**Task:**  
Evaluate the SR image for **hallucinations**-details that appear implausible, inconsistent with the LR image, or semantically incorrect based on what can reasonably be inferred from the LR input.

#### Evaluation Guidelines:
- A **hallucination** refers to invented content in the SR that **cannot be reliably inferred** from the LR, or that appears **semantically incorrect**, **unrealistic**, or **incoherent**.
- Do **not** penalize the SR for lacking detail or for slight texture smoothing-this is expected given the low quality of the LR.
- Focus on signs of **fabricated structures**, **unrealistic patterns**, or **semantically wrong content** (e.g., facial distortions, incorrect text rendering, strange object shapes).

#### Scoring Scale (1-5):
- **1 (Strong Hallucinations):** Clear and frequent semantic distortions or invented details (e.g., distorted faces, unreadable or unrealistic text, fabricated structures).
- **2 (Moderate Hallucinations):** Noticeable hallucinations that are inconsistent with the LR but don't completely break semantic plausibility.
- **3 (Mild Hallucinations):** Some hallucinated textures or minor inconsistencies, but overall visually plausible.
- **4 (Minimal Hallucinations):** Very few and subtle hallucinated details; high consistency with the LR.
- **5 (No Hallucinations):** SR image appears fully consistent with LR input; no visual or semantic artifacts suggesting invented content.

Please respond using **only** the following JSON format:

```json
{
  "score": <integer from 1 to 5>,
  "reasoning": "<Provide a clear explanation for the score, focusing on any fabricated or implausible details in SR relative to the LR input.>"
}
```

\end{lstlisting}
    \end{minipage}
    \caption{
        We show the prompt used for no-reference (NR) HS estimation. (\S\ref{supp:nrhs}). 
        See also the full prompt, in Fig.~\ref{fig:promptfull}, and the illustration of the prompt in Fig.~\ref{fig:prompt}.
    }
    \label{fig:prompt_nr_iqa}
\end{figure*}

\subsection{No-Reference (NR) HS Estimation}
\label{supp:nrhs}
While our FR HS can be applied to both evaluation and optimization, as we do in this paper, its use of an HQ GT input limits some test-time applications.
We therefore considered estimation with an NR model as well.

\noindent\textbf{GPT-NR.}
We first considered a simple modification of our GPT-based approach, by modifying the prompt and not sending the HQ GT to the model (i.e., it only receives the LQ and SR images). The revised prompt for NR HS estimation can be found in Fig.~\ref{fig:prompt_nr_iqa}.
The resulting model, which we denote GPT-NR, therefore attempts to judge the SR image in isolation.
We find that the Spearman correlations to human scores declines significantly, by around $\sim$17\%:
0.51 to 0.42 (majority) and 0.56 to 0.47 (mean).
Pearson correlations also decline, though more modestly: 
0.50 to 0.45 (majority) and 0.55 to 0.50 (mean).
Note that the human scores are decided \textit{with} access to the GT, just as our standard GPT-HS operates; hence, the NR model has access to less information than the human judgments to which we are correlating, and some loss in performance is expected.
Overall, these results suggest that significant aspects of our hallucination measures can still be captured \textit{without} access to GT, albeit with slightly reduced accuracy in terms of human judgments.
Since our uses for HS in this paper (evaluation and reward-based fine-tuning) occur in scenarios with access to the GT, we utilize our FR models instead and leave their application to future work.

\noindent\textbf{CNN-NR.}
We also tested our CNN-based HS proxy in an NR form
(see also \S\ref{supp:proxy:model}), where the RN50 predictor only has access to the LQ and SR image.
Similar to the GPT-NR case, we find that correlation to human mean scores suffers a decline of just over 10\%, specifically 0.51 to 0.45 (Spearman) and 0.49 to 0.44 (Pearson), while correlation to human majority incurs a more modest decline (0.44 to 0.43 and 0.43 to 0.41 for Spearman and Pearson).

\section{Additional Details and Results for Mitigating Hallucination in GSR}
\label{sec:add-results}

{
\setlength{\tabcolsep}{0.04in}
\fontsize{9}{10}\selectfont
\aboverulesep = 0mm
\belowrulesep = 0mm %
\begin{table*}[t]
    \centering
    \caption{
    \textbf{Complete SR Results.}
        This Table %
        acts as a more complete companion to Table \ref{tab:results_injection} of the main paper, with additional baselines and variations included. 
        We see that Bicubic has the fewest hallucinations (highest HS), which is unsurprising as the method cannot invent new details, with Swin2SR, which focuses on regression (rather than generation), following closely.
        Among the new diffusion models, PiSA tends to obtain a good tradeoff between perceptual quality, fidelity, and hallucinations.
        Our main comparisons are with SeeSR and PASD, versus our modifications via AlignProp.
        We see that the base model tends to have the best pixel-level fidelity (PSNR), but our method improves upon it in every other aspect.
        The CLIP-based variations of our method (chosen because CLIP also has a strong correlation to HS) show good performance, often trading off with our DINO-HS-based approach on the various metrics.
        However, our method using DINO-HS has superior performance in terms of hallucinations, according to all three HS metrics in almost every scenario, without degrading other metrics.
    }
    \begin{adjustbox}{max width=\linewidth}
    \begin{tabular}{ll|ccccccccccc}
    \toprule
      & Model & PSNR $\uparrow$ & SSIM $\uparrow$ & LPIPS $\downarrow$ & DISTS $\downarrow$ & MUSIQ $\uparrow$ & CLIPIQA $\uparrow$ & QAlign $\uparrow$ &  Sharpness $\uparrow$ & GPT-HS $\uparrow$ & Qwen-HS $\uparrow$ & \dinohs $\uparrow$ \\\hline
      \multirow{16}{*}{SS-TS} 
      & Bicubic & 25.04 & 0.634 & 0.704 & 0.337 & 19.86 & 0.312 & 1.15 & 0.90 & \textbf{4.67} & \textbf{3.30} & \textbf{3.67} \\
      & Swin2SR \cite{conde2022swin2sr} & \textbf{25.75} & \textbf{0.681} & 0.473 & 0.295 & 44.37 & 0.299 & 2.20 & 6.57 & 3.38 & 3.17 & 3.39 \\
      & RealESRGAN \cite{wang2021real} & 24.04 & 0.631 & 0.313 & 0.212 & 62.22 & 0.547 & 3.35 & 73.02 & 2.78 & 2.84 & 2.86 \\
      & StableSR \cite{wang2024exploiting} & 23.26 & 0.573 & {0.311} & {0.205} & {65.92} & {0.677} & {3.53} & {105.01} & 3.36 & 3.00 & 3.33\\
      & PiSA \cite{sun2025pixel} & 23.87 & 0.606 & \textbf{0.282} & \textbf{0.193} & \textbf{69.68} & {0.693} & \textbf{3.88} & 73.29 & 3.58 & 3.23 & 3.60\\
      & SUPIR \cite{yu2024scaling} & 23.15 & 0.544 & 0.364 & 0.226 & 62.59 & \textbf{0.705} & 3.78 & \textbf{177.76} & 3.24 & 2.88 & 3.24 \\
      & FaithDiff \cite{chen2025faithdiff} & 23.49 & 0.581 & 0.312 & 0.199 & 69.26 & 0.646 & 3.77 & 79.09 & 2.93 & 2.96 & 3.28 \\
      & DiT4SR \cite{duan2025dit4sr} & 21.77 & 0.548 & 0.345 & 0.211 & 68.09 & 0.664 & 3.72 & 142.04 & 2.54 & 2.64 & 3.17 \\
      \cdashline{2-13}\noalign{\vskip 0.5ex}
      & SeeSR \cite{wu2024seesr} & \textbf{23.68} & {0.604} & 0.319 & 0.197 & 68.67 & 0.694 &3.98 & 84.01 & 2.99 & 2.77 & 3.17 \\
      & \quad+DINO-HS+MUSIQ &  23.23 & 0.595 & \textbf{0.252} & \textbf{0.185} & {70.49} & {0.743} & {3.98} & {135.99} & \textbf{3.87} & \textbf{3.46} & \textbf{3.99} \\
      & \quad+CLIP-ST+MUSIQ & 22.72 & \textbf{0.608} & 0.272 & \textbf{0.185} & \textbf{71.30} & \textbf{0.746} & \textbf{4.22} & \textbf{153.01} & 3.57 & 3.26 & 3.81\\
      & \quad+CLIP-CLS+MUSIQ & 22.48 & 0.601 & 0.292 & 0.189 & 68.73 & 0.684 & 3.94 & 151.27 & 3.54 & 3.17 & 3.63 \\
      
      \cdashline{2-13}\noalign{\vskip 0.5ex}
      & PASD \cite{yang2024pasd} & \textbf{23.55} & {0.598} & 0.369 & 0.214 & 65.54 & 0.635 & 3.75 & 82.59 & 2.54 & 2.42 & 2.48 \\
      & \quad+DINO-HS+MUSIQ & 22.69 & 0.579 & \textbf{0.262} & \textbf{0.186} & \textbf{69.52} & \textbf{0.746} & {3.84} & {175.71} & \textbf{3.83}  & \textbf{3.36} & \textbf{3.90} \\
      & \quad+CLIP-ST+MUSIQ & 22.97 & \textbf{0.614} & 0.273 & \textbf{0.186} & 69.06 & 0.703 & \textbf{3.87} & 125.96 & 3.53 & 3.28 & 3.70 \\
      & \quad+CLIP-CLS+MUSIQ & 21.82 & 0.579 & 0.293 & 0.188 & 66.37 & 0.704 & 3.72 & \textbf{202.98} & 3.57 & 3.16 & 3.59 \\
    \hline\noalign{\vskip 0.5ex}

      \multirow{16}{*}{RealSR}
        & Bicubic & 27.11 & 0.756 & 0.456 & 0.263 & 25.81 & 0.310 & 1.66 & 0.95 & \textbf{4.56} & \textbf{3.63} & \textbf{3.98} \\
        & Swin2SR \cite{conde2022swin2sr} & \textbf{27.29} & \textbf{0.801} & 0.291 & 0.237 & 53.14 & 0.303 & 2.51 & 13.26 & 3.57 & 3.13 & 3.46 \\
        & RealESRGAN \cite{wang2021real} & 25.58 & 0.759 & 0.272 & 0.207 & 60.61 & 0.450 & 3.11 & 48.99 & 2.96 & 2.69 & 2.96 \\
        & StableSR \cite{wang2024exploiting} & 24.65 & 0.708 & 0.300 & 0.214 & 65.88 & 0.623 & 3.28 & {75.74} & 3.22 & 2.68 & 3.31 \\
        & PiSA \cite{sun2025pixel} & 25.50 & 0.742 & \textbf{0.267} & \textbf{0.204} & \textbf{70.14} & \textbf{0.669} & \textbf{3.63} & 51.53 & 3.11 & 2.92 & 3.47 \\
        & SUPIR \cite{yu2024scaling} & 25.09 & 0.674 & 0.374 & 0.250 & 57.60 & 0.623 & 3.32 & 92.59 & 3.33 & 2.87 & 3.48 \\
        & FaithDiff \cite{chen2025faithdiff} & 25.27 & 0.708 & 0.287 & 0.211 & 68.83 & 0.610 & 3.56 & 71.98 & 2.90 & 2.85 & 3.23\\
        & DiT4SR \cite{duan2025dit4sr} & 23.40 & 0.660 & 0.328 & 0.226 & 67.79 & 0.640 & 3.40 & \textbf{102.71} & 2.79 & 2.67 & 3.33 \\
        \cdashline{2-13}\noalign{\vskip 0.5ex}
        
        & SeeSR \cite{wu2024seesr} & \textbf{25.15} & {0.721} & 0.301 & 0.223 & 69.81 & 0.670 & {3.72} & 86.99 & 2.92 & 2.60 & 3.13 \\
        & \quad+DINO-HS+MUSIQ & 23.98 & 0.718 & \textbf{0.278} & \textbf{0.200} & {70.13} & \textbf{0.729} & 3.68 & {106.23} & \textbf{3.45} & \textbf{3.10} & \textbf{3.88} \\
        & \quad+CLIP-ST+MUSIQ & 22.79 & 0.718 & 0.281 & 0.211 & \textbf{70.67} & {0.710} & \textbf{3.93} & \textbf{135.30} & 3.30 & 2.91 & 3.66\\
        & \quad+CLIP-CLS+MUSIQ & 23.22 & \textbf{0.723} & 0.285 & 0.223 & 68.57 & 0.672 & 3.75 & 129.98 & 3.26 & 2.92 & 3.51 \\
        \cdashline{2-13}\noalign{\vskip 0.5ex}
        
        & PASD \cite{yang2024pasd} & \textbf{25.75} & {0.735} & 0.296 & 0.213 & 62.52 & 0.534 & 3.30 & 43.47 & 2.89 & 2.52 & 2.81\\
        & \quad+DINO-HS+MUSIQ & 23.62 & 0.716 & {0.269} & {0.197} & \textbf{69.47} & \textbf{0.719} & {3.59} & {104.88} & \textbf{3.62} & {2.99} & \textbf{3.71}
        \\
        & \quad+CLIP-ST+MUSIQ & 24.14 & \textbf{0.748} & \textbf{0.253} & \textbf{0.194} & 67.68 & 0.643 & 3.59 & 66.06 & 3.44 & \textbf{3.05} & 3.59 \\
        & \quad+CLIP-CLS+MUSIQ & 22.41 & 0.697 & 0.288 & 0.215 & 67.31 & 0.682 & \textbf{3.62} & \textbf{132.26} & 3.17 & 2.77 & 3.38 \\
        \hline\noalign{\vskip 0.5ex}

     \multirow{16}{*}{DRealSR} 
        & Bicubic & \textbf{30.54} & 0.830 & 0.461 & 0.279 & 22.59 & 0.319 & 1.47 & 0.38 & \textbf{4.76} & \textbf{3.95} & \textbf{4.14} \\
        & Swin2SR \cite{conde2022swin2sr} & 29.98 & \textbf{0.843} & 0.330 & 0.251 & 43.58 & 0.325 & 2.23 & 4.07 & 3.68 & 3.69 & 3.63 \\
        & RealESRGAN \cite{wang2021real} & 28.40 & 0.801 & \textbf{0.286} & \textbf{0.211} & 54.87 & 0.454 & 2.91 & 27.07 &3.27 & 3.41 & 3.23 \\
        & StableSR \cite{wang2024exploiting} & 28.03 & 0.754 & 0.328 & 0.227 & 58.51 & 0.636 & 3.06 & {40.08} & 3.51 & 3.41 & 3.45 \\
        & PiSA \cite{sun2025pixel} & 28.31 & 0.780 & 0.296 & 0.217 & {66.10} & \textbf{0.697} & \textbf{3.58} & 30.66 & 3.62 & 3.60 & 3.59 \\
        & SUPIR \cite{yu2024scaling} & 26.78 & 0.668 & 0.434 & 0.278 & 54.49 & 0.630 & 3.20 & \textbf{71.88} & 3.28 & 3.23 & 3.43\\
        & FaithDiff \cite{chen2025faithdiff} & 27.23 & 0.707 & 0.356 & 0.242 & \textbf{66.11} & 0.635 & 3.44 & 47.74 & 2.84 & 3.00 & 3.13\\
        & DiT4SR \cite{duan2025dit4sr} &25.63 & 0.676 & 0.371 & 0.250 & 64.94 & 0.663 & 3.39 & 70.28 & 2.86 & 3.17 & 3.27 \\
        \cdashline{2-13}\noalign{\vskip 0.5ex}
        
        & SeeSR \cite{wu2024seesr} & \textbf{28.07} & \textbf{0.768} & {0.317} & 0.232 & 65.09 & 0.691 & {3.59} & 48.21 & 3.11 & 3.14 & 3.15 \\
        & \quad+DINO-HS+MUSIQ & 26.52 & 0.739 & 0.326 & \textbf{0.221} & {65.19} & \textbf{0.742} & 3.52 & {55.36} & \textbf{3.80} & \textbf{3.65} & \textbf{3.86}
        \\
        & \quad+CLIP-ST+MUSIQ & 25.50 & 0.752 & 0.313 & 0.226 & \textbf{67.31} & {0.739} & \textbf{3.82} & \textbf{67.44} & 3.44 & 3.53 & 3.66\\
        & \quad+CLIP-CLS+MUSIQ & 25.78 & 0.756 & \textbf{0.307} & {0.224} & 63.47 & 0.674 & 3.57 & 65.37 & 3.77 & 3.35 & 3.61\\
        \cdashline{2-13}\noalign{\vskip 0.5ex}
        
        & PASD \cite{yang2024pasd} & \textbf{28.05} & \textbf{0.779} & { 0.319} & 0.230 & 58.48 & 0.572 & 3.27 & 29.66 & 2.72 & 2.85 & 2.70 \\
        & \quad+DINO-HS+MUSIQ & 25.10 & 0.719 & 0.328 & {0.227} & \textbf{65.04} & \textbf{0.729} & {3.41} & {58.42} & \textbf{3.74} & \textbf{3.57} & \textbf{3.75}\\
        & \quad+CLIP-ST+MUSIQ & 25.59 & 0.759 & \textbf{0.291} & \textbf{0.214} & {64.06} & 0.685 & \textbf{3.53} & 42.31 & 3.58 & 3.56 & 3.63 \\
        & \quad+CLIP-CLS+MUSIQ & 24.74 & 0.732 & 0.314 & 0.229 & 58.63 & 0.654 & 3.25 & \textbf{64.90} & 3.44 & 3.39 & 3.59\\
      \bottomrule   
    \end{tabular}
    \end{adjustbox}

    \label{tab:results_injection_full}
\end{table*}
}

\begin{figure*}[t]
    \centering
    \includegraphics[width=0.99\textwidth]{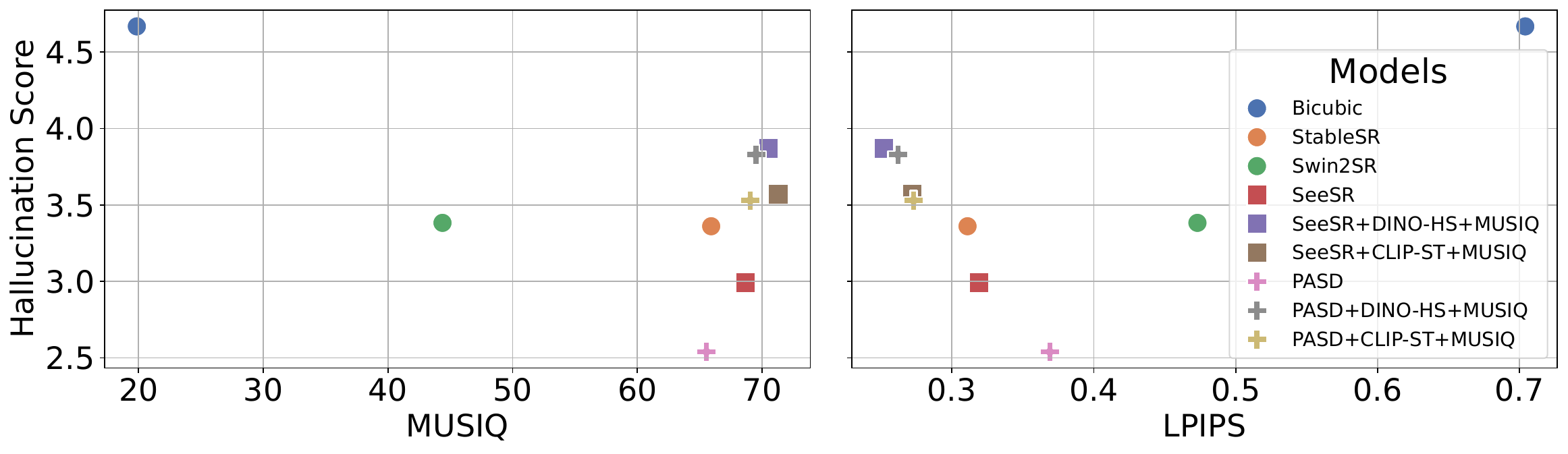}
    \caption{
    \textbf{HS and Perceptual Quality}. We compare methods along HS and Perceptual Quality (MUSIQ, LPIPS) measures on SS-TS dataset. Base models and their aligned variants for SeeSR and PASD are depicted with square (``$\square$") and plus (``$+$") shapes respectively. We observe our aligned variants (using both DINO-HS and CLIP), compared to their base models, improve HS (y-axis) without damaging or even improving over perceptual (LPIPS) and perceived (MUSIQ) quality (x-axis). 
    }
    \label{fig:plot-metric-vs-gpt}
\end{figure*}

\paragraph{Implementation details.}
We use the AlignProp implementation in TRL library from Hugging Face. 
We adapted the code to include diffusion-based GSR pre-trained models with their default configurations obtained from their codebase, which includes SeeSR and PASD. These configurations include the \textit{choice of sampler} (DDIM for SeeSR; UniPC~\cite{zhao2023unipc} for PASD), \textit{prompt extractors from LRI} (degradation-aware tags for SeeSR; captions trained on CoCa for PASD), \textit{added positive} (\texttt{clean, high-resolution, 8k}) and \textit{negative prompts}, and \textit{hyper parameters} including sampling steps (50 for SeeSR; 20 for PASD) and classifier-free guidance weight (5.5 for SeeSR; 9.0 for PASD). Overall, the use of two different model design choices underscores the effectiveness of our proposed reward models within the gradient back-propagation framework used in this paper.

The experiments were performed with one A100 GPU with 80G high-bandwidth memory. We train all the models for $200$ steps using a batch size of 8 with gradient accumulation steps of 4 (effective batch size of $8 \times 4 = 32$), and a learning rate of $1e^{-3}$ with Adam optimizer.

Regarding memory usage, the AlignProp process on SeeSR occupies $\sim$56G of GPU memory. In terms of GPU-hours, the aforementioned fine-tuning for one epoch (which is the default in our paper) takes $\sim$9 hours (on a single A100).

\noindent
\textbf{CLIP and DINO Feature Extraction Details.}

\noindent
$\,\,$ \textit{+DINO-ST+MUSIQ}: we use pretrained DINOv2 ViT-B/14 model with registers \cite{oquab2023dinov2,darcet2023vision}, and form $g$ as the concatenated spatial tokens from intermediate layers with indices $1,3,5,7,9,11$; with $\lambda$ as $0.05$

\noindent
$\,\,$ \textit{+CLIP-ST/CLS+MUSIQ}: we use pretrained OpenCLIP (ViT-B/16) \cite{cherti2023reproducible}, and form $g$ as the concatenated spatial tokens from intermediate layers (same as above) for \textit{CLIP-ST}, and CLS token from the last layer for \textit{CLIP-CLS}; with $\lambda$ as $0.1$ and $0.05$ respectively.

\paragraph{Dataset.}
In addition to \S\ref{sec:results} of the main paper, here we provide more details on the dataset used for AlignProp training. We generate synthetic LRI-GTI pairs from the DIV-2K~\cite{div2k}, DIV-8K~\cite{gu2019div8k}, and Flickr-2K~\cite{agustsson2017ntire} datasets. Specifically, we randomly crop 512$\times$512 images (or GTI) from the original images, and apply Real-ESRGAN \cite{wang2021real} degradations to obtain LRI. We set the degradation level to be the same as StableSR \cite{pan2021exploiting}. In total, we generate 6550 LRI-GTI pairs, with 2400 from DIV-2K, 1500 from DIV-8K, and 2650 from Flickr-2K dataset. We use a random held-out set of 100 images for validation.

\paragraph{Complete SR results.} In addition to the performance on SS-TS and RealSR datasets reported in Table \ref{tab:results_injection} of the main paper, we provide complete results along with performance on DRealSR in Table \ref{tab:results_injection_full}. Across all the three datasets (one synthetic and two real-world), our aligned models improve on HS while maintaining perceived quality (MUSIQ, Sharpness), without damaging or even improving perceptual quality (LPIPS, DISTS).

We further highlight the results along perceptual quality measures in Fig. \ref{fig:plot-metric-vs-gpt}. We plot performance of base models and their aligned variants for SeeSR and PASD with square (``$\square$") and plus (``$+$") shapes respectively. We observe our aligned variants (using both DINO-HS and CLIP) improve over HS (y-axis) while not damaging or even improving over perceptual (LPIPS) and perceived (MUSIQ) quality (x-axis).

\begin{table}[htb]
\centering
\caption{\textbf{Ablation Study on the Choices of CLIP Layers and Impact of MUSIQ Factors}.
As in Table \ref{tab:ablation-combined}, we look at 
architectural variations (\textit{last} vs.\ \textit{interm})
and loss weight changes (strength of the MUSIQ weight $\lambda$), but with our CLIP-based approach, instead of DINO-HS.
We encounter similar results: 
(i) using \textit{last} instead of \textit{interm} improves HS, but causes a collapse in quality (MUSIQ);
(ii) we can control the tradeoff between HS and MUSIQ by varying the MUSIQ-based regularization strength ($\lambda$);
and (iii) the presence of the MUSIQ penalty tends to improve LPIPS at the expense of PSNR. 
}
\label{tab:ablation-combined-clip}
    \begin{adjustbox}{max width=0.99\linewidth}
\begin{tabular}{@{}lcccccc@{}}
\toprule
\multirow{2}{*}{Metric} & \multirow{2}{*}{SeeSR} & \multicolumn{2}{c}{+ CLIP-ST} & \multicolumn{3}{c}{+ CLIP-ST interm + $\lambda \cdot $MUSIQ} \\ 
\cmidrule(l){3-4} \cmidrule(l){5-7}  
 &  & last & interm & $\lambda\mathord{=}0.2$ & $\lambda\mathord{=}0.1$ & $\lambda\mathord{=}0.05$ \\ 
 \midrule
PSNR $\uparrow$ & 23.68 & 25.22 & 23.95 & 23.15 & 22.72 & 23.90 \\
LPIPS $\downarrow$ & 0.319 & 0.367 & 0.303 & 0.274 & 0.272 & 0.267  \\
MUSIQ $\uparrow$ & 68.67 & 9.07 & 33.25 & 71.90 & 71.30 & 64.78 \\
GPT-HS $\uparrow$ & 2.99 & 4.05 & 3.88 & 3.60 & 3.57 & 3.77  \\ \bottomrule
\end{tabular}%
\end{adjustbox}
\end{table}

\paragraph{Ablations and Variations.}
In addition to Table \ref{tab:ablation-combined} in the main paper, which shows ablations with SeeSR and our HS proxy variants, we considered a series of alternatives, including different reward models and variations thereof.

\noindent$\,\bullet\,$\textit{CLIP-based Reward.}
In addition to including results with CLIP in Table \ref{tab:results_injection_full}, we show more CLIP-aligned variants in Table \ref{tab:ablation-combined-clip}.
We observe similar trends, where (i) intermediate layers (\texttt{interm}) results in higher perceptual (LPIPS) and perceived (MUSIQ) quality compared to last layer only (\texttt{last}), with a trade-off between fidelity, quality and HS; and (ii) higher MUSIQ factors ($\lambda$) leads to higher perceived quality (MUSIQ).

\noindent$\,\bullet\,$\textit{LPIPS-based reward.}
We also considered using LPIPS as the basis of our reward for fine-tuning.
Results on the SS-TS dataset are shown in Table~\ref{supp:table:lpipsablation}.
We see that the resulting LPIPS-based model cannot improve HS effectively
(compared to fine-tuning with DINO-HS; see also Table~\ref{tab:results_injection}).
This may not be surprising, given that LPIPS correlates far less with HS (human or MLLM-based) than DINO or DINO-HS.

\newcommand{\textbfc}[1]{\textbf{#1}}

\begin{table*}[h!]
\centering
\caption{
    \textbf{Replacing our deep features proxy HS estimator with LPIPS.} 
    All values are computed on the SS-TS test set.
    As in our standard case, to maintain comparability, we use an additional MUSIQ term with the LPIPS reward.
    While LPIPS as a reward generally does well, it is not able to effectively improve HS compared to fine-tuning with DINO-HS.
}
    \begin{adjustbox}{max width=0.93\linewidth}
\begin{tabular}{lccccccccc}
\hline
\textbf{Model} & \textbf{PSNR} $\uparrow$ & \textbf{SSIM} $\uparrow$ & \textbf{LPIPS} $\downarrow$ & \textbf{DISTS} $\downarrow$ & \textbf{MUSIQ} $\uparrow$ & \textbf{CLIPIQA} $\uparrow$ & \textbf{QAlign} $\uparrow$ & \textbf{Sharpness} $\uparrow$ & \textbf{GPT-HS} $\uparrow$ \\
\hline
SeeSR & \textbfc{23.68} & \textbfc{0.604} & 0.319 & 0.197 & 68.67 & 0.694 & 3.98 & 84.01 & 2.99 \\ 
    \quad+LPIPS+MUSIQ & 23.66 & 0.602 & \textbfc{0.248} & 0.199 & \textbfc{71.49} & 0.710 & \textbfc{3.99} & 101.13 & 3.32 \\
    \quad+DINO-HS+MUSIQ (default) &  23.23 & 0.595 & {0.252} & \textbf{0.185} & {70.49} & \textbf{0.743} & {3.98} & \textbf{135.99} & \textbf{3.87} \\ %
\cdashline{1-10}\noalign{\vskip 0.5ex}
PASD & \textbfc{23.55} & 0.598 & 0.369 & 0.214 & 65.54 & 0.635 & 3.75 & 82.59 & 2.54 \\
\quad+LPIPS+MUSIQ & 22.92 & \textbfc{0.599} & \textbfc{0.257} & 0.195 & \textbfc{71.83} & 0.735 & \textbfc{3.94} & 119.68 & 3.22 \\
\quad+DINO-HS+MUSIQ (default) & 22.69 & 0.579 & {0.262} & \textbf{0.186} & {69.52} & \textbf{0.746} & {3.84} & \textbf{175.71} & \textbf{3.83} \\ %
\hline
\end{tabular}
\end{adjustbox}
\label{supp:table:lpipsablation}
\end{table*}

\noindent$\,\bullet\,$\textit{Number of steps.}
In Table \ref{supp:table:numsteps}, 
we show the results of halving or doubling the number of fine-tuning steps used in AlignProp-based training.
We find that halving the number of steps lowers HS without improving other metrics, suggesting under-training.
In contrast, doubling the number of steps further improves HS, but at the expense of perceptual quality (\ie, NR-IQA scores), in addition, of course, to a significant increase in training time.
We therefore suggest our default settings as a good balance between reducing hallucinations, maintaining (or improving) realism, and computational time cost.

\begin{table*}[h!]
\centering
\caption{
    \textbf{Number of steps.} 
    We consider halving and doubling the training time of our fine-tuning approach.
    Compared to the default mode, which sees 6.4K samples, these variations see 3.2K and 12.8K, respectively. 
    We evaluate with SeeSR on the SS-TS, using our reward based on DINO-HS and MUSIQ.
    We see that decreasing the number of steps leads to slightly lower HS values.
    On the other hand, while doubling the number of steps increases HS, it does so at the expense of several NR-IQA metrics.
    We therefore suggest our default setting as a good balance between HS, NR-IQA, and training time.
}
    \begin{adjustbox}{max width=0.99\linewidth}
\begin{tabular}{lccccccccccc}
\hline
Model & PSNR $\uparrow$ & SSIM $\uparrow$ & LPIPS $\downarrow$ & DISTS $\downarrow$ & MUSIQ $\uparrow$ & CLIPIQA $\uparrow$ & QAlign $\uparrow$ &  Sharpness $\uparrow$ & GPT-HS $\uparrow$ & Qwen-HS $\uparrow$ & \dinohs $\uparrow$ \\\hline
SeeSR \cite{wu2024seesr} & \textbf{23.68} & {0.604} & 0.319 & 0.197 & 68.67 & 0.694 & \textbf{3.98} & 84.01 & 2.99 & 2.77 & 3.17 \\
    \quad+DINO-HS+MUSIQ ($1/2\times$ steps) %
    & 23.34 & \textbf{0.611} & 0.260 & 0.186 & 69.26 & 0.728 & 3.96 & 117.90 & 3.84 & 3.40 & 3.90 \\
    \quad+DINO-HS+MUSIQ (default) %
    &  23.23 & 0.595 & {0.252} & {0.185} & \textbf{70.49} & \textbf{0.743} & \textbf{3.98} & \textbf{135.99} & 3.87 & {3.46} & {3.99} \\
    \quad+DINO-HS+MUSIQ ($2\times$ steps) %
    & 23.40 & 0.602 & \textbf{0.247} & \textbf{ 0.183} & 69.86 & 0.725 & 3.92 & 117.07 & \textbf{3.98} & \textbf{3.49 }& \textbf{4.04} \\
    
\hline
\end{tabular}
\end{adjustbox}
\label{supp:table:numsteps}
\end{table*}

\begin{figure*}
    \centering
    \includegraphics[width=0.99\textwidth]{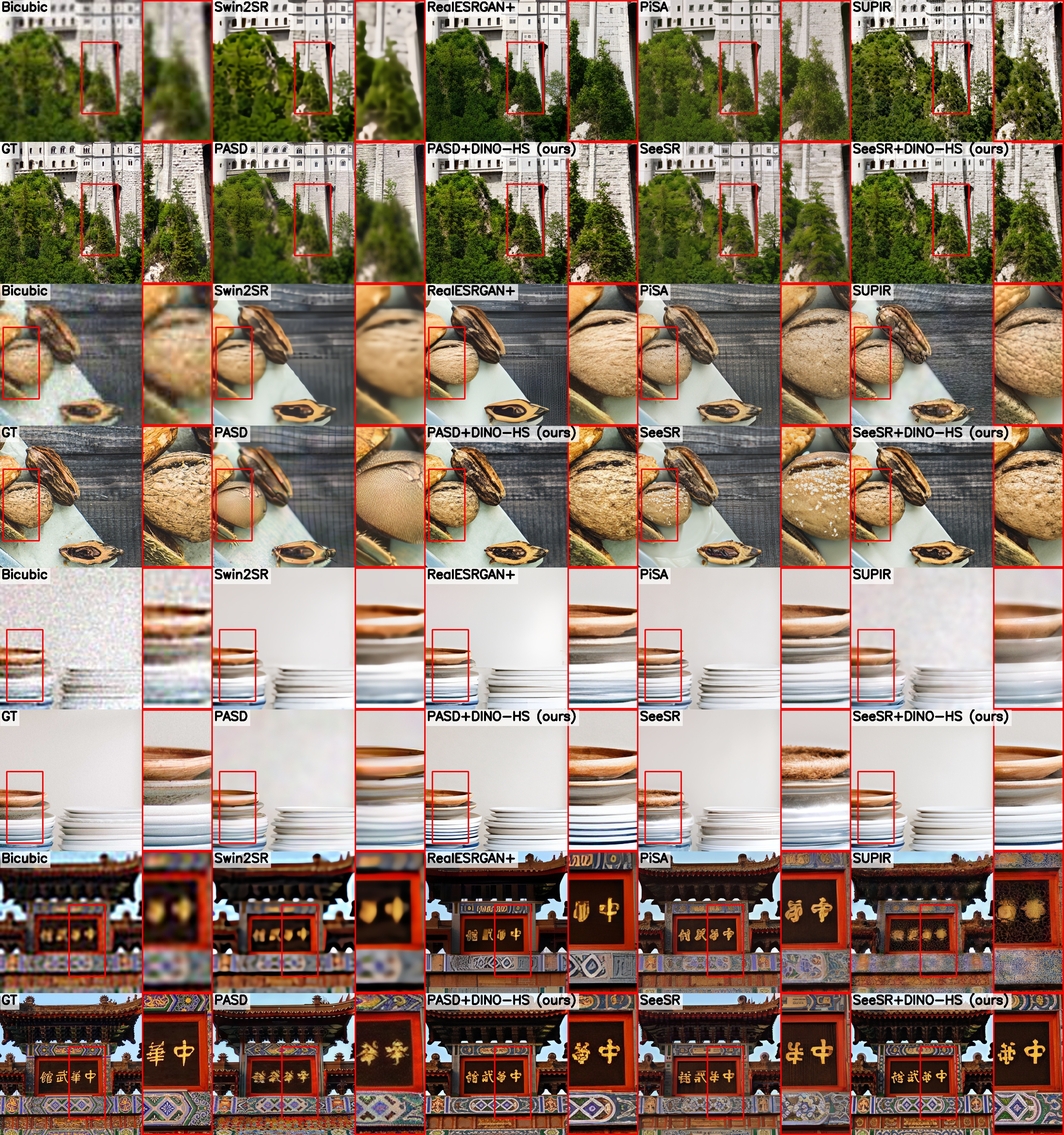}
    \caption{Additional comparative results (I). Note the primary point of comparison is the base model (SeeSR or PASD) versus our fine-tuned version (SeeSR/PASD+DINO-HS), but we provide other models for reference as well. 
    In general, we see that our altered models tend to have more realistic textures and fewer extreme semantic errors. 
    For example, in the first image-set, we see that both the trees and the stone wall in our outputs are far more similar to the GT (versus the base models), without sacrificing image quality. 
    In image-set two, our fine-tuning reduces the severe semantic (PASD) and textural (SeeSR) errors in the appearance of the nut, with image-set three shows similar improvements.
    Finally, the last two rows show a difficult image involving Chinese characters: while no method obtains the fully correct details, our models have greater fidelity to both the symbols and the diamond-shaped pattern underneath,
    while again maintaining realism.
    See also Fig.~\ref{fig:supp:sstsresults1}.
    }
    \label{fig:supp:sstsresults0}
\end{figure*}

\begin{figure*}
    \centering
\includegraphics[width=0.99\textwidth]{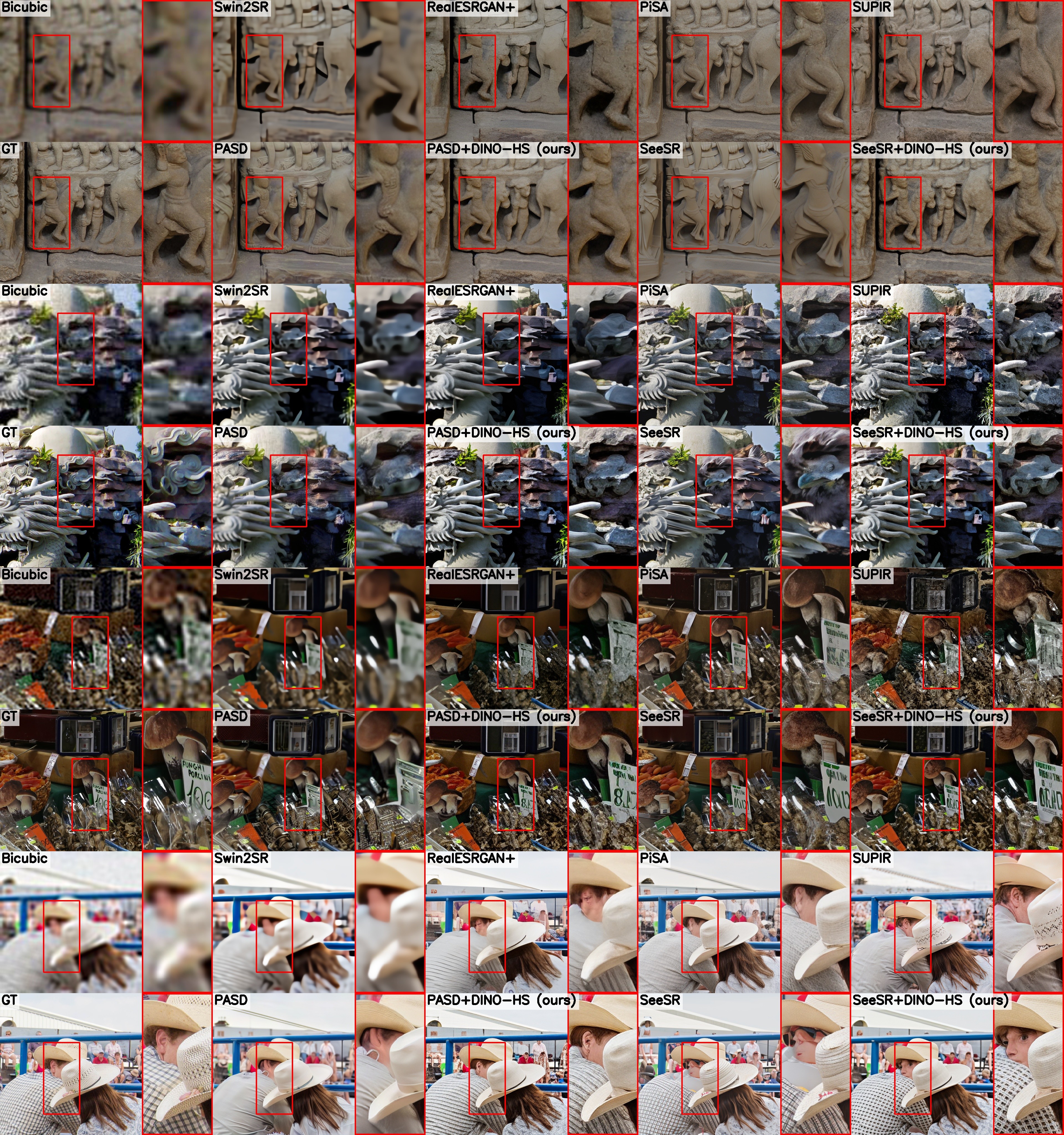}
    \caption{Additional comparative results (II). Note the primary point of comparison is the base model (SeeSR or PASD) versus our fine-tuned version (SeeSR/PASD+DINO-HS), but we provide other models for reference as well. 
    In general, we see that our altered models tend to have more realistic textures and fewer extreme semantic errors. 
    For instance, the appearance of the stone in image-set one of our HS-corrected methods is more faithful, while in image-set two our methods fix oversmoothing (PASD) and dramatic semantic errors (SeeSR).  
    The last row shows a failure case, where our method applied to SeeSR is unable to fix the mistaken human pose from the original model.
    See also Fig.~\ref{fig:supp:sstsresults0}.
    }
    \label{fig:supp:sstsresults1}
\end{figure*}

\paragraph{Qualitative results.}
We provide more qualitative results from our aligned models (both SeeSR and PASD) in Figs.~\ref{fig:supp:sstsresults0} and \ref{fig:supp:sstsresults1}, along with a suite of baselines ranging from powerful perception-oriented diffusion models (SUPIR \cite{yu2024scaling} and PiSA \cite{sun2024pisasr}) to more distortion-oriented single-pass models (Swin2SR \cite{conde2022swin2sr} and RealESRGAN+ \cite{wang2021real}). 
We observe that DINO-HS fine-tuning is often able to reduce mistakes in the 
semantics (\eg, Fig.~\ref{fig:supp:sstsresults0}, second and third image-sets; Fig.~\ref{fig:supp:sstsresults1}, second image-set)
and 
repair poor mid-level textural fidelity 
(\eg, the first image-set of both Fig.~\ref{fig:supp:sstsresults0} and Fig.~\ref{fig:supp:sstsresults1})
yet maintains perceptual quality, sharpness, and realism.

\section{Additional Explanatory Remarks}
\label{supp:extraremarks}
In this section, we provide additional remarks about HS and our reward-based fine-tuning, for which we had insufficient room in the main paper.

\noindent
\textbf{How is HS different from existing IQA?}
\noindent
Let us consider the FR IQA case first. When a reference is available, it would seem that we can simply use an existing FR metric to determine which GSR model is more hallucinatory.
However, we suggest this may hold only for artifacts that FR-IQA methods are trained to detect. For example, LPIPS and DISTS are sensitive to mid-level distortions, like textural changes, but miss semantic alterations. Conversely, high-level features like CLIP may overlook subtle issues (e.g., nonsense symbols replacing text). As shown in Fig.~\ref{fig:mllm_sample_outputs_seesr_supp}, the MLLM detects incorrect text on signs in SR images -- something models like DINO may miss. Finally, low-level FR metrics like SSIM are too sensitive, picking up simple blur (which usually does not qualify as a realistic hallucination under our definition) or plausible but not pixel-perfect outputs (e.g., even slightly shifting the images can immensely impact such metrics). 
Regardless, we do find that the existing approaches best correlated to GPT-based HS (and human scoring) are based on FR deep feature distances, which are much more semantics-aware.

The NR-IQA case is more easily seen to be orthogonal. Indeed, we find that MUSIQ and sharpness are \textit{negatively} correlated to HS (as well as human judgments), because they reward realism, even if the result is completely implausible with respect to the LQ or semantically mutated compared to the GT.

\noindent
\textbf{What is the Role of Saliency?}
\noindent
One potentially unintuitive aspect of hallucinations is the role of saliency. Consider the case of artifacts in non-salient regions, where people are less likely to notice the errors. For instance, consider severe alterations to background vegetation - here, severe can mean both semantic (new branches or wrong plants) and in terms of pixel distances.
By our definition of hallucinations in SRIs (\S\ref{sec:def}), new textural details that a human observer \textit{would not notice} as out-of-place are considered to be low hallucinatory. Importantly, our definition of hallucination is orthogonal to general image quality: degradation in vegetation regions may be very severe if considered as a generic type of artifact (e.g., as noise, it could be considered severe, as measured by PSNR or classifier error), but it might not be severe as a \textit{hallucination} (if it is not perceptually noticeable).

For this reason, notice that HS can be impacted by cropping or field-of-view, as the image is evaluated holistically in its full context (just as human judges do).
Since salient regions in a crop can sometimes become non-salient when considered in a larger image, it is potentially possible for a low HS crop to reside in a larger image with a high HS, and for this to align with human judgments as well.
We leave investigations of such possibilities to future work.

\noindent
\textbf{Does HS care about localized artifacts?}
Since the MLLM has access to full image and we output a global score, it may not be immediately obvious that artifacts localized to small regions will appropriately affect the HS output.
However, in our evaluation setting, each HS score is accompanied by a detailed reasoning response from GPT, indicating why that specific HS score is given to the SRI. We can see how and why localized artifacts affect HS via this explanation.
For instance, as shown in the first example of Fig.~\ref{fig:mllm_sample_outputs_stablesr_supp}, HS identifies the SRI ``altering the content of the shirts with different logos and text compared to the GT image'' and rates the image with a score of 1. Assuming the reasoning reflects the underlying logic determining the score, this suggests that the model is able to assess smaller local regions in the image (\eg, the logo region) to determine the final HS.

\noindent
\textbf{Why utilize MUSIQ in the reward, when it anticorrelates to HS?}
We note that MUSIQ is trained to align with human judgments of technical and aesthetic quality on datasets where blur is treated as a defect, so it tends to score sharper images higher. In contrast, our paper shows that HS correlates better with metrics such as PSNR, and prefers more conservative and blurry results (e.g., in Table \ref{tab:results_injection}, bicubic upsampling has the highest HS). As a result, MUSIQ exhibits a negative correlation with HS and human mean ratings. We also observe this sharpness-hallucination tradeoff in our ablation study: according to Tables \ref{tab:ablation-combined} and \ref{tab:ablation-combined-clip}, performing reward-backpropagation using DINO/CLIP alone (without MUSIQ) leads to degraded perceptual/perceived quality (sharpness), yet higher HS. 
Ideally, we would sacrifice as little image quality as possible, while reducing hallucinations.
Indeed, if we try to optimize HS in isolation, we may end up with excessively blurry outputs (similar to, e.g., bicubic upsampling).
On the other hand, increasing the weight of the MUSIQ reward improves perceptual sharpness, but can harm HS. Based on these findings, we propose to combine our HS proxy reward with MUSIQ, which stems the deterioration in realism, allowing us to strike a balance between perceptual quality and hallucination degree.
Of course, our fine-tuning approach is agnostic to the exact choice of NR-IQA model used for this quality preservation regularizer (though MUSIQ has been shown to perform well for SR \cite{zhang2025augmenting}); 
hence, as NR-IQA models improve over time, we can apply such advances to our method as well.

\section{More Example Outputs from GPT}
\label{supp:qualitative}
\begin{figure*}[t]
    \centering
    \includegraphics[width=0.99\textwidth]{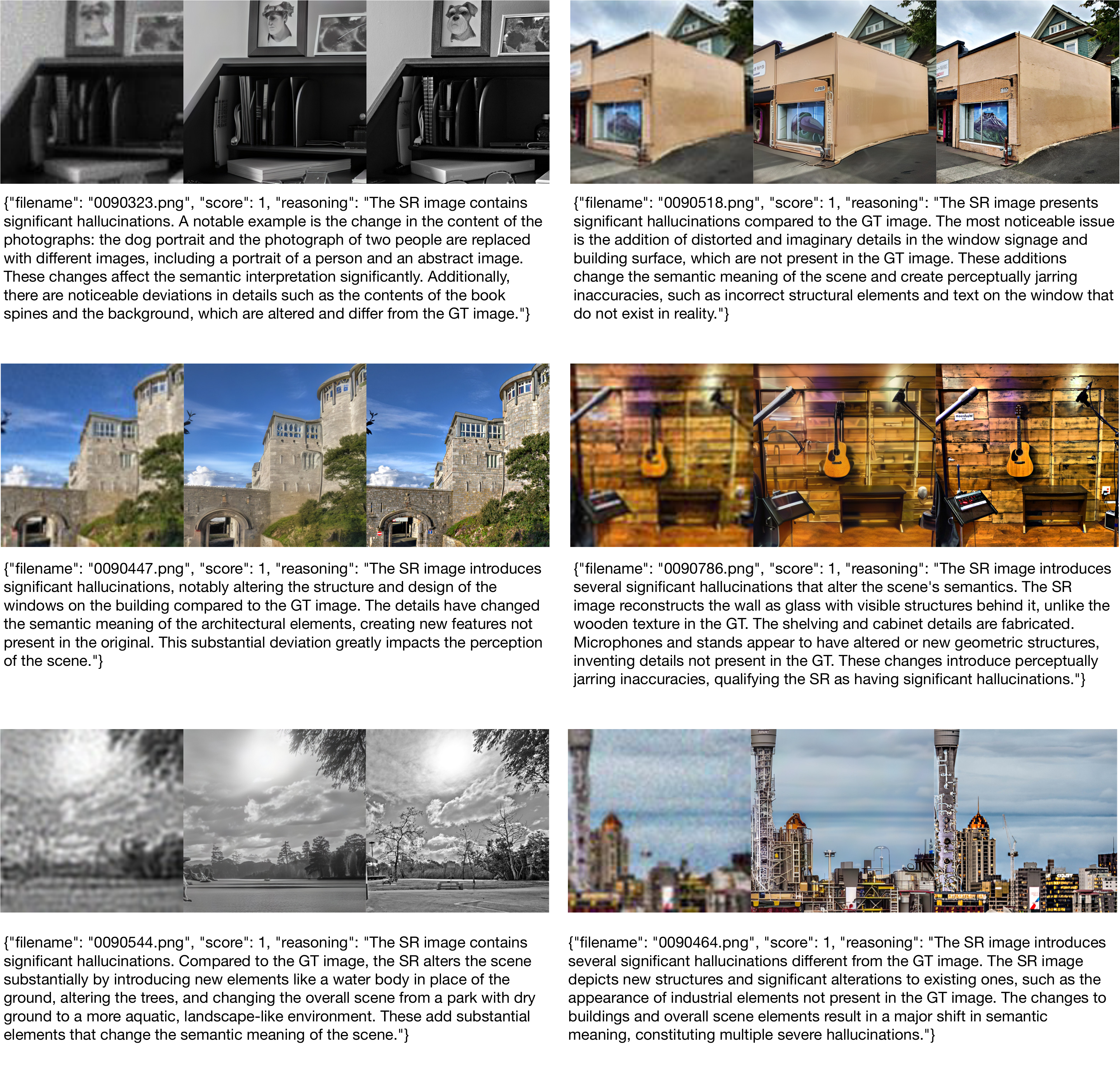}
    \caption{In this figure, we show six example outputs from the GPT-4o given the LRI (left), SRI (middle), GTI (right) and the prompt as inputs. Each output includes a numerical score on a 1-5 scale accompanied by detailed explanations justifying the assigned score. The results demonstrate the MLLM's ability to effectively identify critical hallucination issues in each image and assign accurate hallucination scores accordingly. Images are from the PASD outputs on the images in LSDIR training set. Note that PASD is not trained on LSDIR.}
    \label{fig:mllm_sample_outputs_pasd_supp}
\end{figure*}

\begin{figure*}[t]
    \centering
    \includegraphics[width=0.99\textwidth]{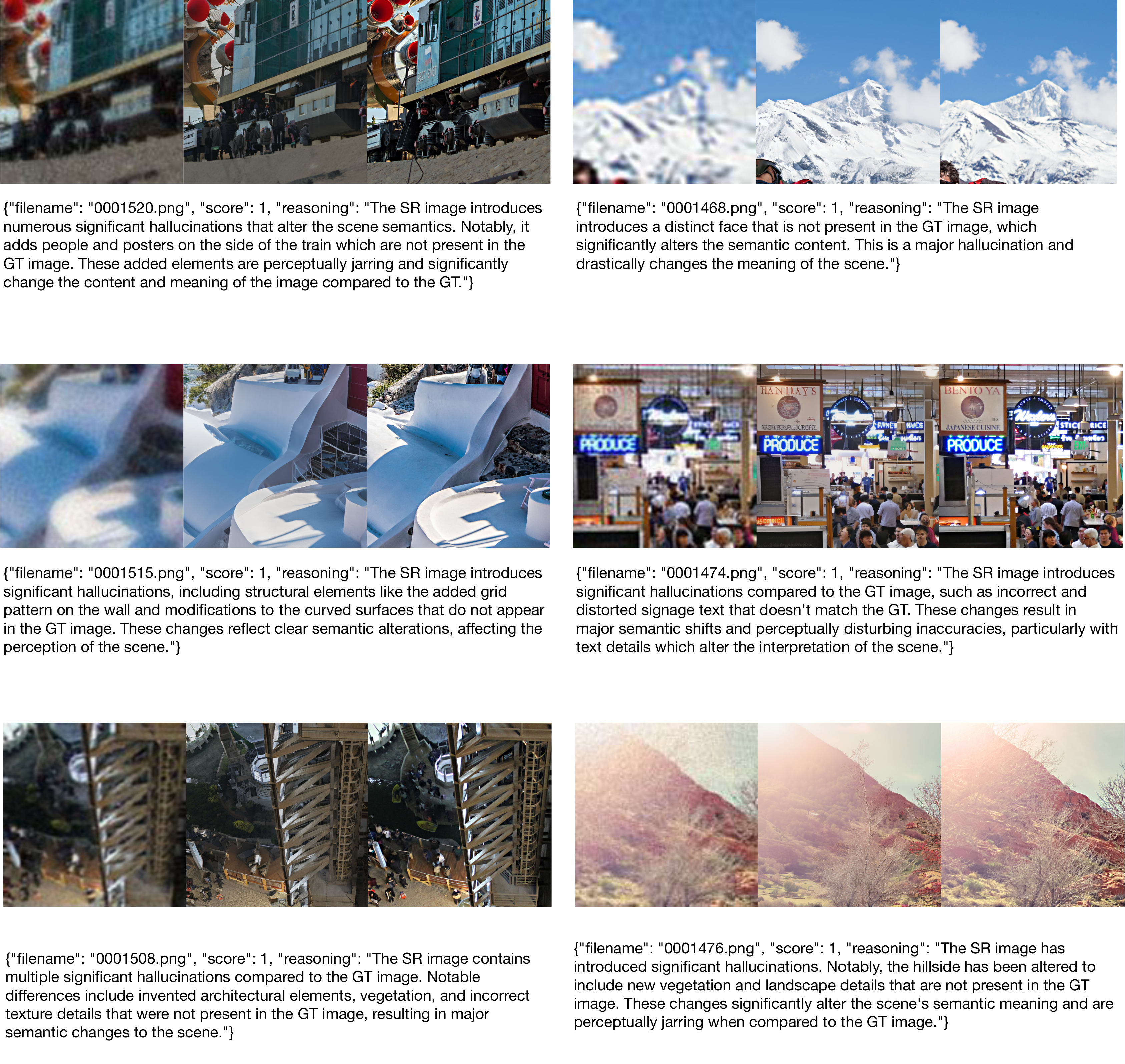}
    \caption{In this figure, we show six example outputs from the GPT-4o given the LRI (left), SRI (middle), GTI (right) and the prompt as inputs. Each output includes a numerical score on a 1-5 scale accompanied by detailed explanations justifying the assigned score. The results demonstrate the MLLM's ability to effectively identify critical hallucination issues in each image and assign accurate hallucination scores accordingly. Images are from the SeeSR outputs on the DIV2k training set. Note that SeeSR is not trained on DIV2k.}
    \label{fig:mllm_sample_outputs_seesr_supp}
\end{figure*}

\begin{figure*}[t]
    \centering
    \includegraphics[width=0.99\textwidth]{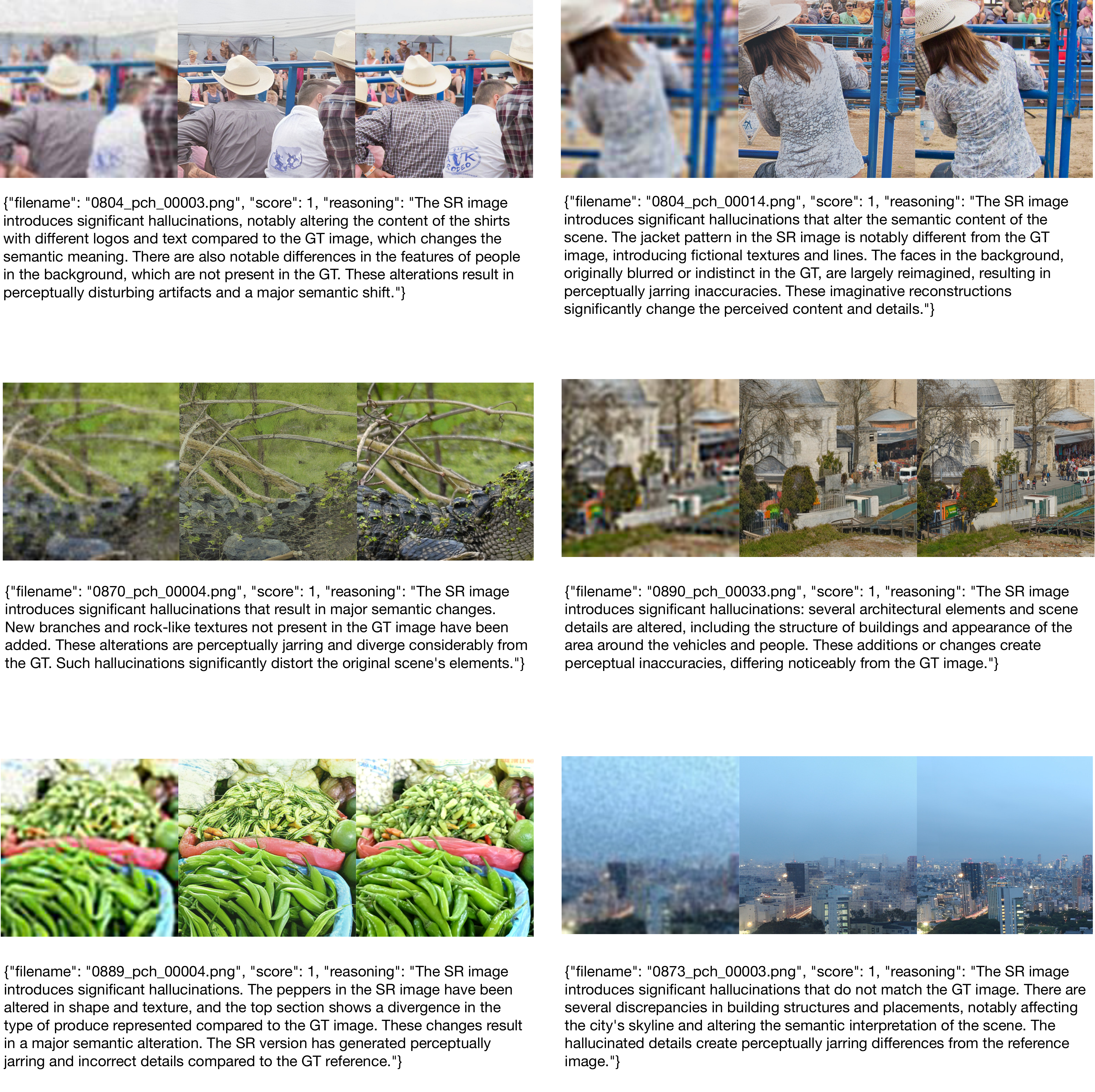}
    \caption{In this figure, we show six example outputs from the GPT-4o given the LRI (left), SRI (middle), GTI (right) and the prompt as inputs. Each output includes a numerical score on a 1-5 scale accompanied by detailed explanations justifying the assigned score. The results demonstrate the MLLM's ability to effectively identify critical hallucination issues in each image and assign accurate hallucination scores accordingly. Images are from the StableSR outputs on the DIV2k validation set.}
    \label{fig:mllm_sample_outputs_stablesr_supp}
\end{figure*}

\begin{figure*}[t]
    \centering
    \includegraphics[width=0.9\textwidth]{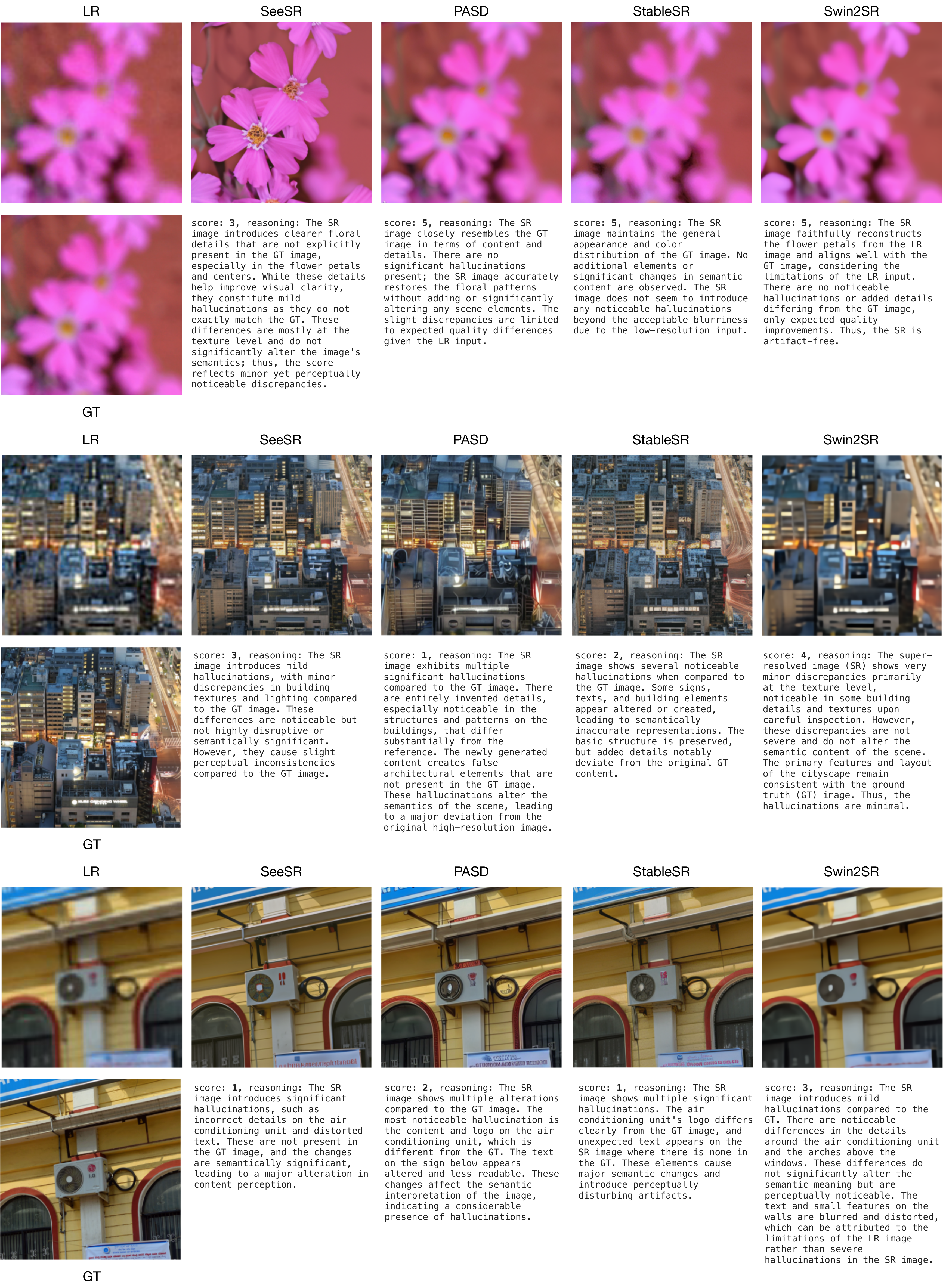}
    \vspace{-3.1mm}
    \caption{In this figure, we show more example outputs from GPT-4o (GPT-HS) given the LR, SR, and GT images, plus the prompt, as inputs. Each output includes a numerical score on a 1-5 scale accompanied by detailed explanations justifying the assigned score. The results demonstrate the MLLM's ability to effectively identify critical hallucination issues in each image and assign accurate hallucination scores accordingly. Images are from the SS-TS test set.}
    \label{fig:mllm_sample_gpt_score_supp}
\end{figure*}

To better understand hallucination issues in state-of-the-art diffusion-based generative SR models, we provide more example GPT-HS outputs for PASD (Fig.~\ref{fig:mllm_sample_outputs_pasd_supp}), SeeSR (Fig.~\ref{fig:mllm_sample_outputs_seesr_supp}), and StableSR (Fig.~\ref{fig:mllm_sample_outputs_stablesr_supp}) 
focusing on instances
{with severe hallucinations, which are the motivation for this work.} 
For each example, we show the LRI (left), SRI (middle), GTI (right), and outputs from the MLLM. Moreover, we show additional example outputs with minor or moderate hallucinations in Fig.~\ref{fig:mllm_sample_gpt_score_supp}. 
In all of these examples, we can clearly see that the MLLM is able to identify different types of hallucinations in the SR outputs across various scenarios. 

\end{document}